\documentclass[10pt,twocolumn,letterpaper]{article}

\usepackage{iccv}
\usepackage{times}
\usepackage{epsfig}
\usepackage{graphicx}
\usepackage{amsmath}
\usepackage{amssymb}

\usepackage{multirow} 
\usepackage{makecell} 
\usepackage[table,dvipsnames]{xcolor} 
\usepackage{colortbl} 
\usepackage{diagbox} 
\usepackage{siunitx}
\usepackage{enumitem} 
\usepackage{arydshln} 
\usepackage{tabularx}
\usepackage{booktabs} 
\usepackage{xspace} 
\usepackage[subrefformat=parens,labelformat=parens]{subcaption} 
\usepackage[Export]{adjustbox} 
\usepackage{pifont} 
\usepackage{balance} 

\usepackage{titling}
\pretitle{\begin{center}\Large \bf}
\posttitle{\par\end{center} \vskip 0.5em}
\preauthor{\begin{center} \large \begin{tabular}[t]{c}}
\postauthor{\end{tabular}\par\end{center}}
\predate{\begin{center}\large}
\postdate{\par\end{center} \vskip -0.75em}

\usepackage[pagebackref=true,breaklinks=true,colorlinks,bookmarks=false]{hyperref}

\usepackage[capitalize]{cleveref}

\def\eg{\emph{e.g.}\xspace} 
\def\ie{\emph{i.e.}\xspace}

\def\etal{\emph{et al.}\xspace}

\newcommand{\ours}{Point-SLAM\xspace}
\newcommand{\boldparagraph}[1]{\vspace{0.1em}\noindent{\bf #1}}

\newcommand{\greencheck}{{\color{green}\checkmark}}
\newcommand{\redx}{{\color{red}\ding{55}}}

\colorlet{colorFst}{Green!25}       
\colorlet{colorSnd}{SpringGreen!45} 
\colorlet{colorTrd}{Yellow!30}      
\colorlet{colorLow}{darkgray!30}    
\newcommand{\fs}{\cellcolor{colorFst}\bf}   
\newcommand{\nd}{\cellcolor{colorSnd}}      
\newcommand{\rd}{\cellcolor{colorTrd}}      
\newcommand{\lo}{\color{colorLow}}          

\definecolor{gray}{rgb}{0.65,0.65,0.65}
\definecolor{mycol}{rgb}{0.90,0.95,1.0}

\iccvfinalcopy 

\def\iccvPaperID{1873} 

\begin{document}

\title{Point-SLAM: Dense Neural Point Cloud-based SLAM}

\author{
Erik~Sandström$^{1}$\thanks{Equal contribution.} \hspace{3em} 
Yue Li$^{1}$\footnotemark[1] \hspace{3em} 
Luc~Van~Gool$^{1,2}$ \hspace{3em} 
Martin~R.~Oswald$^{1,3}$ \\
$^{1}$ETH Zürich, Switzerland \hspace{3em}
$^{2}$KU Leuven, Belgium \hspace{3em}
$^{3}$University of Amsterdam, Netherlands
}

\date{}

\maketitle

\begin{abstract}
  We propose a dense neural simultaneous localization and mapping (SLAM) approach for monocular RGBD input which anchors the features of a neural scene representation in a point cloud that is iteratively generated in an input-dependent data-driven manner. We demonstrate that both tracking and mapping can be performed with the same point-based neural scene representation by minimizing an RGBD-based re-rendering loss.
  In contrast to recent dense neural SLAM methods which anchor the scene features in a sparse grid, our point-based approach allows dynamically adapting the anchor point density to the information density of the input.
  This strategy reduces runtime and memory usage in regions with fewer details and dedicates higher point density to resolve fine details.
  Our approach performs either better or competitive to existing dense neural RGBD SLAM methods in tracking, mapping and rendering accuracy on the Replica, TUM-RGBD and ScanNet datasets. The source code is available at \url{https://github.com/eriksandstroem/Point-SLAM}.
\end{abstract}

\section{Introduction}
%
Dense visual simultaneous localization and mapping (SLAM) is a long-standing problem in computer vision where dense maps have widespread applications in augmented and virtual reality (AR, VR), robot navigation and planning tasks~\cite{hane2011stereo}, collision detection~\cite{Chen2023catnips}, detailed occlusion reasoning~\cite{ross2022bev}, and interpretation~\cite{yu2018ds} of scene content which is vital for scene understanding and perception.
\begin{figure}[t]
  \centering
  \footnotesize
  \setlength{\tabcolsep}{1pt}
  \renewcommand{\arraystretch}{1}
  \newcommand{\sz}{0.48}
  \begin{tabular}{ccc}
    & NICE-SLAM~\cite{zhu2022nice}  & \ours (ours) \\
    \rotatebox{90}{\hspace{12pt}Anchor Points} &
    \includegraphics[width=\sz\linewidth]{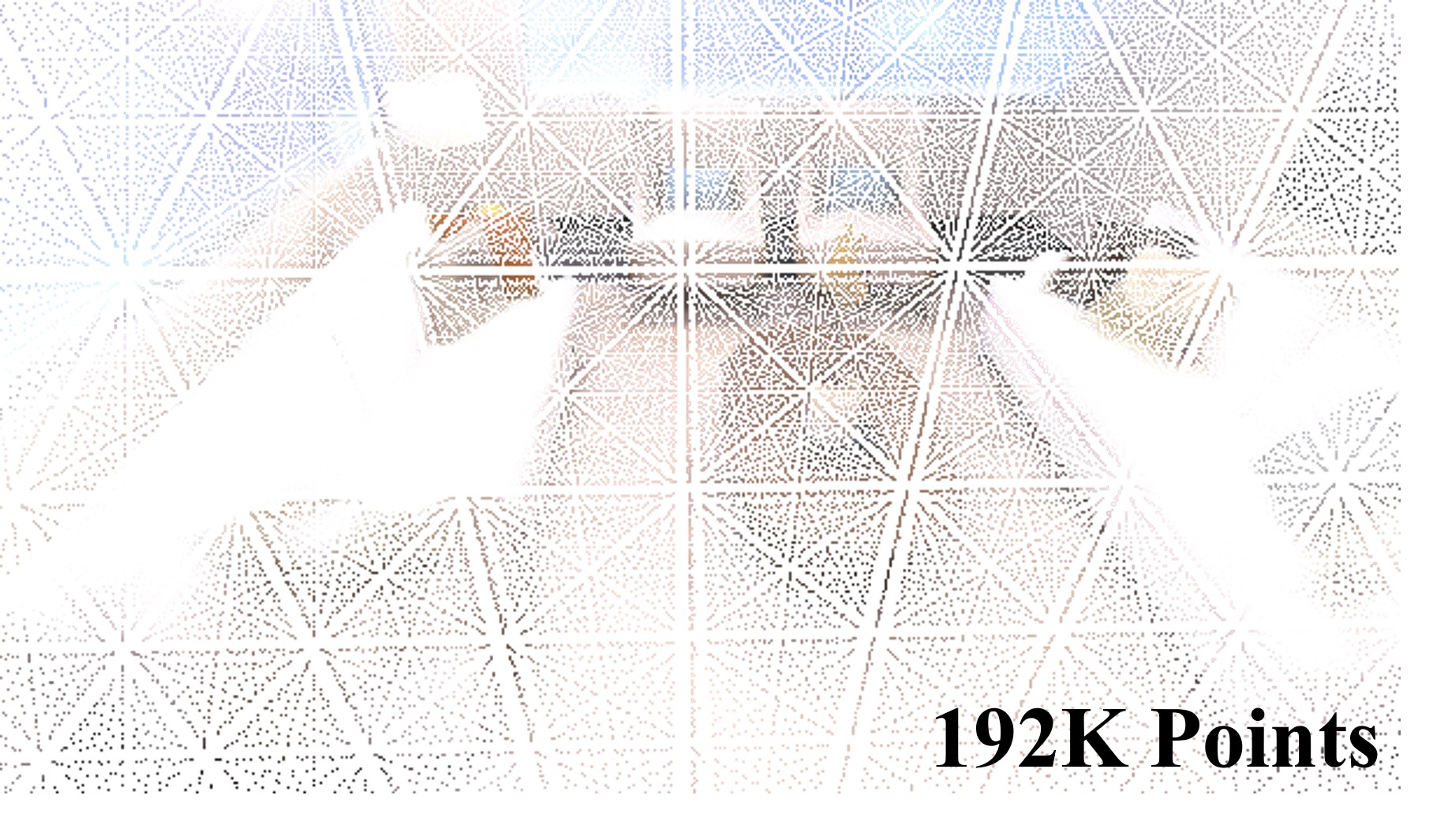} & 
    \includegraphics[width=\sz\linewidth]{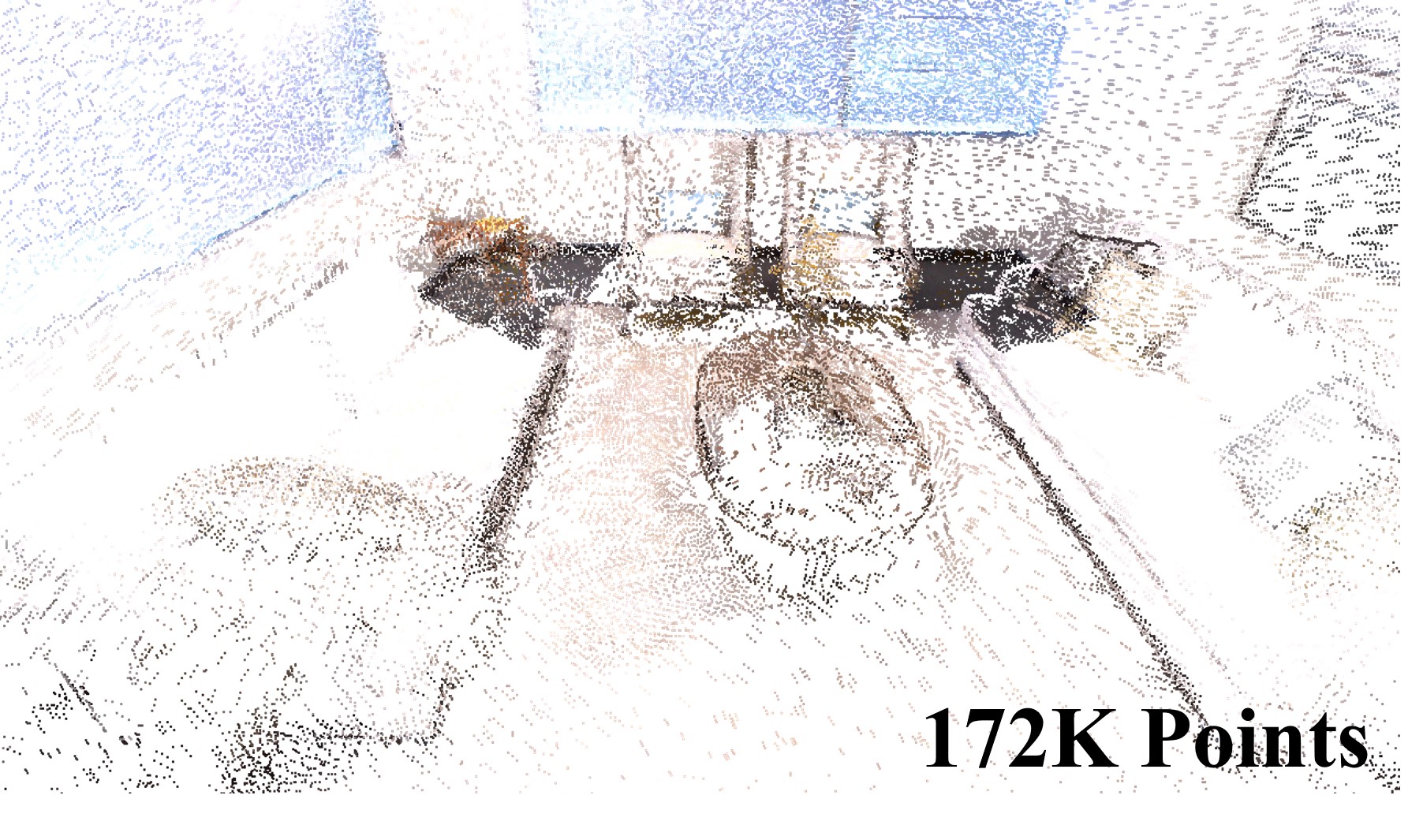} \\
    \rotatebox{90}{\hspace{18pt}Rendering} &
    \includegraphics[width=\sz\linewidth]{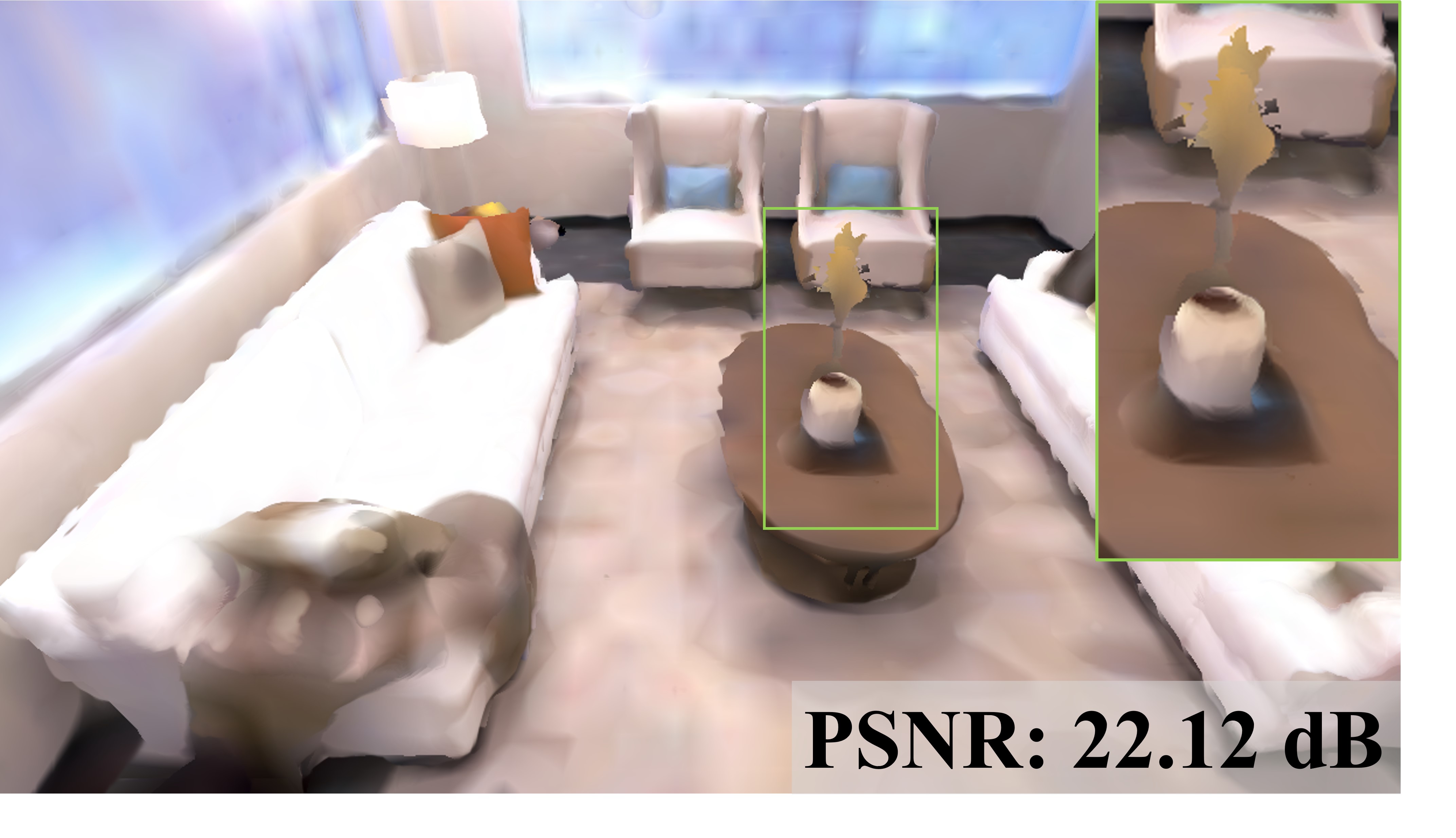} & 
    \includegraphics[width=\sz\linewidth]{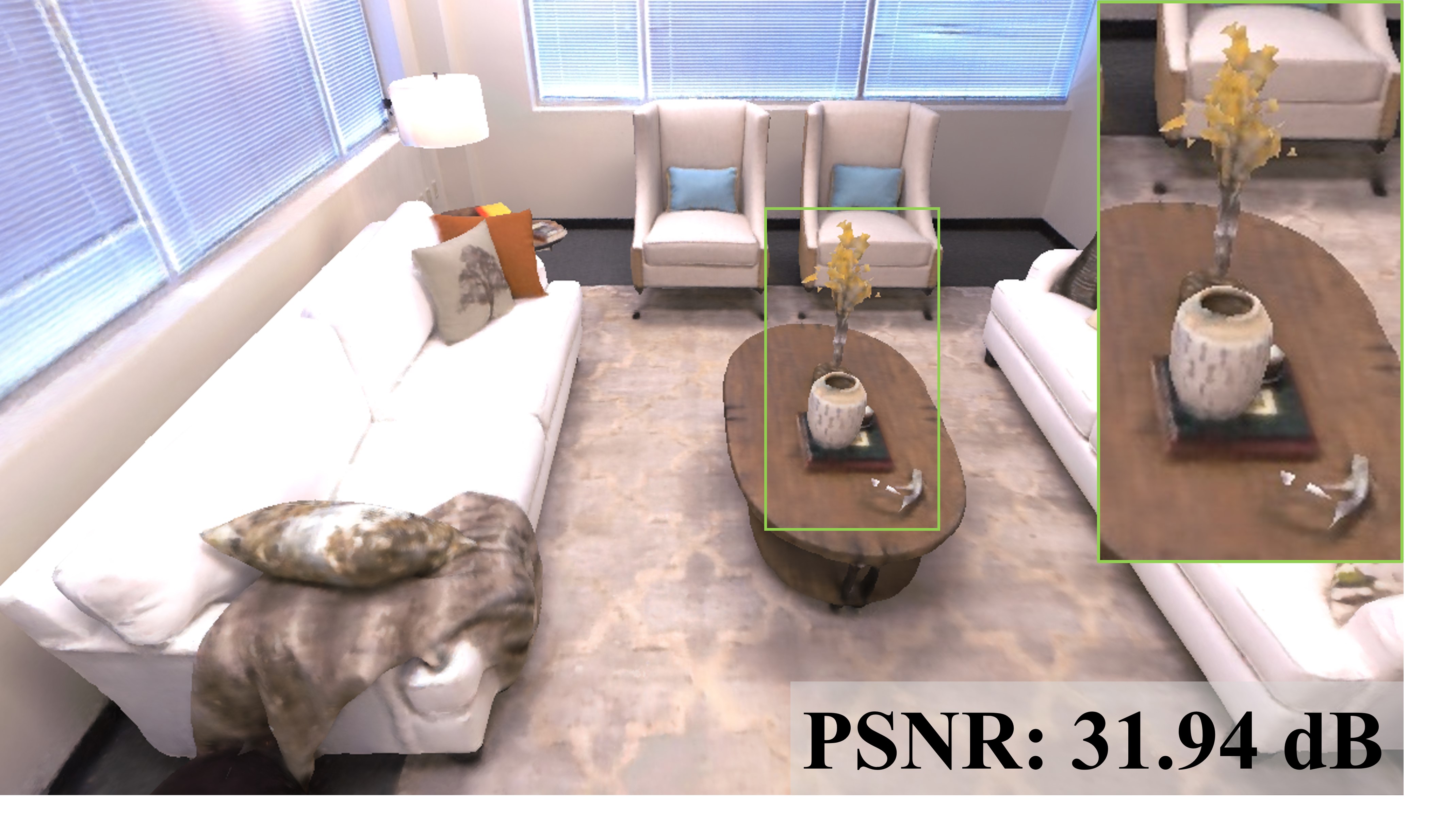} \\
  \end{tabular}    
  \begin{tabular}{cc}
    \rotatebox{90}{\hspace{9pt}Ground Truth} &
    \includegraphics[width=\sz\linewidth]{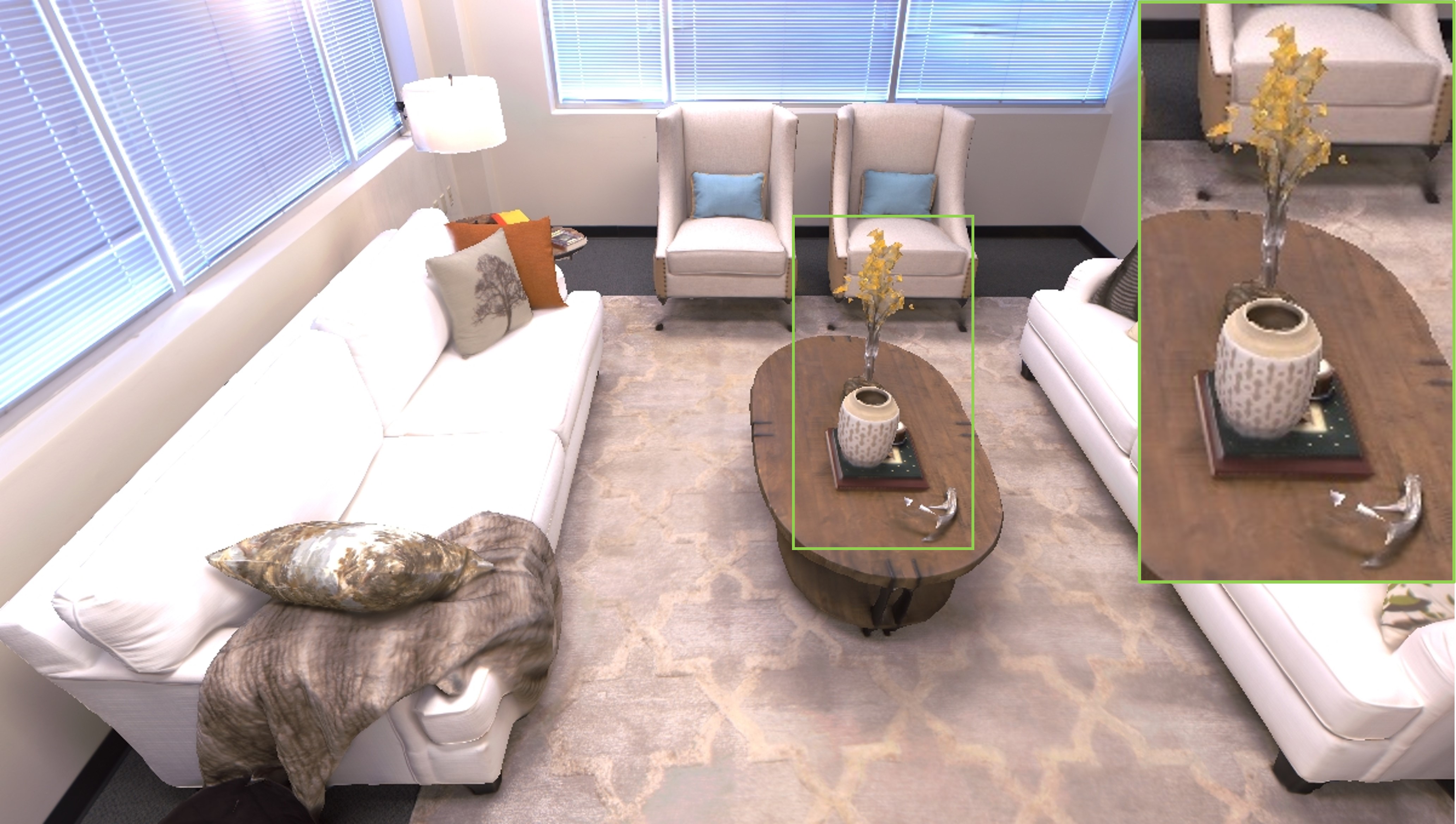} \\
  \end{tabular}
  \caption{\textbf{\ours Benefits.} Due to the spatially adaptive anchoring of neural features, \ours can encode high-frequency details more effectively than NICE-SLAM which leads to superior performance in rendering, reconstruction and tracking accuracy while attaining competitive runtime and memory usage.
  The \textbf{first row} shows the feature anchor points. For NICE-SLAM we show the centers of non-empty voxels located on a regular grid, while the density of anchor points for \ours depends on depth and image gradients. The row below depicts resulting renderings showing substantial differences on areas with high-frequency textures like the vase, blinds, floor or blanket.
  }
  \label{fig:teaser}
\end{figure}

To estimate a dense map via SLAM, tracking and mapping steps have traditionally been employed with different scene representations which creates undesirable data redundancy and independence since the tracking is then often performed independently of the estimated dense map.
Camera \textbf{tracking} is frequently done with sparse point clouds or depth maps, \eg via frame-to-model tracking~\cite{newcombe2011kinectfusion,whelan2012kintinuous,chen2013scalable,niessner2013voxel_hashing,Kahler2015infiniTAM} and with incorporated loop closures~\cite{dai2017bundlefusion,zhou2013elastic,cao2018real}.
For dense \textbf{mapping} the most common scene representations are voxel grids~\cite{newcombe2011kinectfusion,newcombe2011dtam}, voxel hashing~\cite{niessner2013voxel_hashing,dai2017bundlefusion,Kahler2015infiniTAM,kahler2015hierarchical}, octrees~\cite{fuhrmann2011fusion,steinbrucker2013large,marniok2017efficient}, or point/surfel clouds~\cite{zhou2013elastic,cao2018real,schops2019bad}.
The introduction of learned scene representations~\cite{park2019deepsdf,mescheder2019occupancy,chen2019learning,Mildenhall2020NeRF:Synthesis} has led to rapid progress for learning-based online mapping methods~\cite{Weder2020RoutedFusion,weder2021neuralfusion,mihajlovic2021deepsurfels,huang2021di,li2022bnv,ortiz2022isdf} and offline methods~\cite{peng2020convolutional,azinovic2022neural,wang2022gosurf,yu2022monoSDF}. 
However, most of these methods require ground truth depth or 3D for model training and may not generalize to unseen real-world scenarios at test time. 
To eliminate the potential domain gap between train and test time, recent SLAM methods rely on test time optimization via volume rendering~\cite{Sucar2021IMAP:Real-Time,yang2022vox,zhu2022nice}.
Compared to traditional approaches, neural scene representations have attractive properties for mapping like improved noise and outlier handling~\cite{weder2021neuralfusion}, better hole filling and inpainting capabilities for unobserved scene parts~\cite{yang2022vox,zhu2022nice}, and data compression~\cite{park2019deepsdf, wang2022neuris}. 
Like DTAM~\cite{newcombe2011dtam} or BAD-SLAM~\cite{schops2019bad} recent neural SLAM methods~\cite{zhu2022nice,yang2022vox,Sucar2021IMAP:Real-Time} only use a single scene representation for both tracking and mapping but they rely either on a regular grid structure~\cite{zhu2022nice,yang2022vox} or a single MLP~\cite{Sucar2021IMAP:Real-Time}.
Inspired by BAD-SLAM~\cite{schops2019bad}, NICE-SLAM~\cite{zhu2022nice} and Point-NeRF~\cite{xu2022point}, the research question we tackle in this work is:
\begin{center}
  \emph{Can point-based neural scene representations be used for tracking and mapping for real-time capable SLAM?}
\end{center}
To this end, we introduce \ours, a point-based solution to dense RGBD SLAM, which allows for a data-adaptive scene encoding.
The key ideas of our method are as follows:
Instead of anchoring the feature points on a regular grid, our approach populates points adaptively depending on information density in the input data which allows for a better memory vs. accuracy trade-off.
For rendering, we depart from the classical splatting technique used for surfels and instead aggregate neural point features in a ray-marching fashion. 
MLP decoders translate these features into scene geometry and color estimates.
Tracking and mapping are performed alternatingly by minimizing an RGBD-based re-rendering loss.
Different from grid-based approaches, we do not model free space and encode only little information around the surface.
We evaluate our proposed method on a selection of indoor RGBD datasets and demonstrate state-of-the-art performance on dense neural RGBD SLAM in terms of tracking, rendering, and mapping - see \cref{fig:teaser} for exemplary results. 
In summary, our \textbf{contributions} include:
\begin{itemize}[itemsep=0pt,topsep=2pt,leftmargin=10pt]
  \item We present \ours, a real-time capable dense RGBD SLAM approach which anchors neural features in a point cloud that grows iteratively in a data-driven manner during scene exploration. We demonstrate that the proposed neural point-based scene representation can be effectively used for both mapping and tracking.
  \item We propose a dynamic point density strategy which allows for computational and memory efficiency gains and trade reconstruction accuracy against speed and memory.
  \item Our approach shows clear benefits on a variety of datasets in terms of tracking, rendering and mapping accuracy.
\end{itemize}
\section{Related Work} \label{sec:rel}

\boldparagraph{Dense Visual SLAM and Mapping.} 
\label{rel-dense-slam}
Curless and Levoy~\cite{curless1996volumetric} laid the groundwork for many 3D reconstruction strategies that employ truncated signed distance functions (TSDF). 
Subsequent developments include KinectFusion~\cite{newcombe2011kinectfusion} and more scalable techniques with voxel hashing~\cite{niessner2013voxel_hashing,Kahler2015infiniTAM,Oleynikova2017voxblox}, octrees~\cite{steinbrucker2013large}, and pose robustness via sparse image features~\cite{7900211}. 
Further extensions involve tracking for SLAM~\cite{newcombe2011dtam,schops2019bad,Sucar2021IMAP:Real-Time,zhu2022nice,cao2018real,yang2022fd} which can also handle loop closures, like BundleFusion~\cite{dai2017bundlefusion}. 
To address the issue of noisy depth maps, RoutedFusion~\cite{Weder2020RoutedFusion} learns a fusion network that outputs the TSDF update of the volumetric grid. NeuralFusion~\cite{weder2021neuralfusion} and DI-Fusion~\cite{huang2021di} extend this concept by learning the scene representation implicitly, resulting in better outlier handling. 
A number of recent works do not need depth input and accomplish dense online reconstruction from RGB cameras only ~\cite{murez2020atlas,choe2021volumefusion,bovzivc2021transformerfusion,stier2021vortx,sun2021neuralrecon,sayed2022simplerecon,li2023dense}.
Lately, methods relying on test time optimization have become popular due to their adaptability to test time constraints. 
For example, Continuous Neural Mapping~\cite{yan2021continual} learns a representation of the scene by means of continually mapping from a sequence of depth maps. 
Neural Radiance Fields~\cite{Mildenhall2020NeRF:Synthesis} inspired works for dense surface reconstruction~\cite{Oechsle2021UNISURF:Reconstruction,Wang2021NeuS:Reconstruction} and pose estimation~\cite{Rosinol2022NeRF-SLAM:Fields,Lin2021BARF:Fields,wang2021nerf,bian2022nope}. 
These works have led to full dense SLAM pipelines~\cite{yang2022vox,zhu2022nice,Sucar2021IMAP:Real-Time,mahdi2022eslam}, which represent the current most promising trend towards accurate and robust visual SLAM. 
See \cite{zollhofer2018state} for a survey on online RGBD reconstruction.
In contrast to our work, none of the neural SLAM approaches supports an input-adaptive scene encoding with high fidelity.

Concurrent to our work, ESLAM~\cite{mahdi2022eslam} tackles RGBD SLAM with axis aligned feature planes and NICER-SLAM~\cite{zhu2023nicer}, NeRF-SLAM~\cite{Rosinol2022NeRF-SLAM:Fields} and Orbeez-SLAM~\cite{chung2022orbeez} focus on RGB-only SLAM.

\begin{figure*}[ht!]
\centering
 \includegraphics[width=\linewidth]{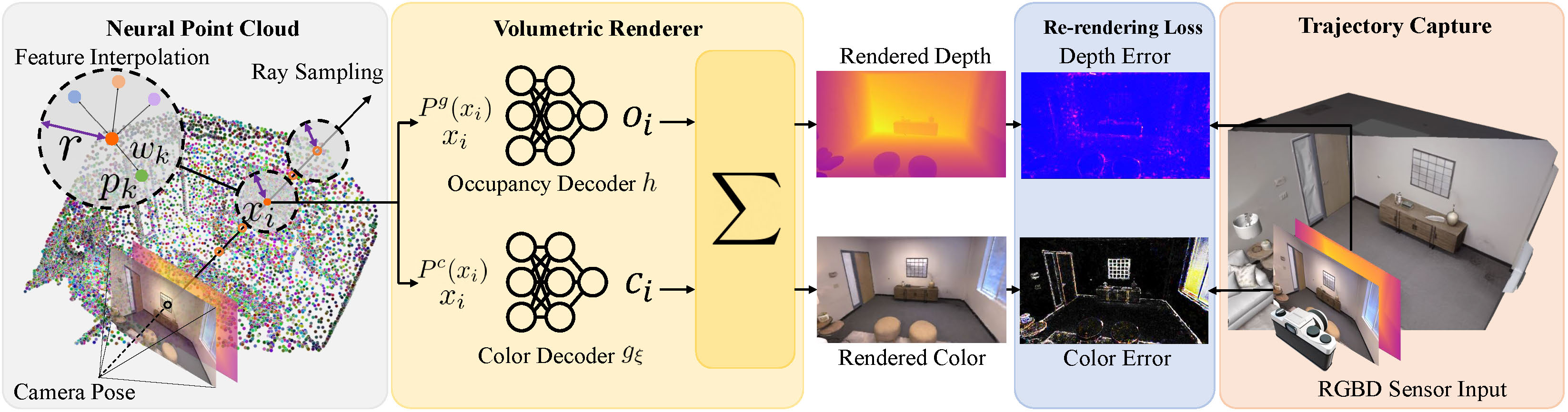}\\
\caption{\textbf{\ours{} Architecture.} Given an estimated camera pose, mapping is performed as follows. We first add a sparse set of neural points to the neural point cloud, and then render depth and color images via volume rendering along the ray. 
For each sampled pixel we sample a set of points $x_i$ along the ray and extract the geometric and color features ($P^g(x_i)$ and $P^c(x_i)$ resp.) at $x_i$, using feature interpolation within the spherical \textcolor{Fuchsia}{search radius r}. Each neural point location $p_k$ is weighted by the distance $w_k$ to the sampled point $x_i$. The features are passed to the occupancy and color decoders ($h$ and $g_{\xi}$ resp.) along with the point coordinate $x_i$ to extract the occupancy $ \mathbf{o}_{i}$ and color $\mathbf{c_i}$. 
By imposing a depth and color re-rendering loss to the sensor input RGBD frame, the neural point features are optimized during mapping. Alternating to the mapping step, we perform tracking by optimizing the camera extrinsics while keeping the map fixed.
}
\label{fig:architecture}
\end{figure*}

\boldparagraph{Scene Representations.}
Most dense 3D reconstruction works can be separated into three categories: (1) \textit{grid-based}, (2) \textit{point-based}, (3) \textit{network-based}.
The \textit{grid-based} representation is perhaps the most explored one and can be further split into methods using dense grids~\cite{zhu2022nice,newcombe2011kinectfusion,Weder2020RoutedFusion,weder2021neuralfusion,curless1996volumetric,sun2021neuralrecon,bovzivc2021transformerfusion,li2022bnv,choi2015robust,zhou2013elastic,zhou2013dense,whelan2012kintinuous,zou2022mononeuralfusion}, hierarchical octrees~\cite{yang2022vox,steinbrucker2013large,marniok2017efficient,chen2013scalable,liu2020neural} and voxel hashing~\cite{niessner2013voxel_hashing,Kahler2015infiniTAM,dai2017bundlefusion,wang2022neus2,muller2022instant} to save memory. 
One advantage of grids is that neighborhood look ups and context aggregations are fast and straightforward.
As their main limitation, the grid resolution needs to be specified beforehand and cannot be trivially adapted during reconstruction, even for octrees. 
This can lead to a suboptimal resolution strategy where memory is wasted in areas with little complexity while not being able to resolve details beyond the resolution choice.
\textit{Point-based} representations offer a solution to the issues facing grids and have successfully been applied to 3D reconstruction~\cite{whelan2015elasticfusion,schops2019bad,cao2018real,chung2022orbeez,Kahler2015infiniTAM,keller2013real,cho2021sp,zhang2020dense}. 
For example, analogous to the resolution in grids, the point density does not need to be specified beforehand and can inherently vary across the scene. 
Further, point sets can be trivially focused around the surface in order not to waste memory on modeling free space. The penalty for this flexibility is a more difficult neighborhood search problem as point sets lack connectivity structure. 
For dense SLAM, neighborhood search can be accelerated by converting the 3D search problem into a 2D one by projecting the point set into a set of keyframes~\cite{whelan2015elasticfusion,schops2019bad}. 
A more elegant and faster solution is to 
register each point within a grid structure~\cite{xu2022point}.
In this work, we argue that points provide a flexible representation that can benefit from a grid structure for fast neighborhood search. 
Contrary to previous point- or surfel-based SLAM approaches~\cite{whelan2015elasticfusion,schops2019bad,cao2018real}, we benefit from neural implicit features from which rendering is performed through volumetric alpha compositing. 
\textit{Network-based} methods for dense 3D reconstruction offer a continuous representation by modeling the global scene implicitly through coordinate-MLPs~\cite{azinovic2022neural,Sucar2021IMAP:Real-Time,Wang2021NeuS:Reconstruction,Rosinol2022NeRF-SLAM:Fields,ortiz2022isdf,yan2021continual,yariv2021volume,park2019deepsdf,mescheder2019occupancy}.
Benefiting from a simple formulation that is continuous and compressed, network-based methods can recover maps and textures of high quality, but are not suitable for online scene reconstruction for two main reasons: 1) they do not allow for local scene updates, 2) for growing scene size the network capacity cannot be increased at runtime.
In this work, we adopt neural implicit representations popularized by network-based methods, but allow for scalability and local updates by anchoring neural point features in 3D space.

Outside the domain of the aforementioned three groups, a few works have studied other representations such as parameterized surface elements~\cite{vakalopoulou2018atlasnet} and axis aligned feature planes~\cite{mahdi2022eslam,peng2020convolutional}. 
Parameterized surface elements generally struggle with formulating a flexible shape template while feature planes struggle with scene reconstructions containing multiple surfaces, due to their overly compressed representation. 
Therefore, we believe that these approaches are not suitable for dense SLAM. 
Instead we look to model our scene space as a collection of unordered points with corresponding optimizable features.

\section{Method} \label{sec:methods}
%
This section details how our neural point cloud is deployed as the sole representation for dense RGBD SLAM. 
Given an estimated camera pose, points are iteratively added to the scene as new areas are explored (\cref{sec:npc_rep}). 
We make use of per-pixel image gradients to achieve a dynamic point density which aids in resolving fine details while compressing the representation elsewhere. 
We further detail how depth and color rendering is performed (\cref{sec:rendering}), with which we minimize a re-rendering loss for both mapping and tracking (\cref{sec:map_and_track}). 
An overview of our method is provided in \cref{fig:architecture}.

\subsection{Neural Point Cloud Representation}  \label{sec:npc_rep}
%
We define our neural point cloud as a set of $N$ neural points
\begin{equation}
  P = \{(p_i, f^g_i, f^c_i) \, | \, i=1,...,N\} \enspace,
\end{equation}
each anchored at location $p_i \in \mathbb{R}^3$ and with a geometric and color feature descriptor $f^g_i \in \mathbb{R}^{32}$ and $f^c_i \in \mathbb{R}^{32}$. 

\boldparagraph{Point Adding Strategy.} 
For every mapping phase and a given estimated camera pose, we sample $X$ pixels uniformly across the image plane and $Y$ pixels among the top $5Y$ pixels with the highest color gradient magnitude. 
Using the available depth information, the pixels are unprojected into 3D where we search for neighbors within a radius $r$. 
If no neighbors are found, we add three neural points along the ray, centered at the depth reading $D$ and then offset by $(1 - \rho)D$ and $(1 + \rho)D$ with $\rho \in (0,1)$ being a hyperparameter accounting for the expected depth noise. 
If neighbors are found, no points are added.
We use a normally distributed initialization of the feature vectors. 
The three points act as a limited update band that is depth dependent in order to model the common noise characteristic of depth cameras. 
As more frames are processed, our neural point cloud grows progressively to represent the exploration of the scene, but converges to a bounded set of points when no new scene parts are visited. 
Contrary to many voxel-based representations, it is not required to specify any scene bounds before the reconstruction. 

\boldparagraph{Dynamic Resolution.} For computational and memory efficiency, we employ a dynamic point density across the scene. 
This allows \ours to efficiently model regions with few details while high point densities are imposed where it is needed to resolve fine details. 
We implement this by allowing the nearest neighbor search radius $r$ to vary according to the color gradient observed from the sensor. 
We use a clamped linear mapping to define the search radius $r$ based on the color gradient:
\begin{equation}
  \!\!r(u,v)= 
  \begin{cases}
    r_l & \text{if } \nabla I(u,v) \geq g_u \\
    \beta_1\nabla I(u,v) \!+\! \beta_2 & \text{if } g_l \leq \nabla I(u,v) \leq g_u \\
    r_u & \text{if } \nabla I(u,v) \leq g_l \enspace, \\
  \end{cases}
  \label{eq:r}
\end{equation}
where $\nabla I(u,v)$ denotes the gradient magnitude at the pixel location $(u,v)$. 
We use a lower and upper bound $(r_l, r_u)$ for the search radius to control the compression level and memory usage. 
For more details about parameter choices, we refer to the supplementary material.

\subsection{Rendering}  \label{sec:rendering}
%
To render depth and color, we adopt a volume rendering strategy. 
Given a camera pose with origin $\mathbf{O}$, we sample a set of points $x_i$ as
\begin{align}
    x_i = \mathbf{O} + z_i\mathbf{d}, \quad i \in \{1, ..., M\} \enspace,
    \label{eq:point-sample}
\end{align}
where $z_i \in \mathbb{R}$ is the point depth and $d \in \mathbb{R}^3$ the ray direction. Specifically, we sample $5$ points spread evenly between $(1 - \rho)D$ and $(1 + \rho)D$, where $D$ is the sensor depth at the pixel to be rendered. 
This is in contrast to voxel-based frameworks~\cite{zhu2022nice,yang2022vox} which need to carve the empty space between the camera and the surface, thus requiring significantly more samples. 
For example, NICE-SLAM~\cite{zhu2022nice} uses 48 samples (16 around the surface and 32 between the camera and the surface). 
With fewer samples along the ray, we achieve a computational speed-up during rendering. 
After the points $x_i$ have been sampled, the occupancies $\mathrm{o}_{i}$ and colors $\mathbf{c}_i$ are decoded using MLPs following~\cite{zhu2022nice} as
\begin{align} 
    \mathrm{o}_{i} = h\big(x_i, P^g(x_i)\big) \qquad \mathbf{c}_i = g_\xi\big(x_i, P^c(x_i)\big) \enspace.
    \label{eq:occ-from-net}
\end{align}
We denote the geometry and color decoder MLPs by $h$ and $g_\xi$, respectively, where $\xi$ are the trainable parameters of $g$. 
We use the same architecture for $h$ and $g$ as~\cite{zhu2022nice} and use their provided pretrained and fixed middle geometric decoder $h$. 
The decoder input is the 3D point $x_i$, to which we apply
a learnable Gaussian positional encoding~\cite{tancik2020fourier} to mitigate the limited band-width of MLPs, and the associated feature. 
We further denote $P^g(x_i)$ and $P^c(x_i)$ as the geometric and color features extracted at point $x_i$ respectively. 
For each point $x_i$ we use the corresponding per-pixel query radius $2r$, where $r$ is computed according to \cref{eq:r}. 
Within the radius $2r$, we require to find at least two neighbors. 
Otherwise, the point is given zero occupancy. 
We use the closest eight neighbors and use inverse squared distance weighting for the geometric features, \ie
\begin{equation}
  \!\!\!
  P^g(x_i) = \sum_k \frac{w_k}{\sum_k w_k}f^g_k 
  \;\;\text{ with }
  w_k = \frac{1}{||p_k - x_i||^2} \enspace.
\end{equation}
For the color features, inspired by~\cite{xu2022point}, we impose a nonlinear preprocessing on the extracted neighbor features $f^c_k$ such that 
\begin{equation}
  f^c_{k, x_i} = F_{\theta}(f^c_k, p_k - x_i) \enspace,
\end{equation}
where $F$ is a one-layer MLP parameterized by $\theta$, with $128$ neurons and $\mathrm{softplus}$ activations. We use the same Gaussian positional encoding for the relative point vector $(p_k - x_i)$ as used by the geometry and color decoders. This yields 
\begin{equation}
  P^c(x_i) = \sum_k \frac{w_k}{\sum_k w_k}f^c_{k, x_i} \enspace.
\end{equation}
For pixels without depth observation, we render by marching along the ray from the depth $30 cm$ to $1.2D_{max}$, where $D_{max}$ is the maximum frame depth. 
We use $25$ samples within this interval. 
This technique acts as a hole filling technique, but does not fill in arbitrarily large holes, which can cause large completion errors.
Next, we describe how the per-point occupancies $\mathrm{o}_{i}$ and colors $\mathbf{c}_{i}$ are used to render the per-pixel depth and color using volume rendering. 
We construct a weighting function, $\alpha_i$ as described in \cref{eq:weight-func}. This weight represents the discretized probability that the ray terminates at point $x_i$.
\begin{align}
    \alpha_i = \mathrm{o}_{\mathbf{p}_i} \prod_{j=1}^{i-1}(1-\mathrm{o}_{\mathbf{p}_j}) \enspace.
    \label{eq:weight-func}
\end{align}
The rendered depth is computed as the weighted average of the depth values along each ray, and equivalently for the color according to \cref{eq:rgbd-render}.
\begin{align}
   \hat{D} = \sum_{i=1}^N \alpha_i z_i, \quad 
   \hat{I} = \sum_{i=1}^N \alpha_i \mathbf{c}_i
   \label{eq:rgbd-render}
\end{align}
We also compute the variance along the ray as 
\begin{align} 
   \hat{S}_{D} = \sum_{i=1}^N \alpha_i \big(\hat{D} - z_i \big)^2 
   \label{eq:epi-unc-render}
\end{align}
For more details, we refer to \cite{zhu2022nice}.
%
%
\subsection{Mapping and Tracking}  \label{sec:map_and_track}
%
\boldparagraph{Mapping.}
During mapping, we render $M$ pixels uniformly across the RGBD frame and minimize the re-rendering loss to the sensor reading $D$ and $I$ as 
\begin{align}
    \mathcal{L}_{map} = \sum_{m = 1}^M \lvert D_m - \hat{D}_m \rvert_1 + \lambda_m\lvert I_m - \hat{I}_m\rvert_1 \enspace,
    \label{eq:unc-loss-l1}
\end{align}
which combines a geometric $L_1$ depth loss and a color $L_1$ loss with hyperparameter $\lambda_m$ for given ground truth values $\hat{D}_m, \hat{I}_m$. 
The loss optimizes the geometric and color features $f^g$ and $f^c$ as well as the parameters $\xi$ and $\theta$ of the color decoder $g$ and interpolation decoder $F$ respectively. 
For each mapping phase, we first optimize using only the depth term in order to initialize the color optimization well. 
We then add the color loss for the remaining 60 $\%$ of iterations.
Following the same strategy as~\cite{zhu2022nice}, we make use of a database of keyframes to regularize the mapping loss.
We sample a set of keyframes which have a significant overlap with the viewing frustum of the current frame and add pixel samples from the keyframes. 
More details are provided in the supplementary material.

\boldparagraph{Tracking.}
In a separate process to mapping, we perform tracking by optimizing the camera extrinsics $\{\mathbf{R}, \mathbf{t}\}$ at each frame. 
We sample $M_t$ pixels across the frame and initialize the new pose with a simple constant speed assumption that transforms the last known pose with the relative transformation between the second last pose and the last pose. 
The tracking loss $\mathcal{L}_{\mathrm{track}}$ combines a color term weighted by $\lambda_t$ and a geometric term weighted by the standard deviation of the depth prediction:
\begin{align}
  \mathcal{L}_{\mathrm{track}} &= 
   \sum_{m=1}^{M_t} \frac{\lvert D_m - \hat{D}_m \rvert_1}
      {\sqrt{\hat{S}_{D}}} + \lambda_t \lvert I_m - \hat{I}_m\rvert_1 \enspace
  \label{eq:og-track-depth-loss}
\end{align}

\subsection{Exposure Compensation}
For scenes with significant exposure changes between frames, we use an additional module to reduce color differences between corresponding pixels. 
Inspired by~\cite{Rematas2021UrbanFields}, we learn a per-image latent vector which is fed as input to an exposure MLP $G_{\phi}$ with parameters $\phi$. 
The network $G$ is shared between frames and optimized at runtime. 
It outputs an affine transformation ($3\times 3$ matrix and $3\times 1$ translation) which is used to transform the color prediction from \cref{eq:rgbd-render} before being fed to the tracking or mapping loss. For more details see the supplementary material.

\section{Experiments}
\label{sec:exp}
We first describe our experimental setup and then evaluate our method against state-of-the-art dense neural RGBD SLAM methods on Replica~\cite{straub2019replica} as well as the real world TUM-RGBD~\cite{Sturm2012ASystems} and the ScanNet~\cite{Dai2017ScanNet} datasets.
Further experiments and details are in the supplementary material.

\boldparagraph{Implementation Details.}
For efficient nearest neighborhood search, we use the FAISS library~\cite{johnson2019billion} which supports GPU processing.
We use $\rho = 0.02$ on Replica and TUM-RGBD and $\rho = 0.04$ on ScanNet. 
We set $r_l = 0.02$, $r_u = 0.08$, $g_u = 0.15$, $g_l = 0.01$ and $\beta_1=-\frac{2}{3}$, $\beta_2 = \frac{13}{150}$. 
For all datasets, $X = 6000$. For Replica $Y = 1000$ and for ScanNet and TUM-RGBD $Y = 0$. For tracking, we sample $M_t=1.5K$ pixels uniformly on Replica. 
On TUM-RGBD and ScanNet, we first compute the top $75K$ pixels based on the image gradient magnitude and sample $M_t=5K$ out of this set. 
For mapping, we sample uniformly $M=5K$ pixels for Replica and $10K$ pixels for TUM-RGBD and ScanNet.
Although we specify a number of mapping iterations, we use an adaptive scheme which takes the number of newly added points into account.
The number of mapping iterations is computed as $m_i = m_i^dn/300$, where $m_i^d$ is the default mapping iterations and $n$ is the number of added points. 
We clip $m_i$ to lie within $[0.95m_i^d,2m_i^d]$. 
This strategy speeds up mapping when few points are added and helps optimize frames with many new points.
To mesh the scene, we render depth and color every fifth frame over the estimated trajectory and use TSDF Fusion~\cite{curless1996volumetric} with voxel size $1$ cm.
See the supplementary material for more details.

\begin{figure*}[t]
\centering
{\footnotesize
\setlength{\tabcolsep}{1pt}
\renewcommand{\arraystretch}{1.1}
\begin{tabular}{cc}
\resizebox{0.47\linewidth}{!}
{
\begin{subfigure}{0.64\linewidth}
\centering
\footnotesize
\setlength{\tabcolsep}{2pt}
\begin{tabularx}{\linewidth}{llccccccccc}
\toprule
Method & Metric & \texttt{Rm\thinspace0} & \texttt{Rm\thinspace1} & \texttt{Rm\thinspace2} & \texttt{Off\thinspace0} & \texttt{Off\thinspace1} & \texttt{Off\thinspace2} & \texttt{Off\thinspace3} & \texttt{Off\thinspace4} & Avg.\\
\midrule
\multirow{4}{*}{\makecell[l]{NICE-\\SLAM~\cite{zhu2022nice}}}
& Depth L1 [cm] $\downarrow$ & 1.81  & \rd 1.44	& \rd 2.04	& \rd 1.39	& \rd 1.76	&8.33	&4.99	&2.01	&2.97 \\
& Precision [$\%$] $\uparrow$ & \rd 45.86	&\nd 43.76	&\rd 44.38	&\nd 51.40	&\nd 50.80	&\rd 38.37	&\rd 40.85	&\rd 37.35	&\rd 44.10 \\
& Recall [$\%$] $\uparrow$  &\rd 44.10	&\nd 46.12	&\rd 42.78	&\nd 48.66	&\nd 53.08	&\rd 39.98	&\rd 39.04	&\rd 35.77	&\rd 43.69\\
& F1 [$\%$] $\uparrow$ &\rd 44.96	& \nd 44.84	&\rd 43.56	&\nd 49.99	&\nd 51.91	&\rd 39.16	&\rd 39.92	&\rd 36.54	&\rd 43.86\\[0.8pt] \hdashline \noalign{\vskip 1pt}
\multirow{4}{*}{\makecell[l]{Vox-\\Fusion$^{*}$~\cite{yang2022vox}}} 
& Depth L1 [cm] $\downarrow$ & \rd 1.09 & 1.90 & 2.21  & 2.32 & 3.40  & \rd 4.19  & \rd 2.96 & \rd 1.61 & \rd 2.46\\
 &  Precision [$\%$] $\uparrow$ & \nd 75.83 & \rd 35.88 &  \nd 63.10 & \rd 48.51 & \rd 43.50  &  \nd 54.48 & \nd 69.11 & \nd 55.40 & \nd 55.73\\
 & Recall [$\%$] $\uparrow$ & \nd 64.89 & \rd 33.07 & \nd 56.62  & \rd 44.76  & \rd 38.44  &  \nd 47.85 & \nd 60.61  & \nd 46.79 & \nd 49.13\\
   & F1 [$\%$] $\uparrow$  & \nd 69.93 & \rd 34.38 &  \nd 59.67 & \rd 46.54 & \rd  40.81 &  \nd 50.95 & \nd 64.56 & \nd 50.72 & \nd 52.20\\[0.8pt] \hdashline \noalign{\vskip 1pt}
\multirow{1}{*}{ESLAM~\cite{mahdi2022eslam}} & Depth L1 [cm] $\downarrow$ & \nd 0.97  & \nd 1.07  & \nd 1.28  & \nd 0.86 & \nd 1.26  & \nd 1.71  & \nd 1.43  & \nd 1.06 & \nd 1.18 \\[0.8pt] \hdashline \noalign{\vskip 1pt}
\multirow{4}{*}{Ours} 
& Depth L1 [cm] $\downarrow$ &\textbf{0.53}\fs  & \textbf{0.22} \fs  & \textbf{0.46}\fs   & \fs \textbf{0.30}  &  \fs \textbf{0.57}  &  \fs \textbf{0.49} & \fs\textbf{0.51} & \fs\textbf{0.46} & \fs\textbf{0.44} \\
& Precision [$\%$] $\uparrow$ &\textbf{91.95} \fs  &\textbf{99.04} \fs  &\textbf{97.89} \fs   & \fs \textbf{99.00} &  \fs \textbf{99.37} &  \fs \textbf{98.05} & \fs \textbf{96.61} & \fs \textbf{93.98}& \fs\textbf{96.99} \\
& Recall [$\%$] $\uparrow$&\textbf{82.48} \fs  &\textbf{86.43} \fs  &\textbf{84.64} \fs & \fs \textbf{89.06}  & \fs \textbf{84.99} &  \fs \textbf{81.44} &  \fs \textbf{81.17} & \fs\textbf{78.51}  & \fs \textbf{83.59} \\
& F1 [$\%$] $\uparrow$  &\textbf{86.90} \fs  &\textbf{92.31} \fs  &\textbf{90.78} \fs   & \fs \textbf{93.77}  &  \fs \textbf{91.62}  &  \fs  \textbf{88.98} & \fs \textbf{88.22}  & \fs \textbf{85.55} & \fs \textbf{89.77} \\
\bottomrule
\end{tabularx}
\subcaption{}
\label{tab:replica_recon}
\end{subfigure}
}
&
\resizebox{0.53\linewidth}{!}
{
\begin{subfigure}{0.64\linewidth}
\centering
{\footnotesize
\setlength{\tabcolsep}{1pt}
\newcommand{\sz}{0.24}
\renewcommand{\arraystretch}{1}
\begin{tabular}{ccccc}
\rotatebox[origin=c]{90}{\texttt{Office 0}} & 
\includegraphics[valign=c,width=\sz\linewidth]{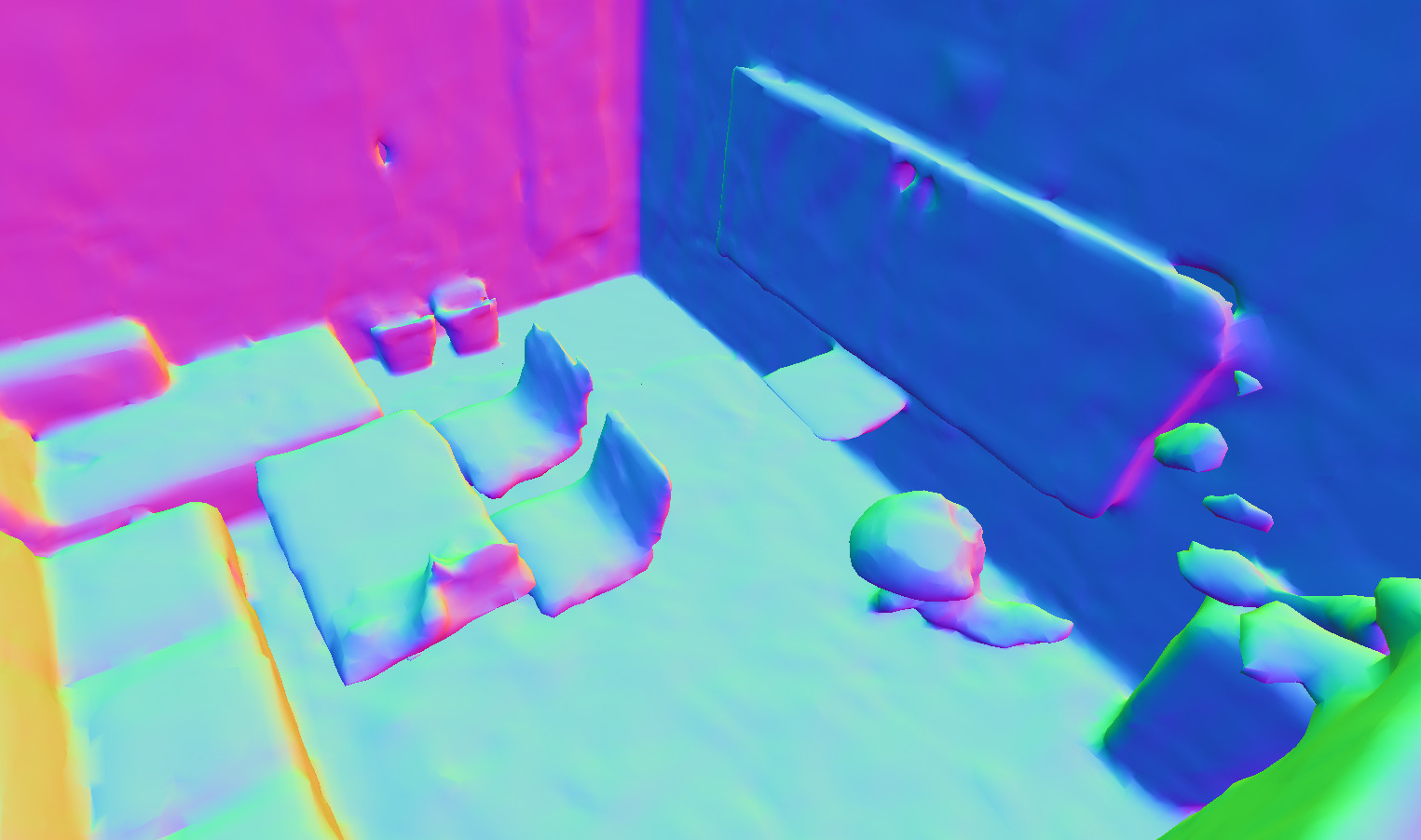} & 
\includegraphics[valign=c,width=\sz\linewidth]{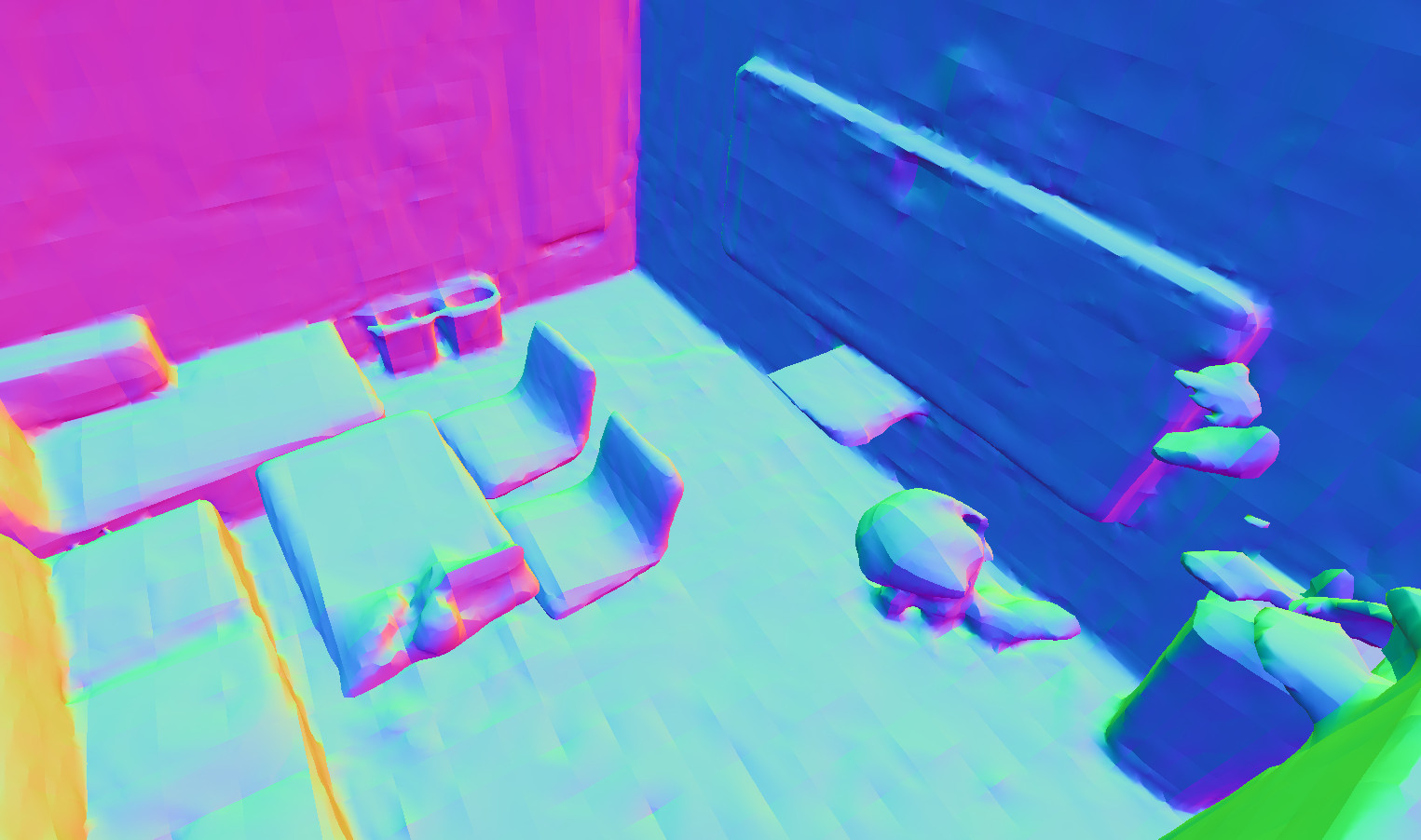} &
\includegraphics[valign=c,width=\sz\linewidth]{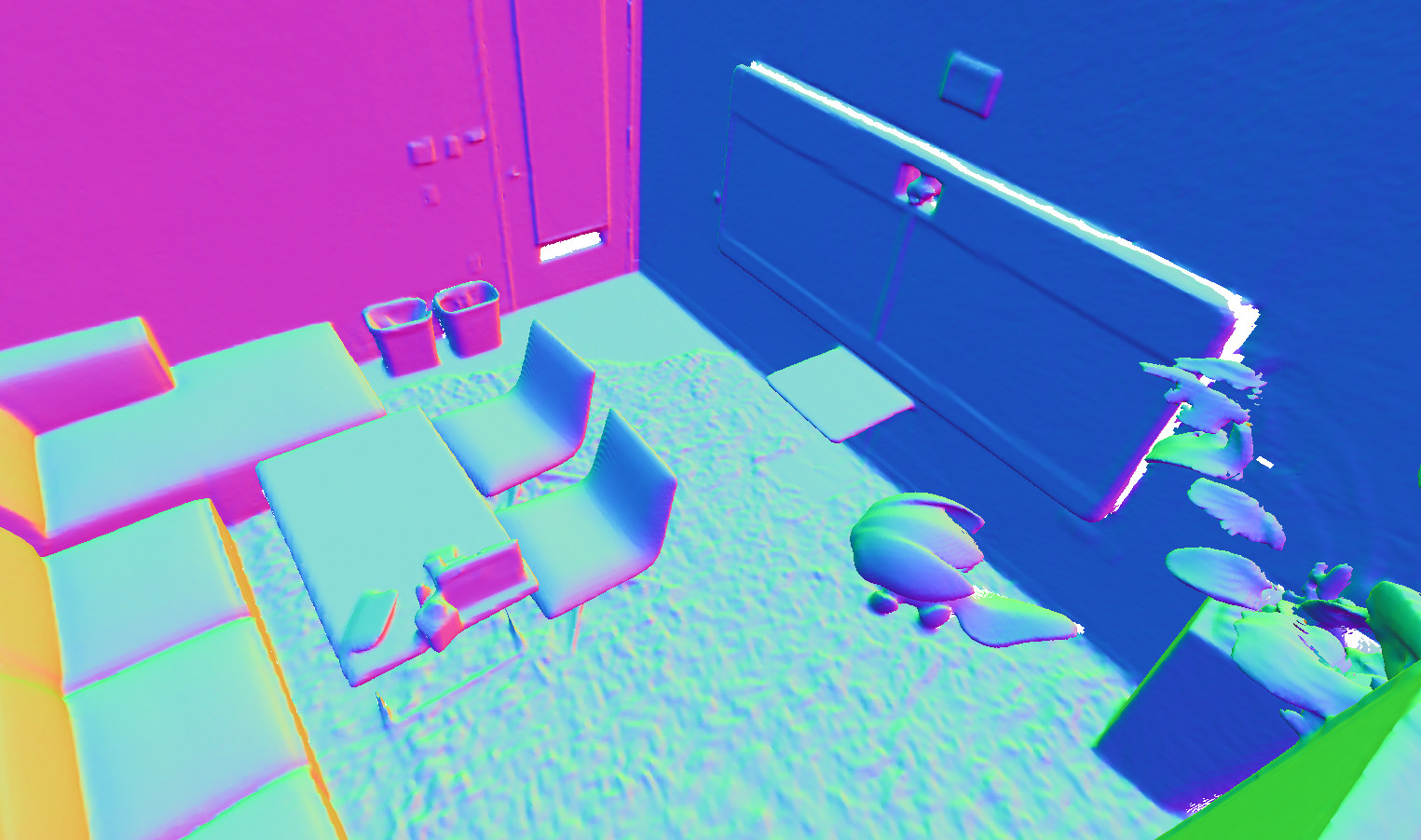} &
\includegraphics[valign=c,width=\sz\linewidth]{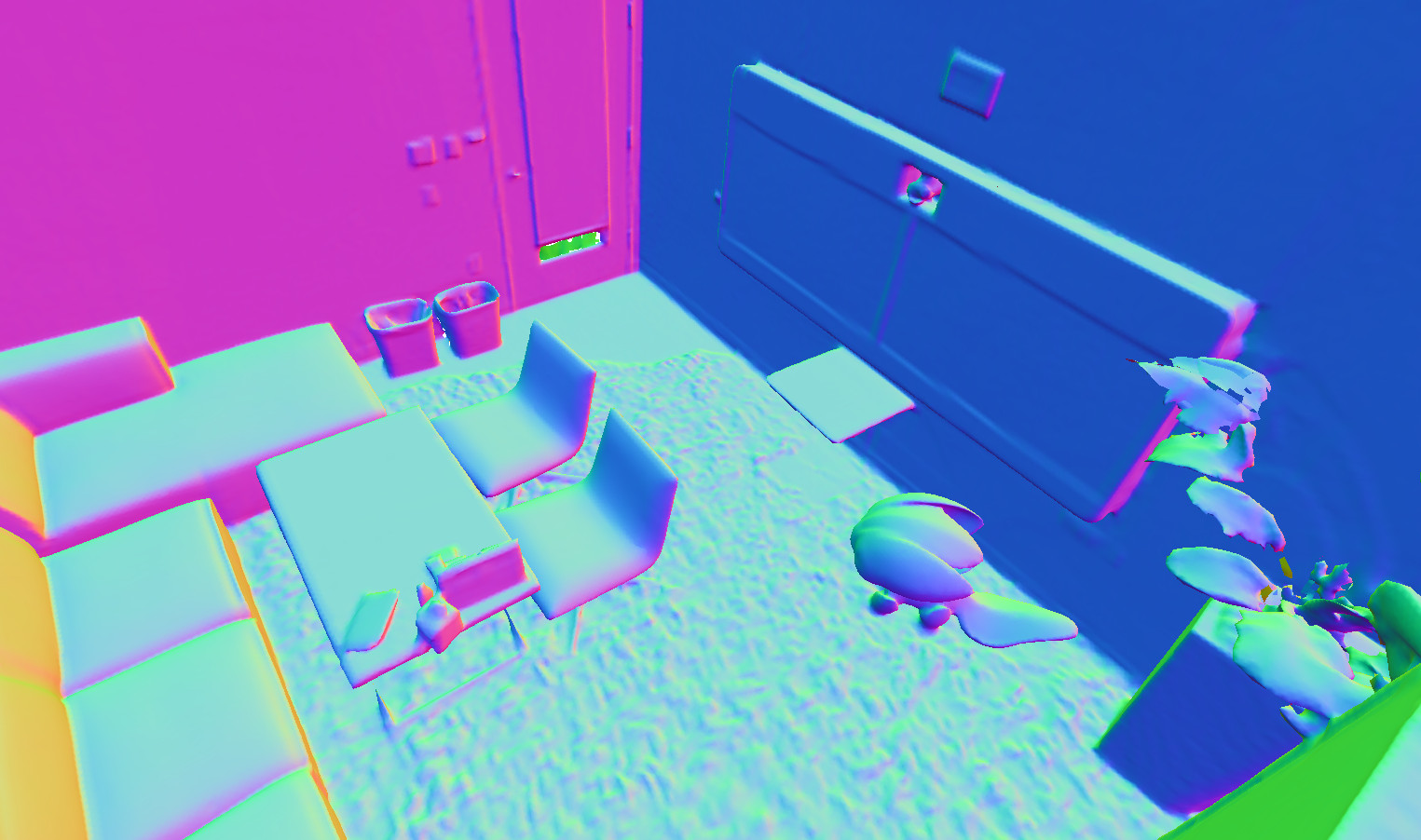} \\
\rotatebox[origin=c]{90}{\texttt{Office 3}} & 
\includegraphics[valign=c,width=\sz\linewidth]{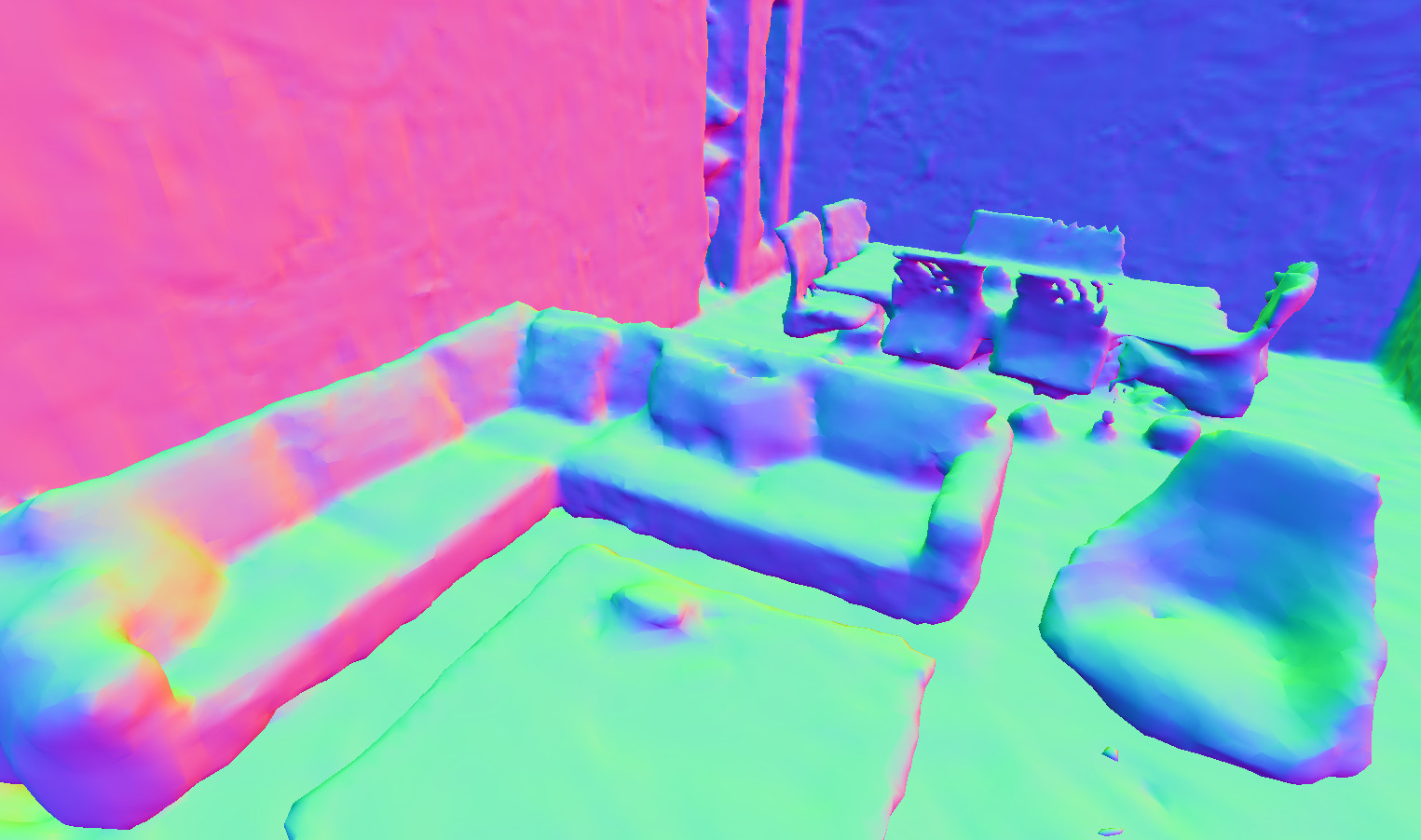} & 
\includegraphics[valign=c,width=\sz\linewidth]{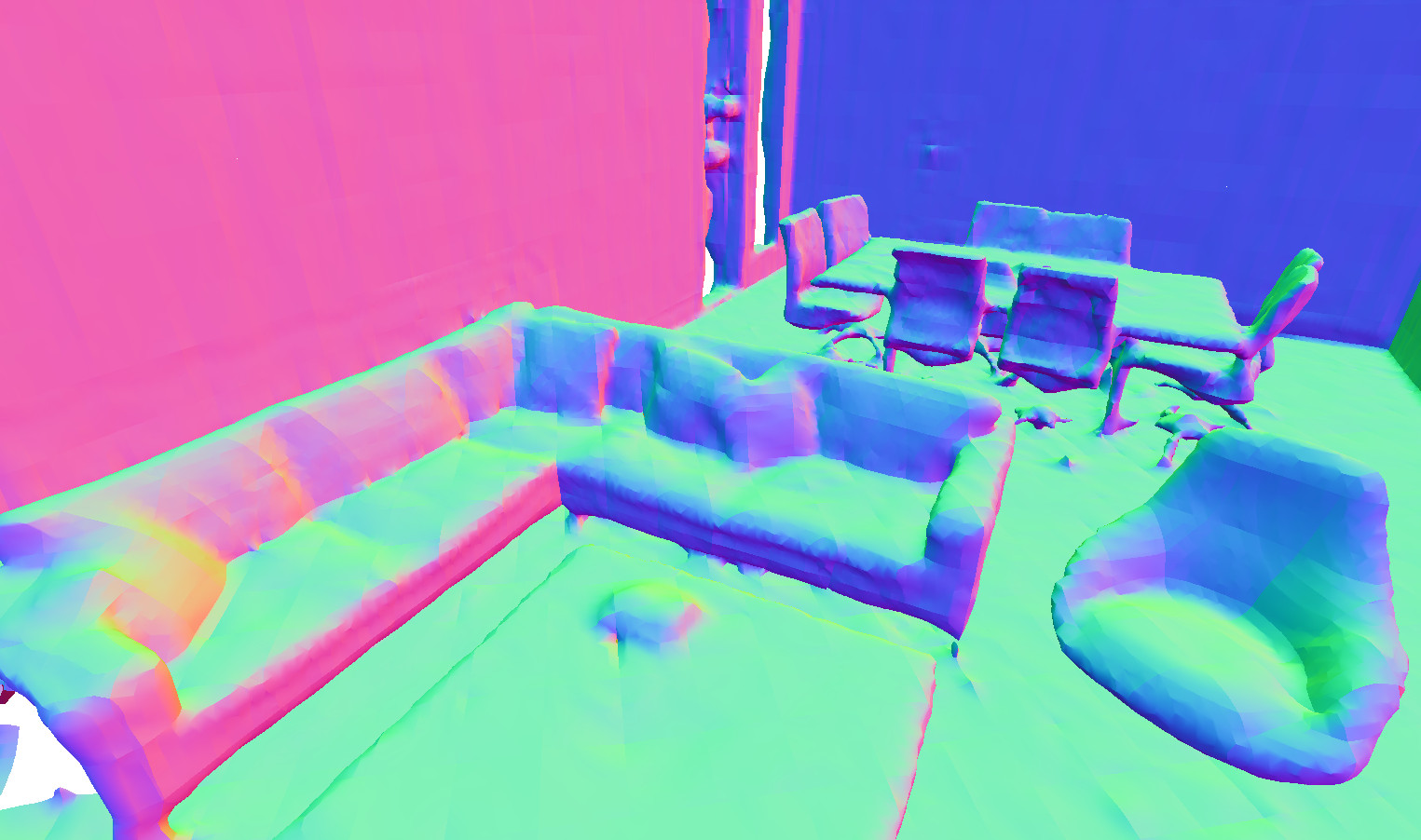} &
\includegraphics[valign=c,width=\sz\linewidth]{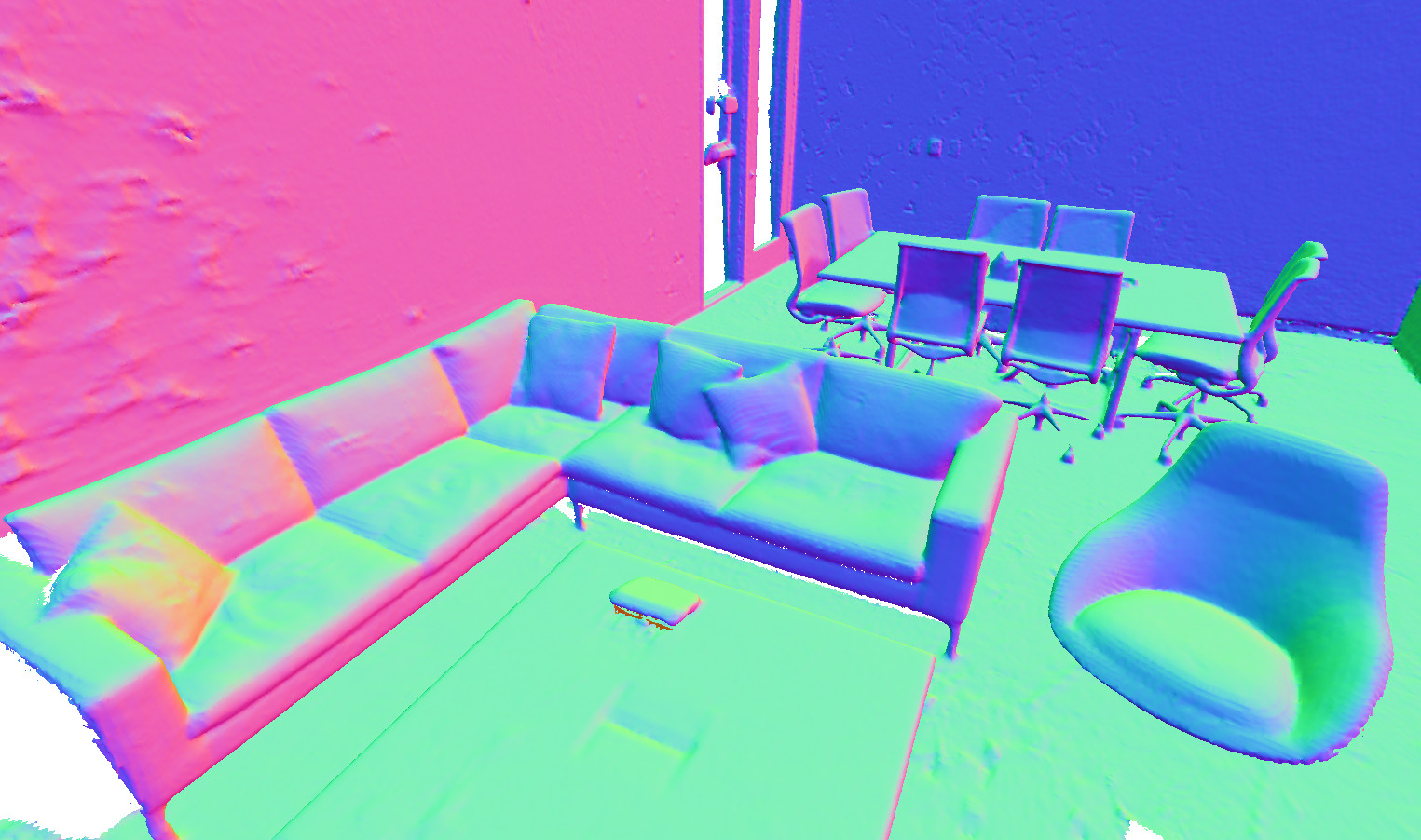} &
\includegraphics[valign=c,width=\sz\linewidth]{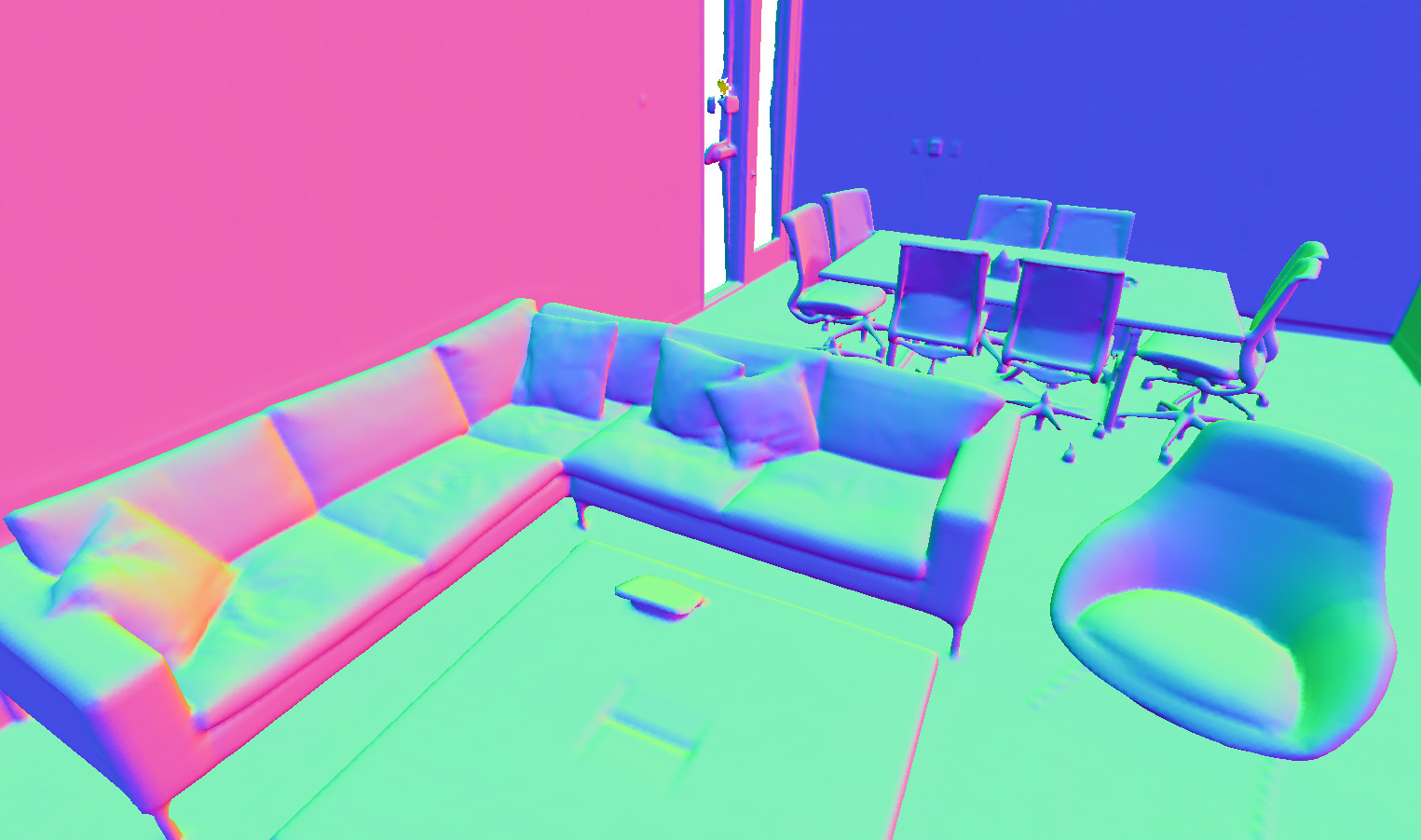} \\
\rotatebox[origin=c]{90}{\texttt{Room 0}} & 
\includegraphics[valign=c,width=\sz\linewidth]{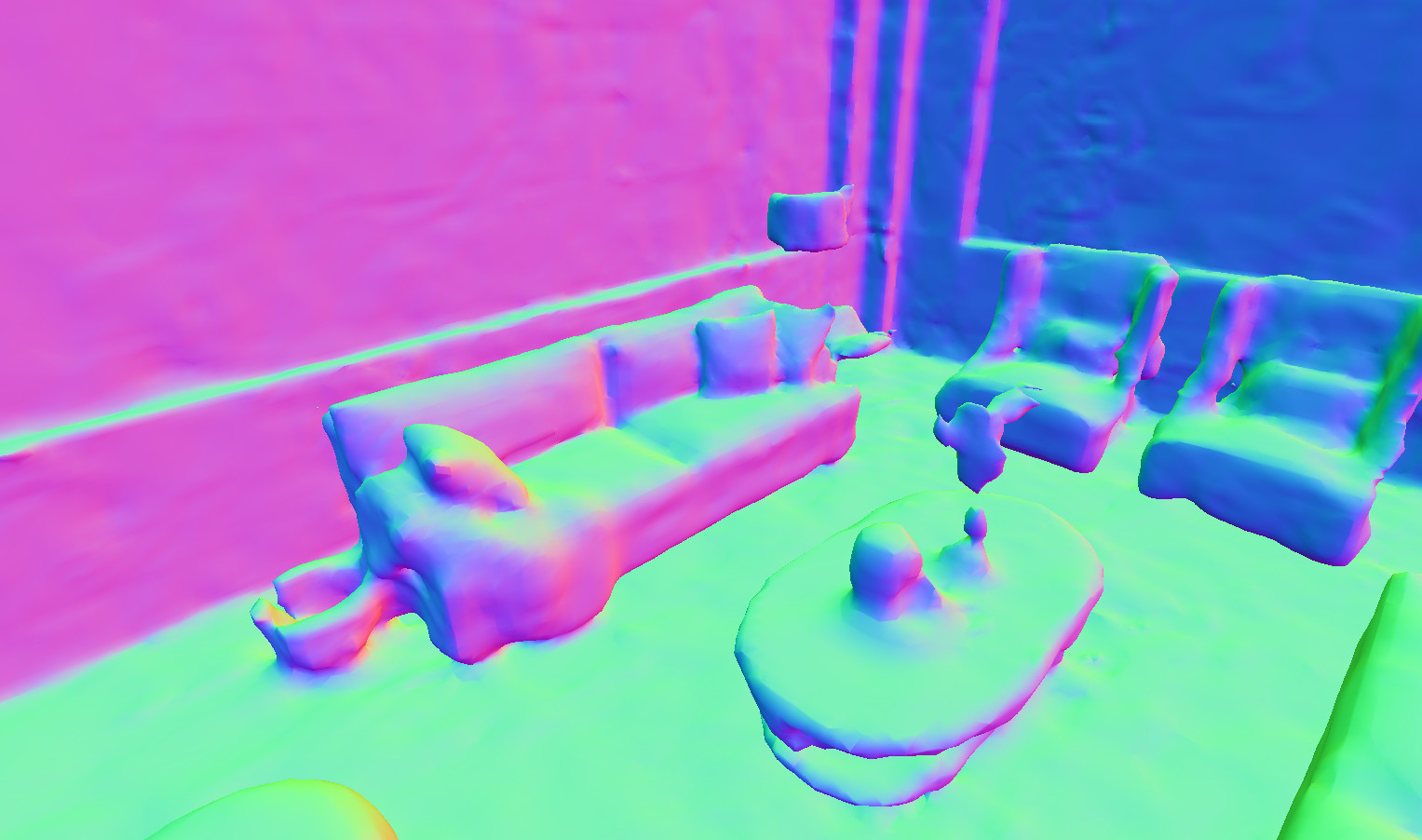} & 
\includegraphics[valign=c,width=\sz\linewidth]{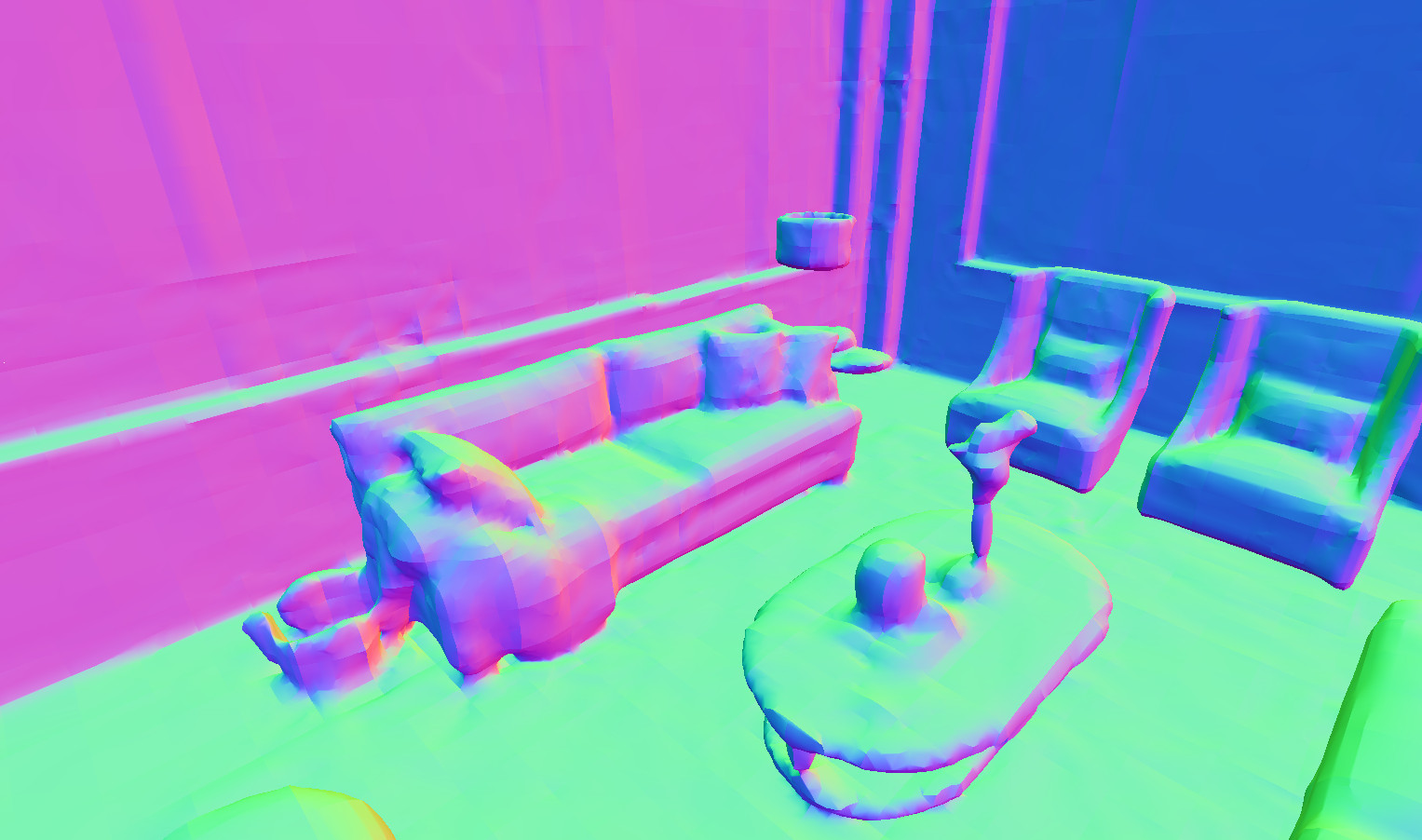} &
\includegraphics[valign=c,width=\sz\linewidth]{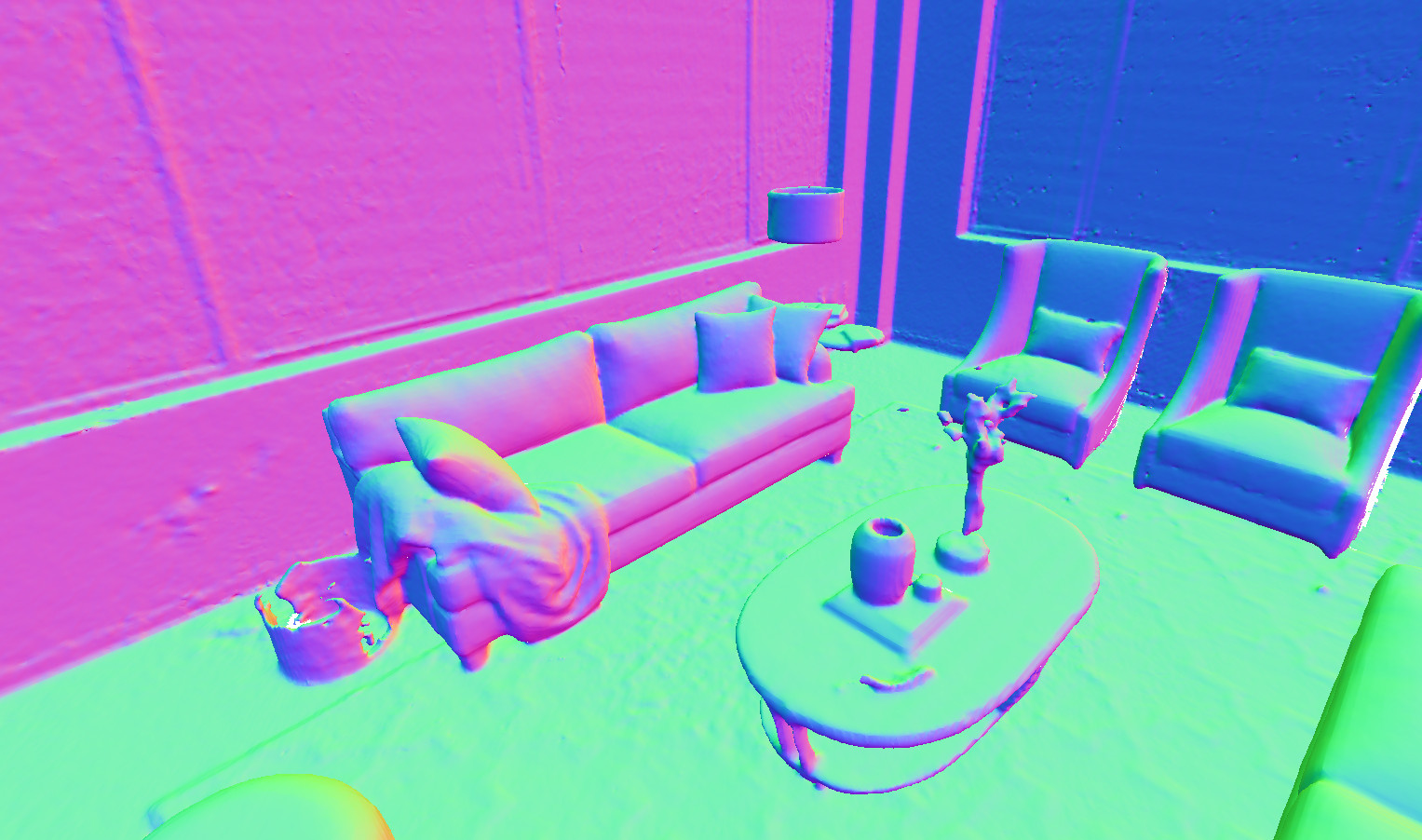} &
\includegraphics[valign=c,width=\sz\linewidth]{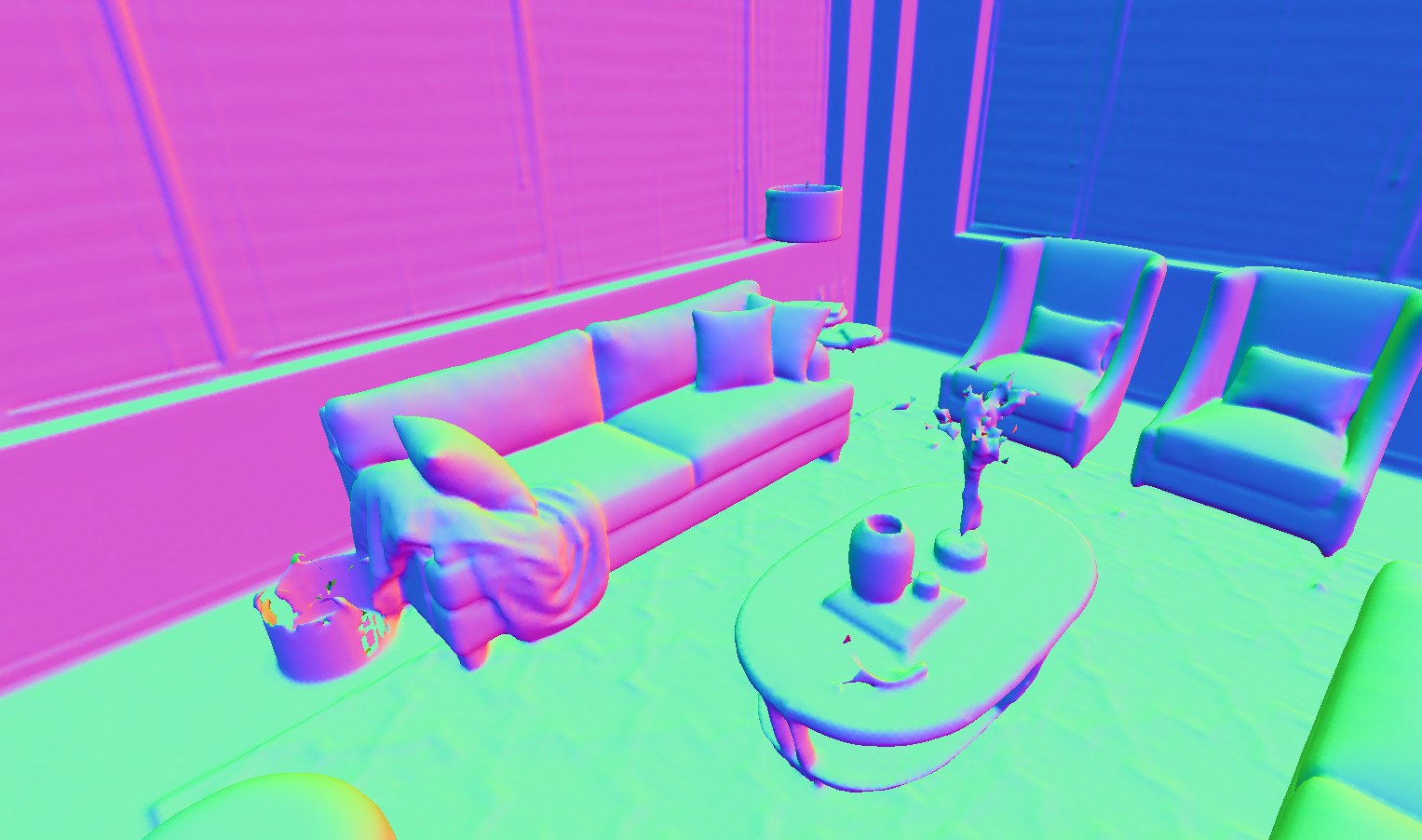} \\
 & NICE-SLAM~\cite{zhu2022nice} & Vox-Fusion$^*$~\cite{yang2022vox} & \ours (ours) & Ground Truth \\
\end{tabular}
}
\subcaption{}
\label{fig:replica_recon}
\end{subfigure}
}
\end{tabular}
}
\vspace{2mm}
\caption{\textbf{Reconstruction Performance on Replica~\cite{straub2019replica}.} \cref{tab:replica_recon}: Our method is able to outperform all existing methods. Best results are highlighted as \colorbox{colorFst}{\bf first}, \colorbox{colorSnd}{second}, and \colorbox{colorTrd}{third}. \cref{fig:replica_recon}: \ours yields on average more precise reconstructions than existing methods, e.g. note the fidelity of the rough carpet reconstruction on \texttt{Office 0}.}
\label{fig:replica_recon_full}
\end{figure*}

\boldparagraph{Evaluation Metrics.}
 The meshes, produced by marching cubes~\cite{lorensen1987marching}, are evaluated using the F-score which is the harmonic mean of the Precision (P) and Recall (R). We use a distance threshold of $1$ cm for all evaluations. We further provide the depth L1 metric as in~\cite{zhu2022nice}. For tracking accuracy, we use ATE RMSE~\cite{Sturm2012ASystems} and for rendering we provide the peak signal-to-noise ratio (PSNR), SSIM~\cite{wang2004image} and LPIPS~\cite{zhang2018unreasonable}. Our rendering metrics are evaluated by rendering the full resolution image along the estimated trajectory every 5th frame. Unless otherwise written, we report the average metric of three runs on seeds 0, 1 and 2.

\boldparagraph{Datasets.} 
The Replica dataset~\cite{straub2019replica} comprises high-quality 3D reconstructions of a variety of indoor scenes. 
We utilize the publicly available dataset collected by Sucar~\etal~\cite{Sucar2021IMAP:Real-Time}, which provides trajectories from an RGBD sensor.
Further, we demonstrate that our framework can handle real-world data by using the TUM-RGBD dataset~\cite{Sturm2012ASystems}, as well as the ScanNet dataset~\cite{Dai2017ScanNet}. 
The poses for TUM-RGBD were captured using an external motion capture system while ScanNet uses poses from BundleFusion~\cite{dai2017bundlefusion}. 

\boldparagraph{Baseline Methods.} 
We primarily compare our method to existing state-of-the-art dense neural RGBD SLAM methods such as NICE-SLAM~\cite{zhu2022nice}, Vox-Fusion~\cite{yang2022vox} and ESLAM~\cite{mahdi2022eslam}. 
We reproduce the results from~\cite{yang2022vox} using the open source code and report the results as Vox-Fusion$^*$. 
For NICE-SLAM, we use $40$ tracking iterations on Replica and mesh the scene at resolution $1 cm$ for a fair comparison.

\begin{table}[tb]
\centering
\setlength{\tabcolsep}{2pt}
\renewcommand{\arraystretch}{1.05}
\resizebox{\columnwidth}{!}
{
\begin{tabular}{lccccccccc}
\toprule
Method & \texttt{Rm 0} & \texttt{Rm 1} & \texttt{Rm 2} & \texttt{Off 0} & \texttt{Off 1} & \texttt{Off 2} & \texttt{Off 3} & \texttt{Off 4} & Avg.\\
\midrule
\multirow{1}{*}{NICE-SLAM~\cite{zhu2022nice}} & \rd 0.97
& \rd 1.31 & \rd 1.07  & \rd 0.88 & \rd 1.00  & \rd 1.06  & \rd 1.10  &\rd 1.13 & \rd 1.06 \\[0.8pt] \noalign{\vskip 1pt}
\lo \multirow{1}{*}{Vox-Fusion~\cite{yang2022vox}} & \lo 0.40 & \lo 0.54 & \lo 0.54 & \lo 0.50 & \lo 0.46 & \lo 0.75 & \lo0.50 & \lo 0.60 & \lo 0.54\\[0.8pt] \noalign{\vskip 1pt}
\multirow{1}{*}{Vox-Fusion$^{*}$~\cite{yang2022vox}} & 1.37 & 4.70 &  1.47 & 8.48 & 2.04  &  2.58 & 1.11 & 2.94 & 3.09\\[0.8pt]  \noalign{\vskip 1pt}
\multirow{1}{*}{ESLAM~\cite{mahdi2022eslam}} & \nd 0.71 & \nd 0.70 & \nd 0.52  & \nd 0.57  & \nd 0.55  & \nd 0.58  & \nd 0.72 & \fs \textbf{0.63} & \nd 0.63 \\[0.8pt]  \noalign{\vskip 1pt}
\multirow{1}{*}{\ours (ours)
} 
& \fs \textbf{0.61}  & \fs \textbf{0.41}  & \fs \textbf{0.37}   & \fs \textbf{0.38} &  \fs 0.48 & \fs \textbf{0.54}  & \fs \textbf{0.69} & \nd 0.72 & \fs \textbf{0.52} \\
\bottomrule
\end{tabular}
}
\caption{\textbf{Tracking Performance on Replica~\cite{straub2019replica}} (ATE RMSE $\downarrow$ [cm]). On average, we achieve better tracking than existing methods. The grayed numbers of \cite{yang2022vox} are from the paper that come from a single run which we could not reproduce. We report an average of $3$ runs for all other methods in this table. Vox-Fusion$^*$ indicates recreated results.}
\label{tab:replica_tracking}
\end{table}

\begin{table*}[t]
\centering
\footnotesize
\setlength{\tabcolsep}{4.5pt}
\begin{tabularx}{\linewidth}{llccccccccc}
\toprule
Method & Metric & \texttt{Room 0} & \texttt{Room 1} & \texttt{Room 2} & \texttt{Office 0} & \texttt{Office 1} & \texttt{Office 2} & \texttt{Office 3} & \texttt{Office 4} & Avg.\\
\midrule
\multirow{3}{*}{NICE-SLAM~\cite{zhu2022nice}}
& PSNR [dB] $\uparrow$ & \rd 22.12 & \nd 22.47 & \nd 24.52 & \nd 29.07 & \nd 30.34 & \rd 19.66 & \rd 22.23 & \rd 24.94 & \nd 24.42 \\
& SSIM $\uparrow$ & \nd 0.689 & \nd 0.757 & \nd 0.814 & \nd 0.874 & \nd 0.886 & \nd 0.797 & \rd 0.801 & \nd 0.856 & \nd 0.809 \\
& LPIPS $\downarrow$ & \rd 0.330 & \rd 0.271 & \nd 0.208 & \nd 0.229 & \nd 0.181 & \nd 0.235 & \nd 0.209 & \nd 0.198 & \nd 0.233\\[0.8pt] \hdashline \noalign{\vskip 1pt}
\multirow{3}{*}{Vox-Fusion$^{*}$~\cite{yang2022vox}} & PSNR [dB] $\uparrow$ & \nd 22.39 & \rd 22.36 & \rd 23.92 & \rd 27.79 & \rd 29.83 & \nd 20.33 & \nd 23.47 & \nd 25.21 & \rd 24.41 \\
& SSIM $\uparrow$ & \rd 0.683 & \rd 0.751 & \rd 0.798 & \rd 0.857 & \rd 0.876 & \rd 0.794 & \nd 0.803 & \rd 0.847 & \rd 0.801\\
& LPIPS $\downarrow$ & \nd 0.303 & \nd 0.269 & \rd 0.234 & \rd 0.241 & \rd 0.184 & \rd 0.243 & \rd 0.213 & \rd 0.199 & \rd 0.236\\[0.8pt] \hdashline \noalign{\vskip 1pt}
\multirow{3}{*}{Ours} 
& PSNR [dB] $\uparrow$  &\fs \textbf{32.40} &\fs \textbf{34.08} &\fs \textbf{35.50}  & \fs \textbf{38.26} & \fs \textbf{39.16} & \fs \textbf{33.99} & \fs \textbf{33.48} & \fs \textbf{33.49} & \fs \textbf{35.17}\\
& SSIM $\uparrow$ &\fs \textbf{0.974} &\fs \textbf{0.977} &\fs \textbf{0.982}	& \fs \textbf{0.983} &	\fs \textbf{0.986} & \fs \textbf{0.960} & \fs \textbf{0.960} & \fs \textbf{0.979} & \fs \textbf{0.975} \\
& LPIPS $\downarrow$ & \fs \textbf{0.113} &\fs \textbf{0.116} &\fs \textbf{0.111} & \fs \textbf{0.100} &	\fs \textbf{0.118} & \fs \textbf{0.156} & \fs \textbf{0.132} & \fs \textbf{0.142} & \fs \textbf{0.124} \\
\bottomrule
\end{tabularx}
\caption{\textbf{Rendering Performance on Replica~\cite{straub2019replica}.} We outperform existing dense neural RGBD methods on the commonly reported rendering metrics. For NICE-SLAM~\cite{zhu2022nice} and Vox-Fusion~\cite{yang2022vox} we take the numbers from~\cite{zhu2023nicer}. For qualitative results, see \cref{fig:replica_rendering}.}
\label{tab:replica_rendering}
\end{table*}

\begin{figure*}[t]
\centering
{\footnotesize
\setlength{\tabcolsep}{1pt}
\renewcommand{\arraystretch}{1}
\newcommand{\sz}{0.23}
\begin{tabular}{ccccc}
\rotatebox[origin=c]{90}{\texttt{Office 0}} & 
\includegraphics[valign=c,width=\sz\linewidth]{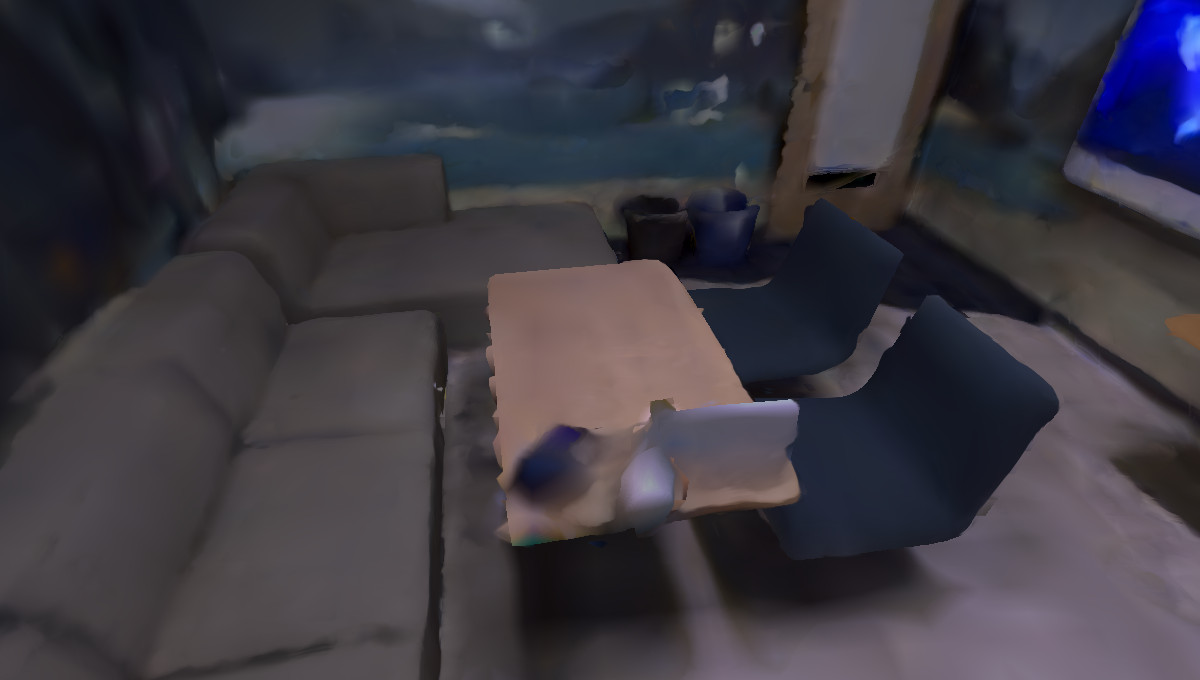} & 
\includegraphics[valign=c,width=\sz\linewidth]{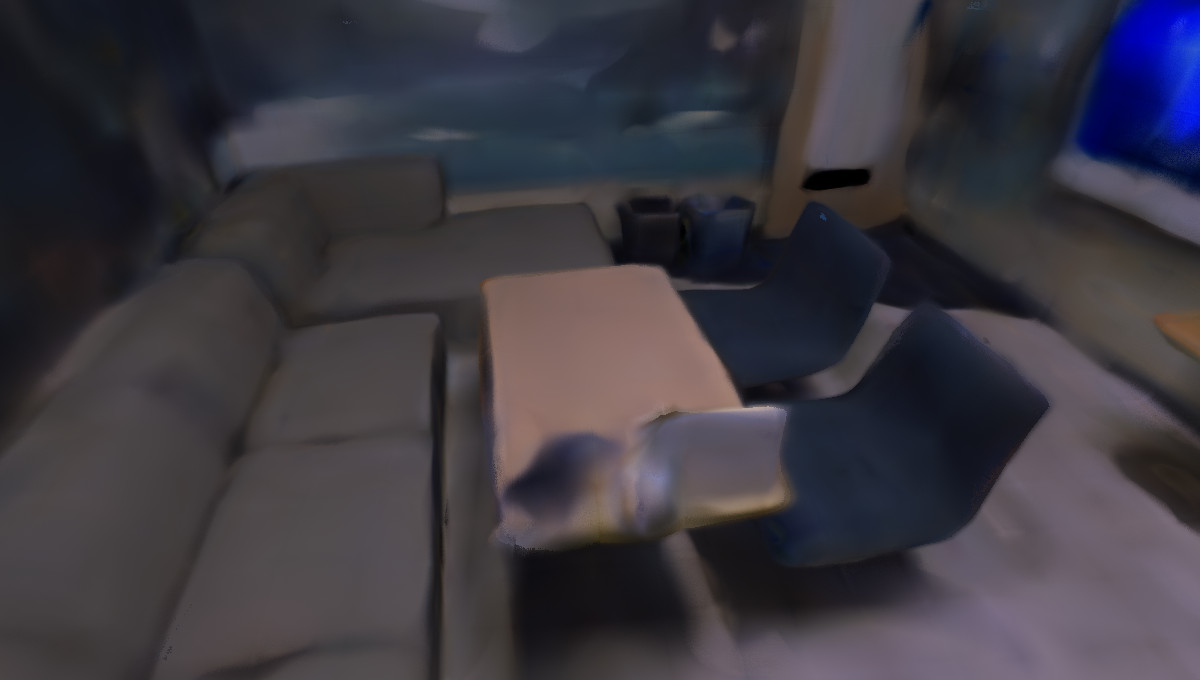} &
\includegraphics[valign=c,width=\sz\linewidth]{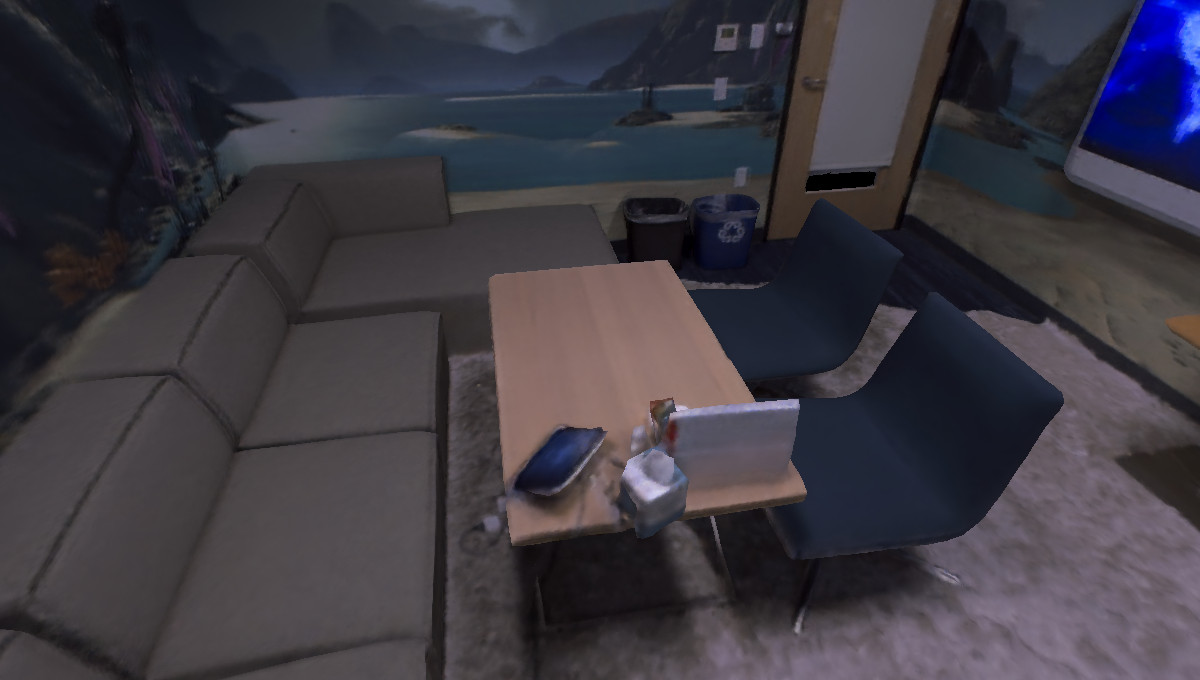} &
\includegraphics[valign=c,width=\sz\linewidth]{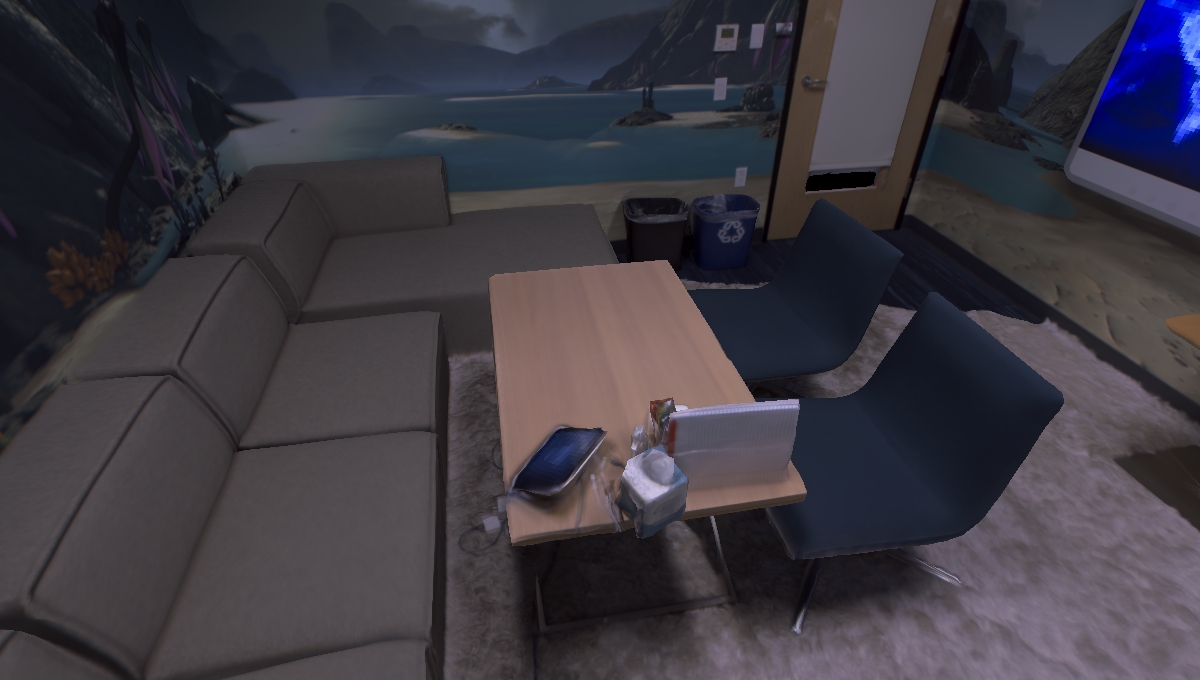} \\
\rotatebox[origin=c]{90}{\texttt{Room 1}} & 
\includegraphics[valign=c,width=\sz\linewidth]{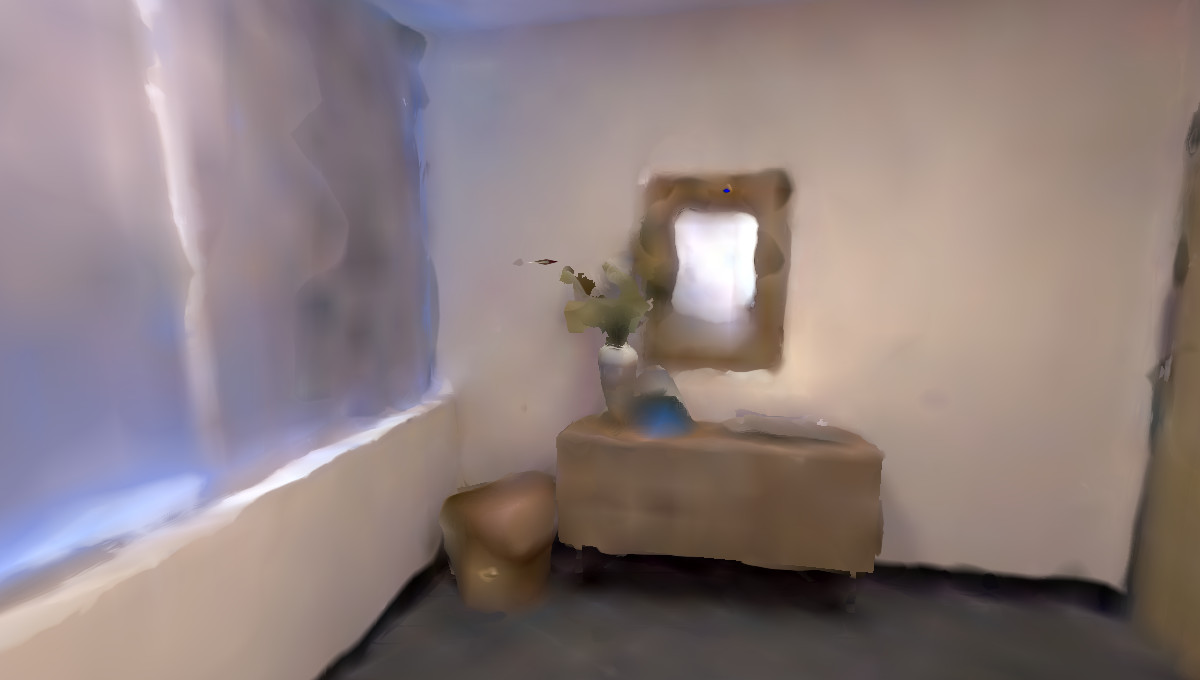} & 
\includegraphics[valign=c,width=\sz\linewidth]{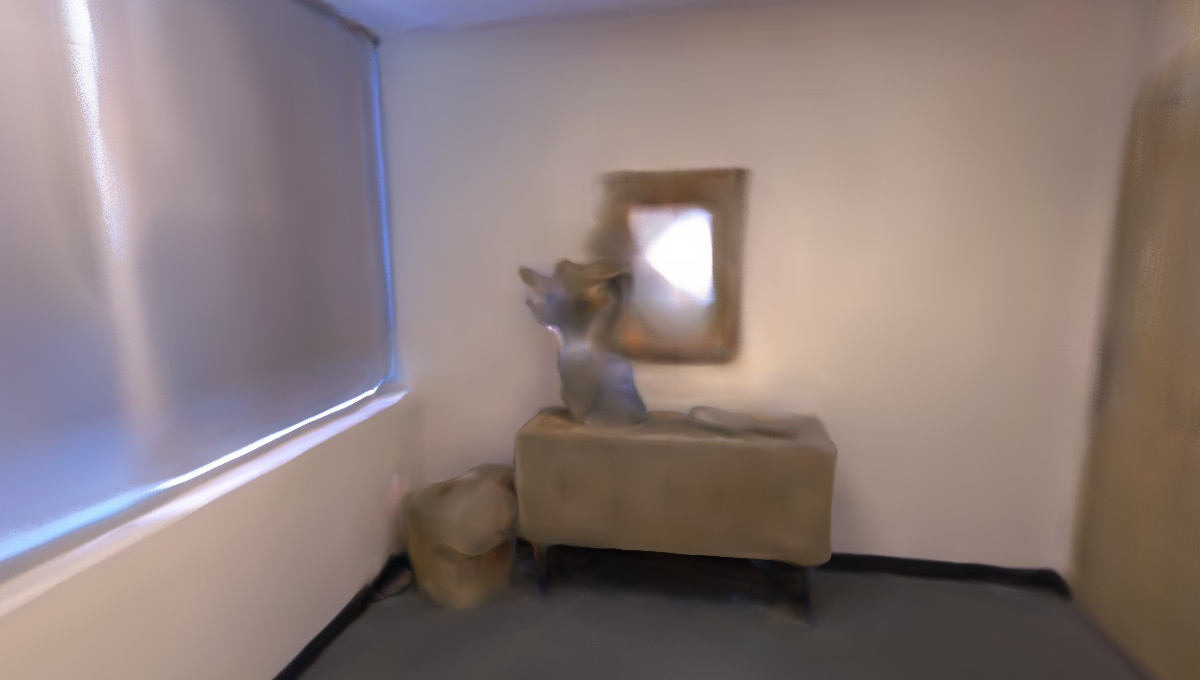} &
\includegraphics[valign=c,width=\sz\linewidth]{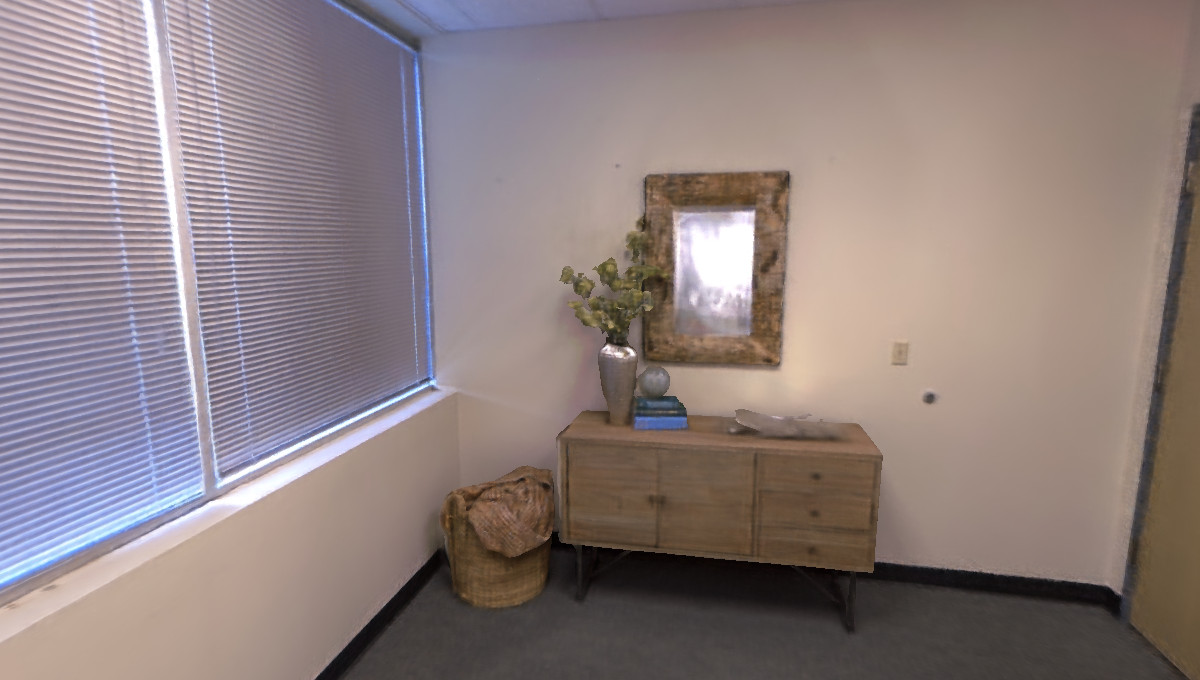} &
\includegraphics[valign=c,width=\sz\linewidth]{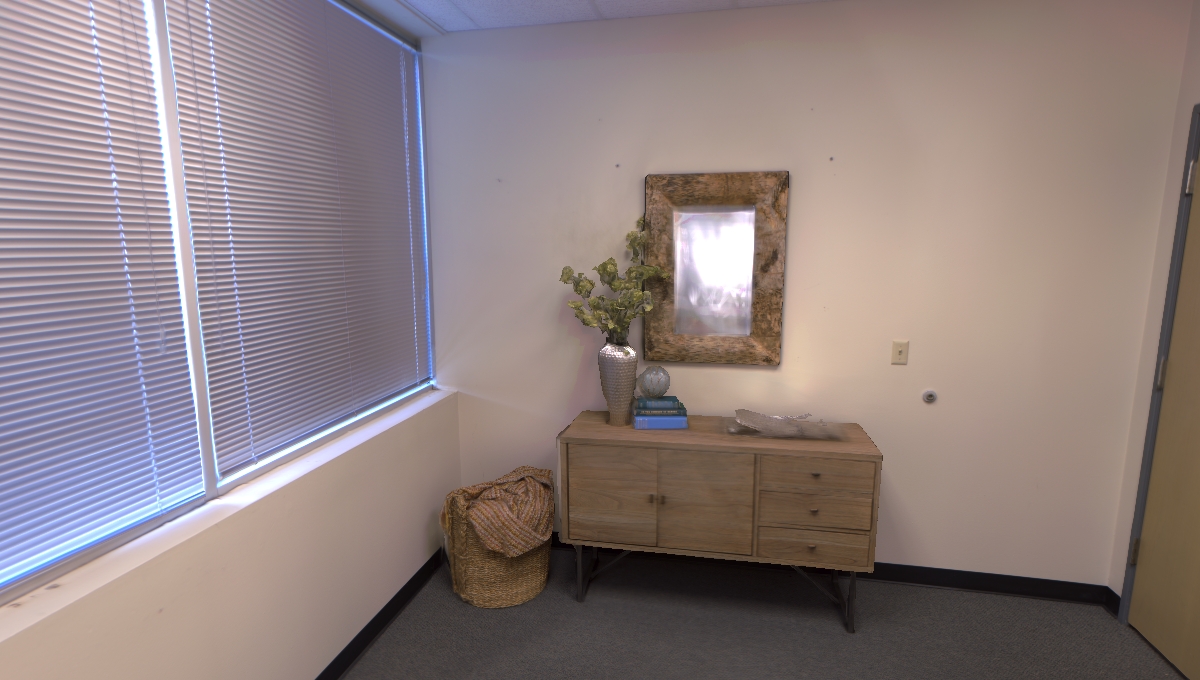} \\
\rotatebox[origin=c]{90}{\texttt{Room 2}} & 
\includegraphics[valign=c,width=\sz\linewidth]{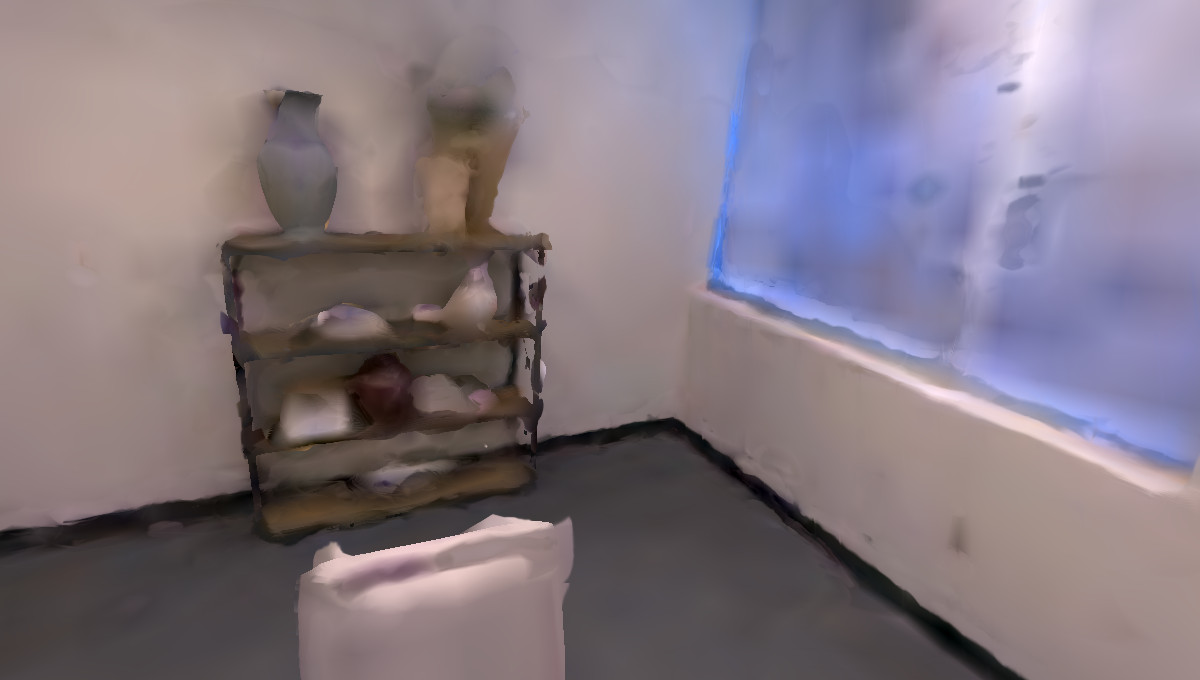} & 
\includegraphics[valign=c,width=\sz\linewidth]{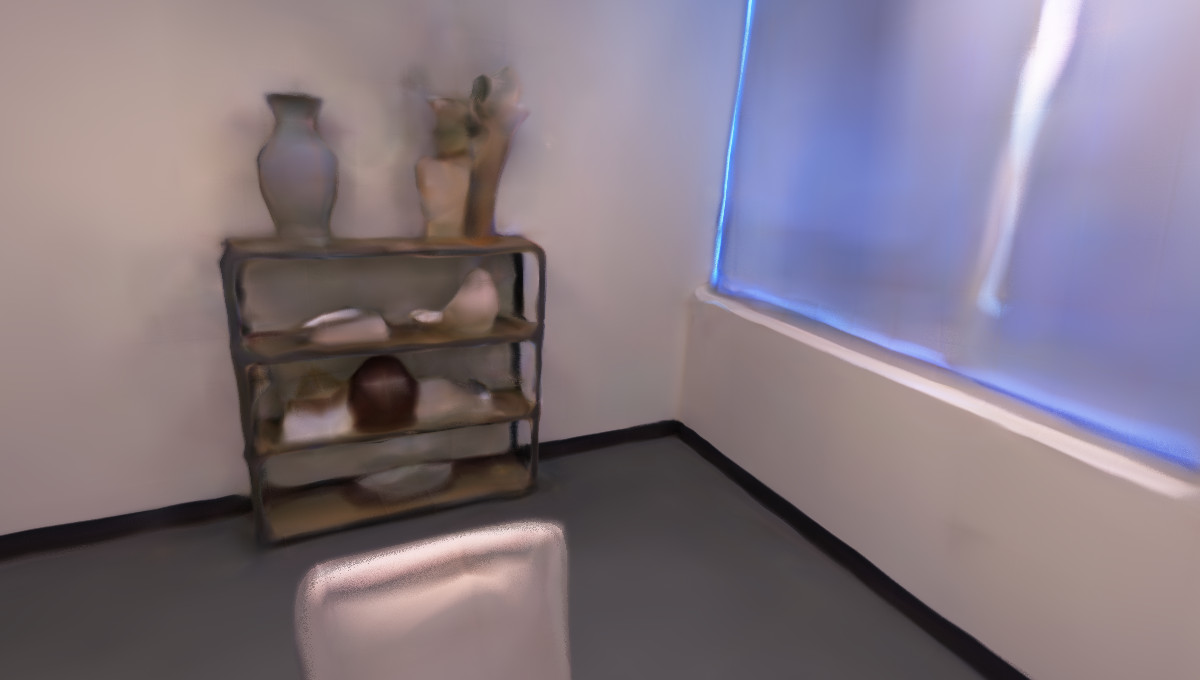} &
\includegraphics[valign=c,width=\sz\linewidth]{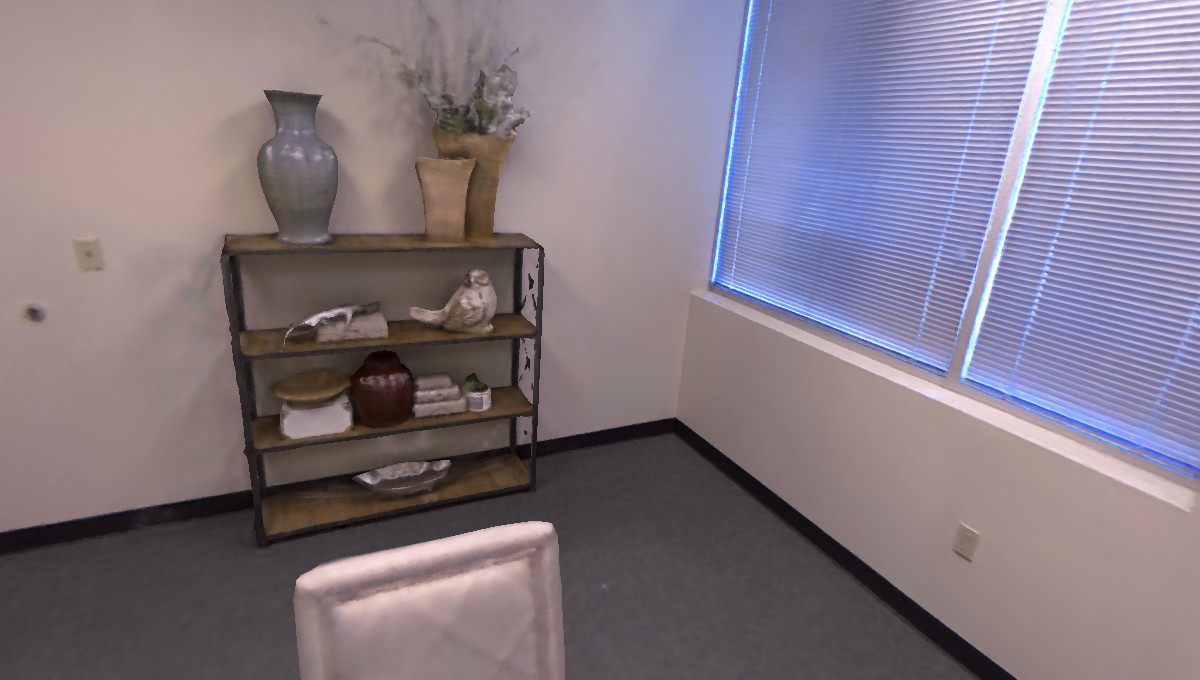} &
\includegraphics[valign=c,width=\sz\linewidth]{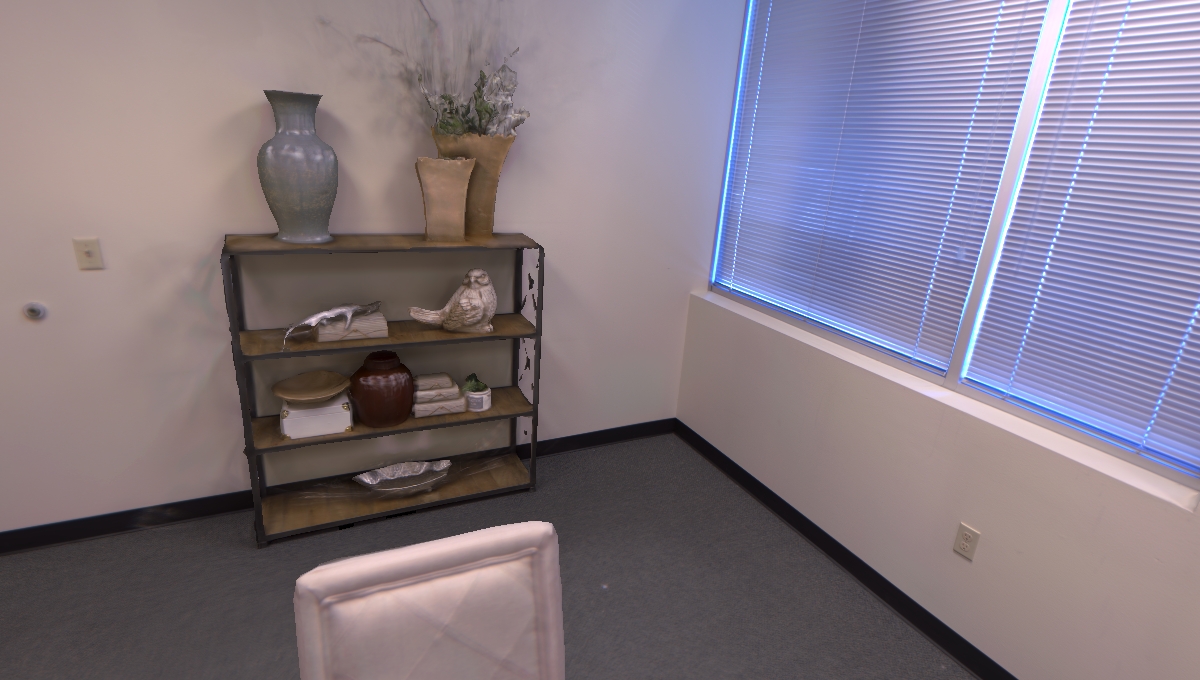} \\
 & NICE-SLAM~\cite{zhu2022nice} & Vox-Fusion$^{*}$~\cite{yang2022vox} & \ours (ours) & Ground Truth \\
\end{tabular}
}
\caption{\textbf{Rendering Performance on Replica~\cite{straub2019replica}}. Thanks to the adaptive density of the neural point cloud, \ours{} is able encode more high-frequency details and to substantially increase the fidelity of the renderings. This is also supported by the quantitative results in \cref{tab:replica_rendering}. }
\label{fig:replica_rendering}
\end{figure*}

\begin{table}[tb]
\centering
\setlength{\tabcolsep}{2pt}
\renewcommand{\arraystretch}{1.05}
\resizebox{\columnwidth}{!}
{
\begin{tabular}{lllllll}
\toprule
   \multirow{2}{*}{Method}& \texttt{fr1/} &  \texttt{fr1/} & \texttt{fr1/} & \texttt{fr2/} & \texttt{fr3/} & \multirow{2}{*}{Avg.} \\
  & \texttt{desk} &  \texttt{desk2} & \texttt{room} & \texttt{xyz} & \texttt{office} &  \\ 
\midrule
DI-Fusion~\cite{huang2021di} & 4.4  &  N/A & N/A & 2.0 & 5.8 &  N/A\\
NICE-SLAM~\cite{zhu2022nice} &4.26  & \rd 4.99  &34.49  &31.73 (6.19)  &3.87 &15.87 \makecell{ (10.76)} \\
Vox-Fusion$^*$~\cite{yang2022vox}  & 3.52  & 6.00  & \rd 19.53 & 1.49 &  26.01 &   11.31 \\
\textbf{\ours{} (Ours)} &4.34  & \nd 4.54  &  30.92  &1.31  &3.48 &  8.92 \\[0.8pt] \hdashline \noalign{\vskip 1pt}
BAD-SLAM~\cite{schops2019bad} & \nd 1.7 &  N/A &   N/A & \nd 1.1 & \nd 1.7 &  N/A\\
Kintinuous~\cite{whelan2012kintinuous} & 3.7  & 7.1 &  \nd 7.5  & 2.9 & 3.0 &  \nd 4.84\\
ORB-SLAM2~\cite{Mur-Artal2017ORB-SLAM2:Cameras} & \fs \textbf{1.6} &   \fs \textbf{2.2} & \fs \textbf{4.7} & \fs \textbf{0.4} & \fs \textbf{1.0} &  \fs 1.98\\
ElasticFusion~\cite{whelan2015elasticfusion} & \rd 2.53 & 6.83 & 21.49 & \rd 1.17 & \rd 2.52 & \rd 6.91\\\bottomrule
\end{tabular}
}
\caption{\textbf{Tracking Performance on TUM-RGBD~\cite{Sturm2012ASystems}} (ATE RMSE $\downarrow$ [cm]). \ours consistently outperforms existing dense neural RGBD methods (top part), and is reducing the gap to sparse tracking methods (bottom part). In parenthesis we report the average over only the successful runs.}
\label{tab:tum}
\end{table}

\begin{table}[tb]
\centering
\setlength{\tabcolsep}{2pt}
\renewcommand{\arraystretch}{1.05}
\resizebox{\columnwidth}{!}
{
\begin{tabular}{llllllll}
\toprule
Method & \texttt{0000} & \texttt{0059} & \texttt{0106} & \texttt{0169} & \texttt{0181} & \texttt{0207} & Avg. \\ \midrule
DI-Fusion~\cite{huang2021di} & 62.99 & 128.00 & 18.50 & 75.80 & 87.88 & 100.19 & 78.89 \\
NICE-SLAM~\cite{zhu2022nice} &\nd12.00 & \nd 14.00 &\fs7.90 &\nd10.90 & \fs \textbf{13.40} &\fs6.20 &\fs10.70 \\
\lo Vox-Fusion~\cite{yang2022vox} & \lo 8.39 & \lo N/A & \lo 7.44 & \lo 6.53 & \lo 12.20 & \lo 5.57 & \lo N/A \\
Vox-Fusion$^*$~\cite{yang2022vox} & 68.84 & \rd 24.18 &\nd 8.41 & \rd27.28 & \rd 23.30 & \rd9.41 & 26.90 \\
 & \rd (16.55) &  &  &  &  &  & \rd (18.52)\\
\textbf{\ours{} (Ours)} & \fs 10.24  &\fs 7.81    & \rd 8.65  & \nd 22.16 & \nd 14.77 & \nd 9.54 &\nd 12.19 \\ 
\bottomrule 
\end{tabular}
}
\caption{\textbf{Tracking Performance on ScanNet~\cite{Dai2017ScanNet}} (ATE RMSE $\downarrow$ [cm]). All scenes are evaluated on the 00 trajectory. We take the numbers from~\cite{mahdi2022eslam} for NICE-SLAM. Tracking failed for one run on Vox-Fusion on scene $\texttt{0000}$. In parenthesis we report the average over only the successful runs.}
\label{tab:scannet}
\end{table}

\subsection{Reconstruction}
\cref{tab:replica_recon} compares our method to NICE-SLAM~\cite{zhu2022nice}, Vox-Fusion~\cite{yang2022vox} and ESLAM~\cite{mahdi2022eslam} in terms of the geometric reconstruction accuracy. 
We outperform all methods on all metrics and report an average improvement of $85$ $\%$, $82$ $\%$ and $63$ $\%$ on the depth L1 metric over NICE-SLAM, Vox-Fusion and ESLAM respectively. 
\cref{fig:replica_recon} compares the mesh reconstructions of NICE-SLAM~\cite{zhu2022nice}, Vox-Fusion~\cite{yang2022vox} and our method to the ground truth mesh. 
We find that our method is able to resolve fine details to a significantly greater extent than previous approaches. 
We attribute this to our neural point cloud which adapts the point density where it is needed (i.e. close to the surface and around fine details) and conserves memory in other areas.

\subsection{Tracking}
We report the tracking performance on the Replica dataset in \cref{tab:replica_tracking}. 
On average we outperform the existing methods. 
We believe this is due to the more accurate scene representation that the neural point cloud provides.
We show that the performance of \ours transfers to real-world data by evaluating on the TUM-RGBD dataset in \cref{tab:tum}. 
We outperform all existing dense neural RGBD methods. 
Nevertheless, there is still a gap to traditional methods which employ more sophisticated tracking schemes including loop closures.
Finally, \cref{tab:scannet} shows our tracking performance on some selected ScanNet scenes, where we activate the exposure compensation module. 
We achieve competitive performance on ScanNet, but find that this dataset is generally more complex due to motion blur and specularities. 
We believe our model is more sensitive to these effects if not modeled properly compared to \eg NICE-SLAM~\cite{zhu2022nice} and Vox-Fusion~\cite{yang2022vox} which employ a large voxel size that leads to more averaging and a reduced sensitivity to specularities. 
We added a more detailed discussion to the supplementary material.

\subsection{Rendering}
\cref{tab:replica_rendering} compares rendering performance and shows improvements over existing dense neural RGBD SLAM methods.
\cref{fig:replica_rendering} shows examplary full resolution renderings where \ours yields more accurate details.

\subsection{Further Statistical Evaluation}
\begin{figure}[t]
\centering
{\footnotesize
\setlength{\tabcolsep}{1pt}
\renewcommand{\arraystretch}{1}
\newcommand{\sz}{0.5}
\begin{tabular}{cc}

\includegraphics[width=\sz\linewidth]{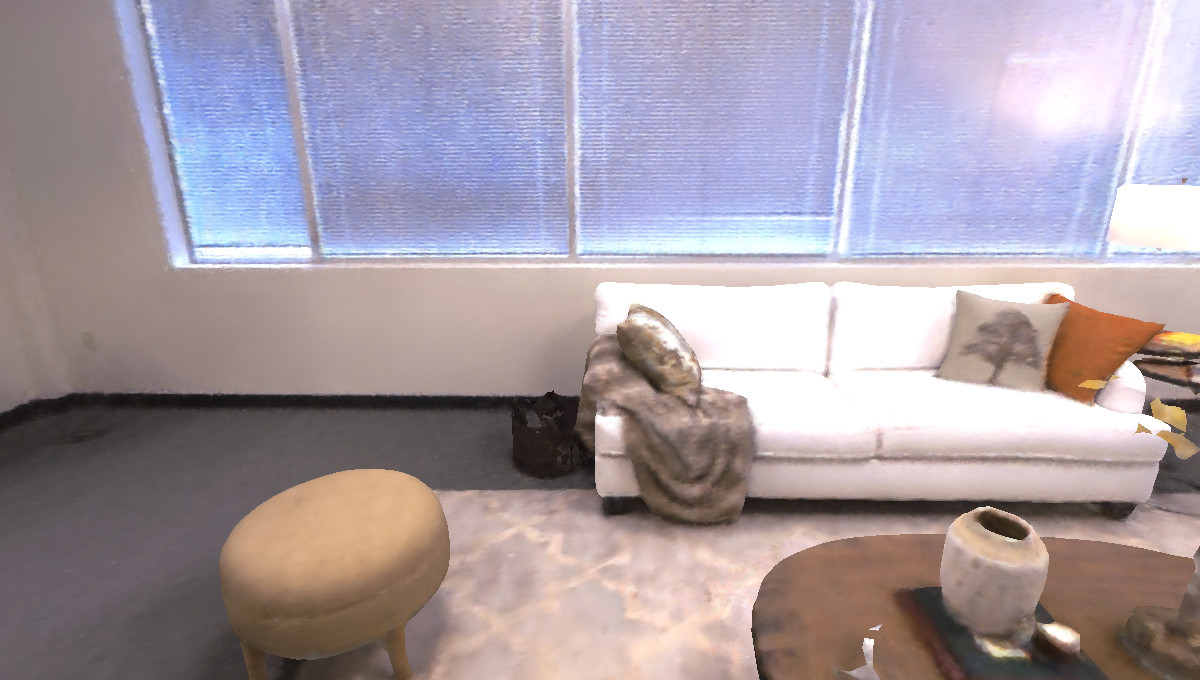} &
\includegraphics[width=\sz\linewidth]
{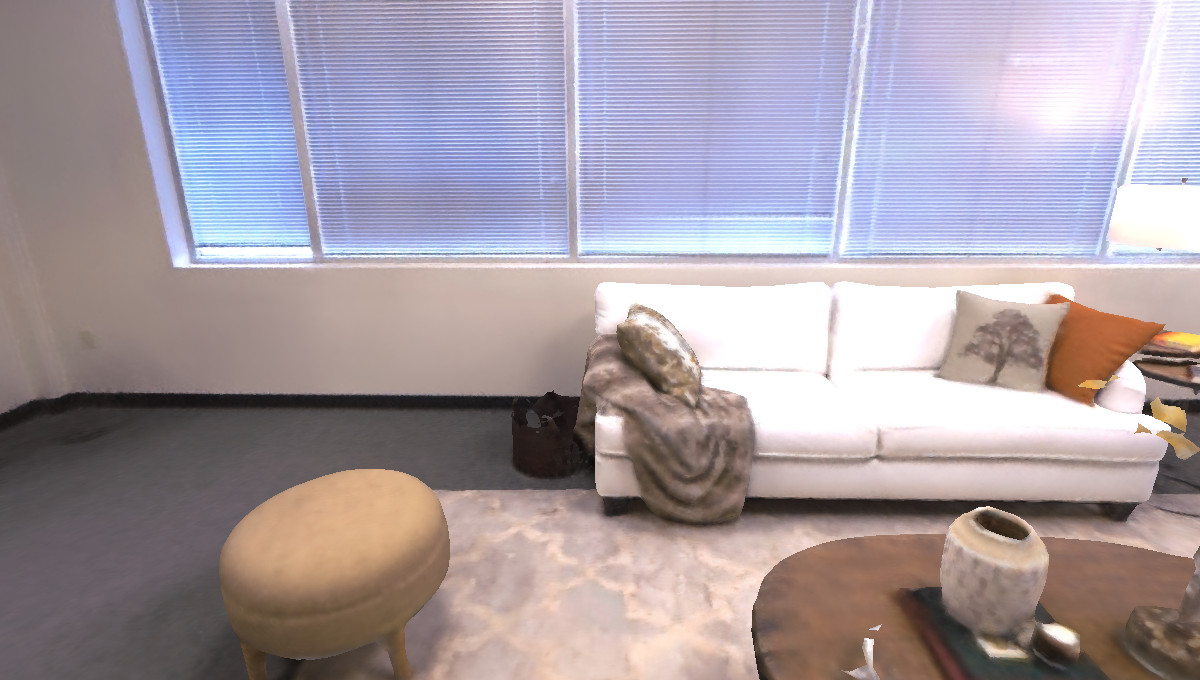}
 \\
Without $F_{\theta}$. PSNR: 27.41  & With $F_{\theta}$. PSNR: 32.09 \\
\end{tabular}
}
\caption{\textbf{Non-Linear Appearance Space.} A non-linear preprocessing via $F_{\theta}$ of the appearance features helps resolve high frequency textures like the blinds, the pot on the table and the tree print on the pillow.}
\label{fig:non_linear}
\end{figure}

%
\boldparagraph{Non-Linear Appearance Space.}
We evaluate \ours on the \texttt{Room 0} scene of the Replica dataset with and without the non-linear preprocessing network $F_{\theta}$. 
\cref{fig:non_linear} shows that a simple linear weighting of the features cannot resolve high frequency textures like the blinds while this can successfully be done when $F_{\theta}$ is optimized during runtime. 
Quantitatively, we evaluate the PSNR over the entire trajectory and show a gain of $17 \%$ ($32.09$ vs. $27.41$). 
We find that for higher tracking errors \eg on TUM-RGBD~\cite{Sturm2012ASystems} or ScanNet~\cite{Dai2017ScanNet}, the MLP $F_{\theta}$ is not helpful and we disable it. 
High-frequency appearance can only be resolved with pixel accurate poses that align the frames correctly.

\boldparagraph{Color Ablation.} We investigate the performance of our pipeline when the RGB input is not used for different settings. \cref{tab:rgb_ablation} reports performance metrics on \texttt{Room 0}. When no RGB is used for tracking, we find that the tracking performance degrades, which negatively affects the depth L1 metric and the rendering quality. The reconstruction performance is mainly determined by the depth input given good camera poses, but since RGB is useful in attaining better poses, we find that RGB information is helpful for both tracking and reconstruction.
\begin{table}[t]
  \centering
  \scriptsize
  \setlength{\tabcolsep}{7.8pt}
  \begin{tabular}{cccccc}
    \toprule
    Mapping & Tracking & ATE RMSE & Depth L1 & F1 & PSNR \\
    RGB & RGB & [cm]$\downarrow$ & [cm]$\downarrow$ & [$\%$]$\uparrow$ & [dB]$\uparrow$ \\
    \midrule
    \redx & \redx & \nd 0.59 & \nd 0.38 & \nd 91.37 & - \\
    \greencheck & \redx & \rd 0.67 & \nd 0.38 & \fs \textbf{91.49} & \nd 30.43 \\
    \greencheck & \greencheck & \fs \textbf{0.36} & \fs \textbf{0.35} & \rd 91.29 & \fs \textbf{32.15} \\
    \bottomrule
  \end{tabular}
  \caption{\textbf{Color Ablation.} The experiment shows that color information is valuable for tracking and marginally for reconstruction.}
  \label{tab:rgb_ablation}
\end{table}

\begin{figure}[t]
\centering
{\footnotesize
\setlength{\tabcolsep}{1pt}
\renewcommand{\arraystretch}{1}
\newcommand{\sz}{0.5}
\begin{tabular}{cc}
\begin{subfigure}{0.50\linewidth}
\includegraphics[width=\linewidth]{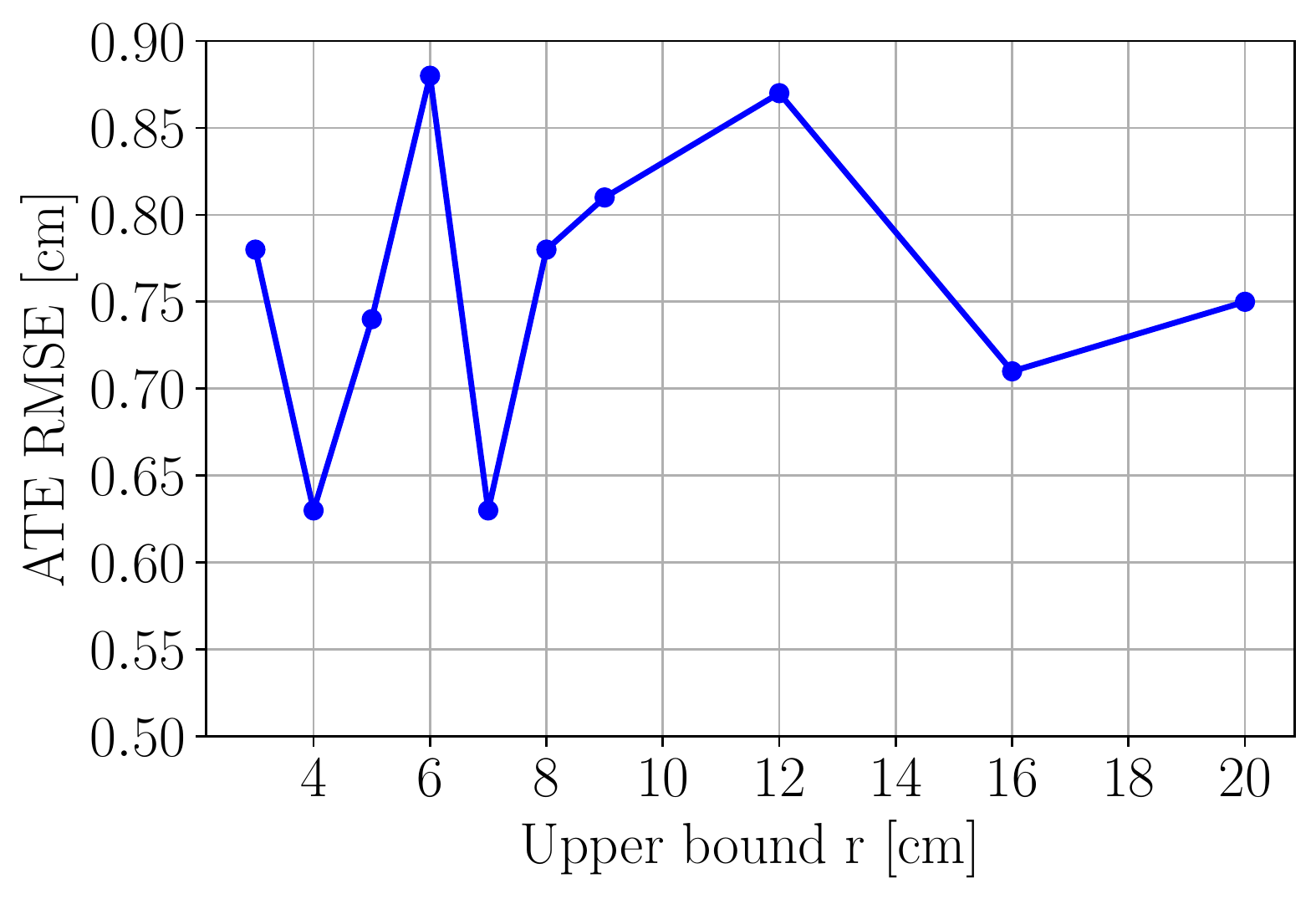}
\subcaption{}
\label{fig:dynamic_res_ablation_a}
\end{subfigure}
&
\begin{subfigure}{0.50\linewidth}
\includegraphics[width=\linewidth]{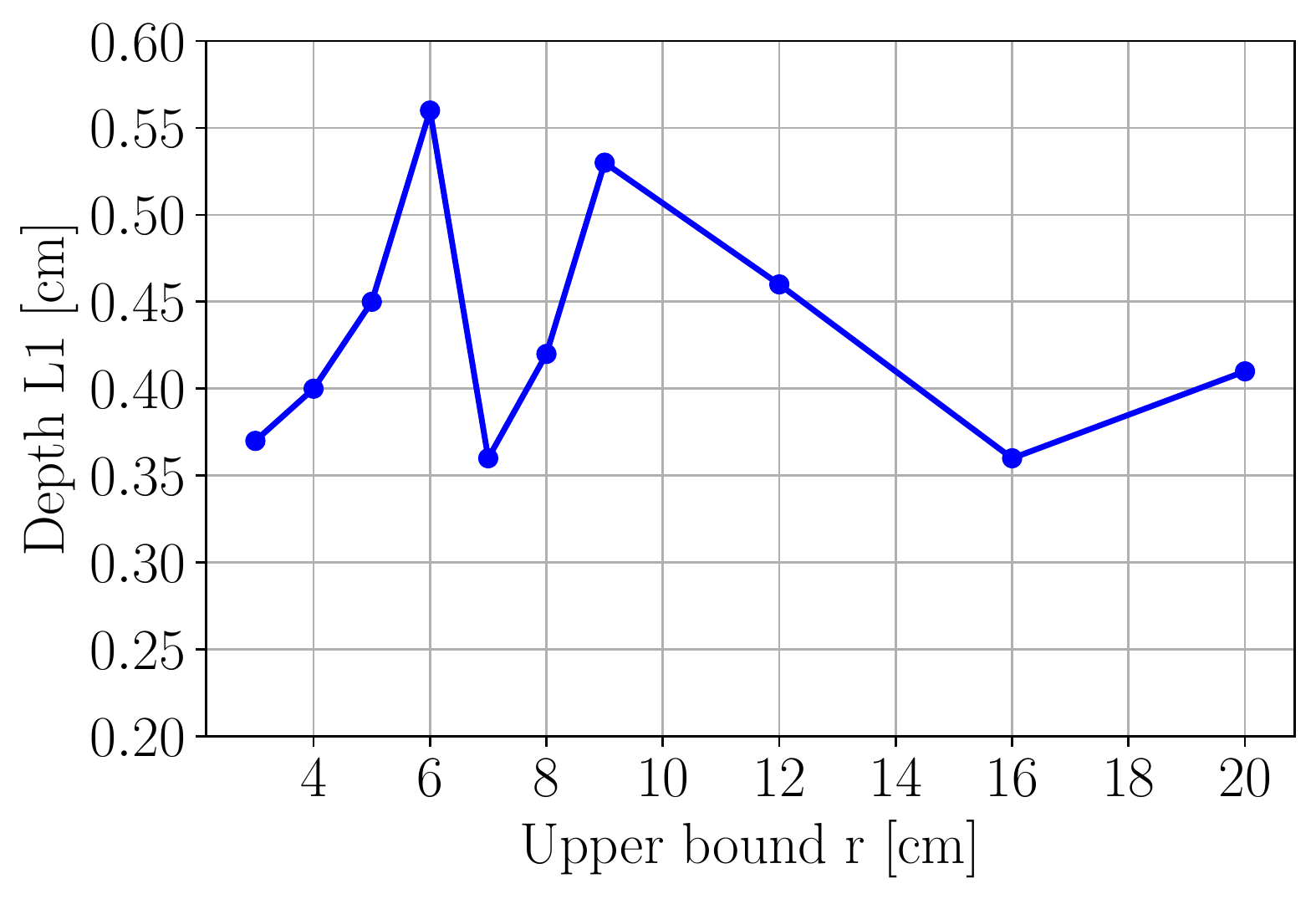}
\subcaption{}
\label{fig:dynamic_res_ablation_b}
\end{subfigure}
\\[3mm]
\begin{subfigure}{0.50\linewidth}
\includegraphics[width=\linewidth]{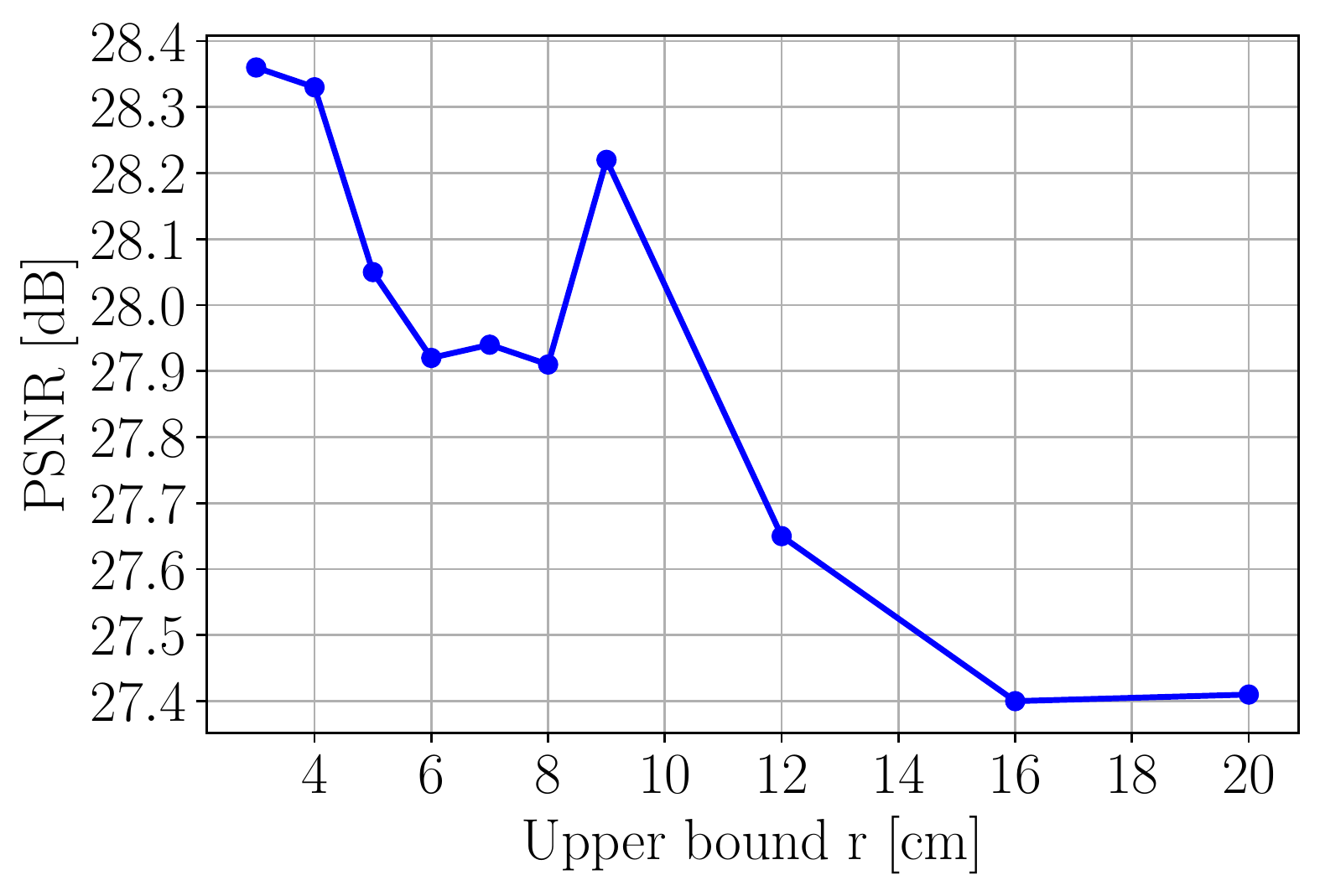}
\subcaption{}
\label{fig:dynamic_res_ablation_c}
\end{subfigure}
&
\begin{subfigure}{0.50\linewidth}
\includegraphics[width=\linewidth]{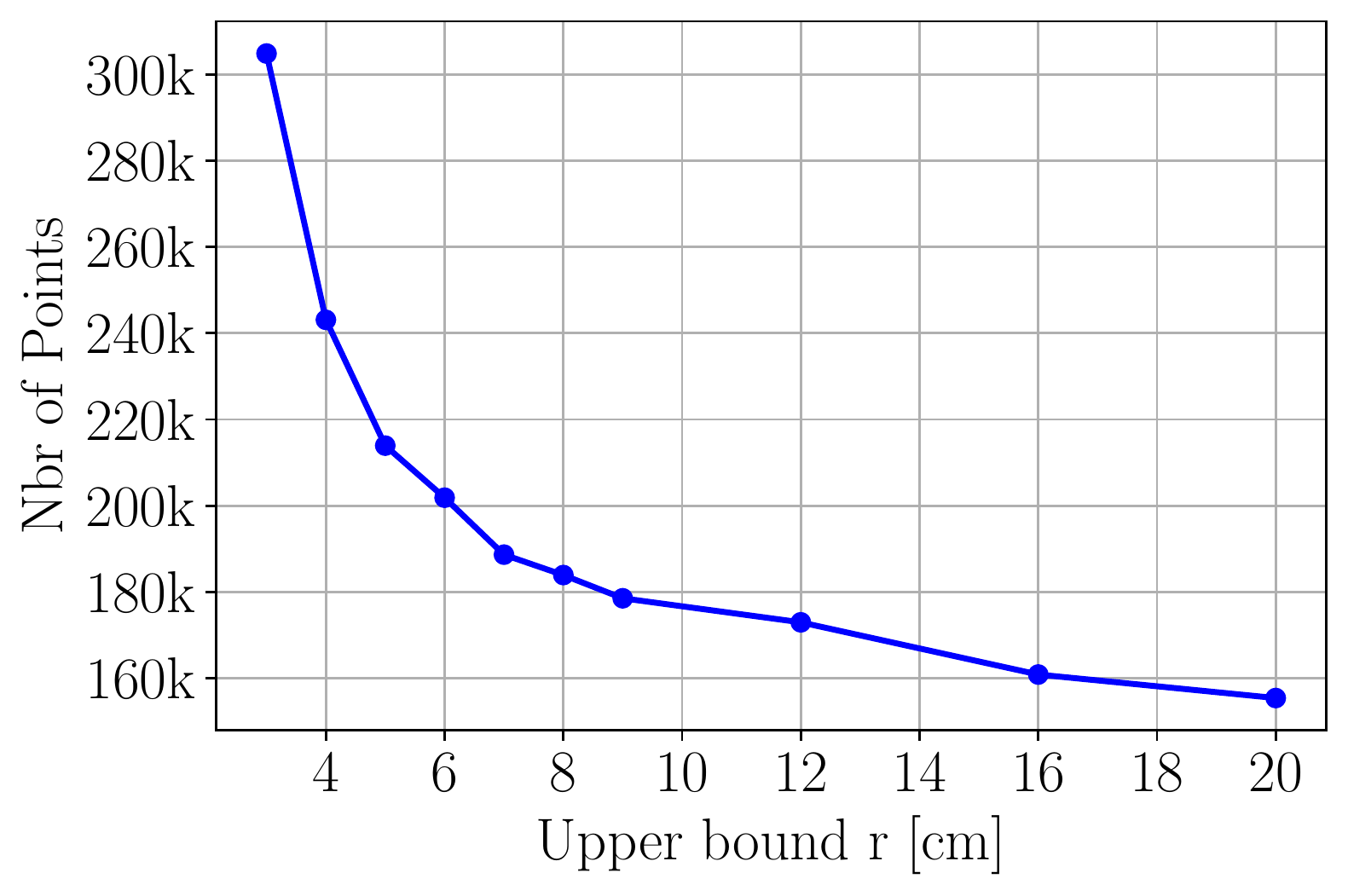}
\subcaption{}
\label{fig:dynamic_res_ablation_d}
\end{subfigure}
\end{tabular}
}
\vspace{2mm}
\caption{\textbf{Dynamic Resolution Ablation}. 
We show the performance metrics for varying upper bounds $r_u$ of the search radius on the \texttt{Room 0} scene. 
Our method is robust to compression regarding the tracking and mapping accuracy ((a) and (b) resp.). 
The rendering quality gradually degrades (c) while the memory usage starts to bottom out around $r_u = 8$ cm. 
We thus choose $r_u = 8$ cm for all experiments.}
\label{fig:dynamic_res_ablation}
\end{figure}

\boldparagraph{Dynamic Resolution Ablation.}
We show that our method is quite robust to the value of $r_u$, the upper bound for the search radius. \cref{fig:dynamic_res_ablation_a,fig:dynamic_res_ablation_b,fig:dynamic_res_ablation_c} display the ATE RMSE, depth L1 and the PSNR respectively as $r_u$ is varied. The tracking and reconstruction metrics are quite robust to $r_u$ while we see a gradual decrease in terms of the PSNR. \cref{fig:dynamic_res_ablation_d} shows the total number of neural points at the end of frame capture, for each $r_u$. We find that the curve bottoms out around $r_u = 8$ cm, which is what we use for all experiments.

\boldparagraph{Memory and Runtime Analysis.}
We report runtime and memory usage on the Replica \texttt{office 0} scene in \cref{tab:memory_runtime}. 
The tracking and mapping time is reported per iteration and frame. 
The decoder size denotes the memory footprint of all MLP networks and includes the networks $G_{\phi}$ and $F_{\theta}$. The embedding size is the total memory footprint of the scene representation. 
The memory usage of \ours falls between NICE-SLAM an Vox-Fusion while the runtime is competitive. 
The runtimes were profiled on a single Nvidia RTX 2080 Ti while Vox-Fusion used an RTX 3090.
\begin{table}[t]
  \centering
  \scriptsize
  \setlength{\tabcolsep}{1.6pt}
  \begin{tabular}{lcccccccc}
    \toprule
    Method & Tracking & Mapping & Tracking & Mapping & Decoder & Embedding \\
     & /Iteration & /Iteration & /Frame & /Frame & Size &  Size  \\
    \midrule
    NICE-SLAM~\cite{zhu2022nice} & \rd 32 ms & \rd 182 ms &\rd 1.32 s &\rd 10.92 s & \fs \textbf{0.47} MB & \rd 95.86 MB  \\
    Vox-Fusion~\cite{yang2022vox} & \fs \textbf{12} ms & \nd 55 ms & \fs \textbf{0.36} s & \fs \textbf{0.55} s & \rd 1.04 MB & \fs \textbf{0.149} MB  \\
    \ours (ours) & \nd 21 ms & \fs \textbf{33} ms &\nd 0.85 s &\nd 9.85 s & \nd 0.51 MB & \nd 27.23 MB  \\
    \bottomrule
  \end{tabular}
  \caption{\textbf{Runtime and Memory Usage on Replica} \texttt{office 0}. The decoder size is the memory of all MLP networks. The embedding size is the total memory of the scene representation. Our memory usage and runtime are competitive.}
  \label{tab:memory_runtime}
\end{table}

\boldparagraph{Limitations.} 
While our framework demonstrates competitive tracking performance on TUM-RGBD and ScanNet, we believe that a more robust system can be built to handle depth noise, by allowing the point locations to be optimized on the fly.
The local adaptation of point densities follows a simple heuristic and should ideally also be learned.
We also think that many of our empirical hyperparameters can be made test time adaptive \eg the keyframe selection strategy as well as the color gradient upper and lower bounds to determine the search radius. 
Finally, while our framework is able to substantially increase the rendering and reconstruction performance over the current state of the art, our system seems more sensitive to motion blur and specularities which we hope to address in future work.
\section{Conclusion}
\label{sec:conclusion}
We proposed \ours, a dense SLAM system which utilizes a neural point cloud for both mapping and tracking.
The data-driven anchoring of features allows to better align them with actual surface locations and the proposed dynamic resolution strategy populates features depending on the input information density.
Overall, this leads to a better balance of memory and compute resource usage and the accuracy of the estimated 3D scene representation. 
Our experiments demonstrate that \ours substantially outperforms existing solutions regarding the reconstruction and rendering accuracy while being competitive with respect to tracking as well as runtime and memory usage.\\
{\small
\boldparagraph{Acknowledgements.}
This work was supported by a VIVO collaboration project on real-time scene reconstruction and research grants from FIFA. 
We thank Danda Pani Paudel and Suryansh Kumar for fruitful discussions.
}

\clearpage

\title{Point-SLAM: Dense Neural Point Cloud-based SLAM \\ --- Supplementary Material --- }

\author{
Erik~Sandström$^{1}$\footnotemark[1] \hspace{3em} 
Yue Li$^{1}$\footnotemark[1] \hspace{3em} 
Luc~Van~Gool$^{1,2}$ \hspace{3em} 
Martin~R.~Oswald$^{1,3}$ \\
$^{1}$ETH Zürich, Switzerland \hspace{3em}
$^{2}$KU Leuven, Belgium \hspace{3em}
$^{3}$University of Amsterdam, Netherlands
}

\date{}

\maketitle

\begin{abstract}
    This supplementary material accompanies the main paper by providing further information for better reproducibility as well as additional evaluations and qualitative results.
\end{abstract}

\appendix
\section{Videos}
\label{sec:videos}
We provide an introductory video to our paper along with this document. The video describes the method and the most important results along with the visualization of the online reconstruction process of our proposed method compared to NICE-SLAM~\cite{zhu2022nice} and Vox-Fusion~\cite{yang2022vox}. Link: \url{https://youtu.be/QFjtL8XTxlU}.

\section{Method}
\label{sec:method}
In the following, we provide more details about our method, specifically the hyperparameter choices for the dynamic resolution strategy and architecture of our exposure compensation network.

\begin{figure}[t]
\centering
{\footnotesize
\setlength{\tabcolsep}{1pt}
\renewcommand{\arraystretch}{1}
\newcommand{\sz}{0.5}
\begin{tabular}{cc}
\begin{subfigure}{0.50\linewidth}
\includegraphics[width=\linewidth]{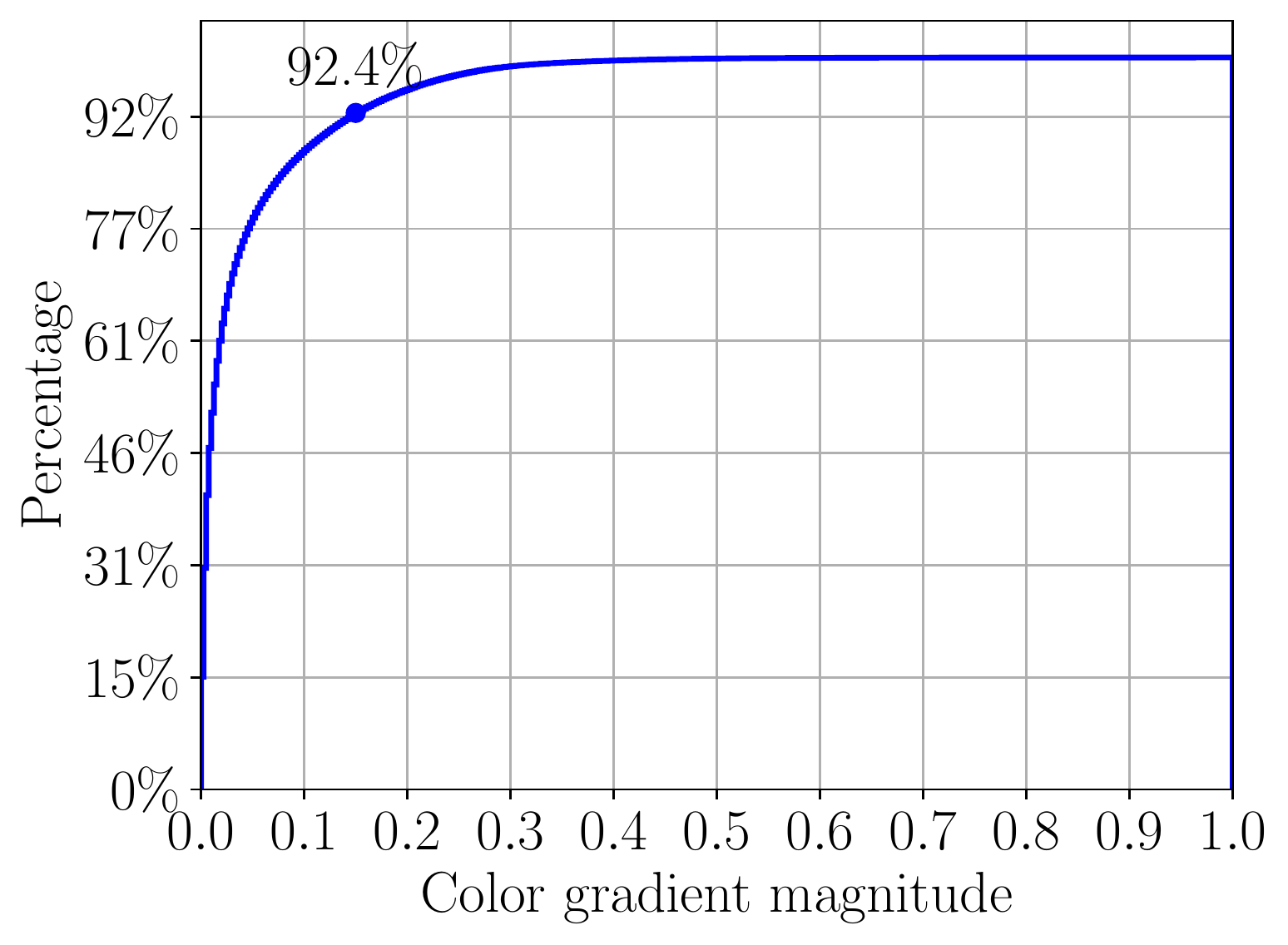}
\subcaption{\texttt{room 0}}
\label{fig:dynamic_r_stat_room0}
\end{subfigure}
&
\begin{subfigure}{0.50\linewidth}
\includegraphics[width=\linewidth]{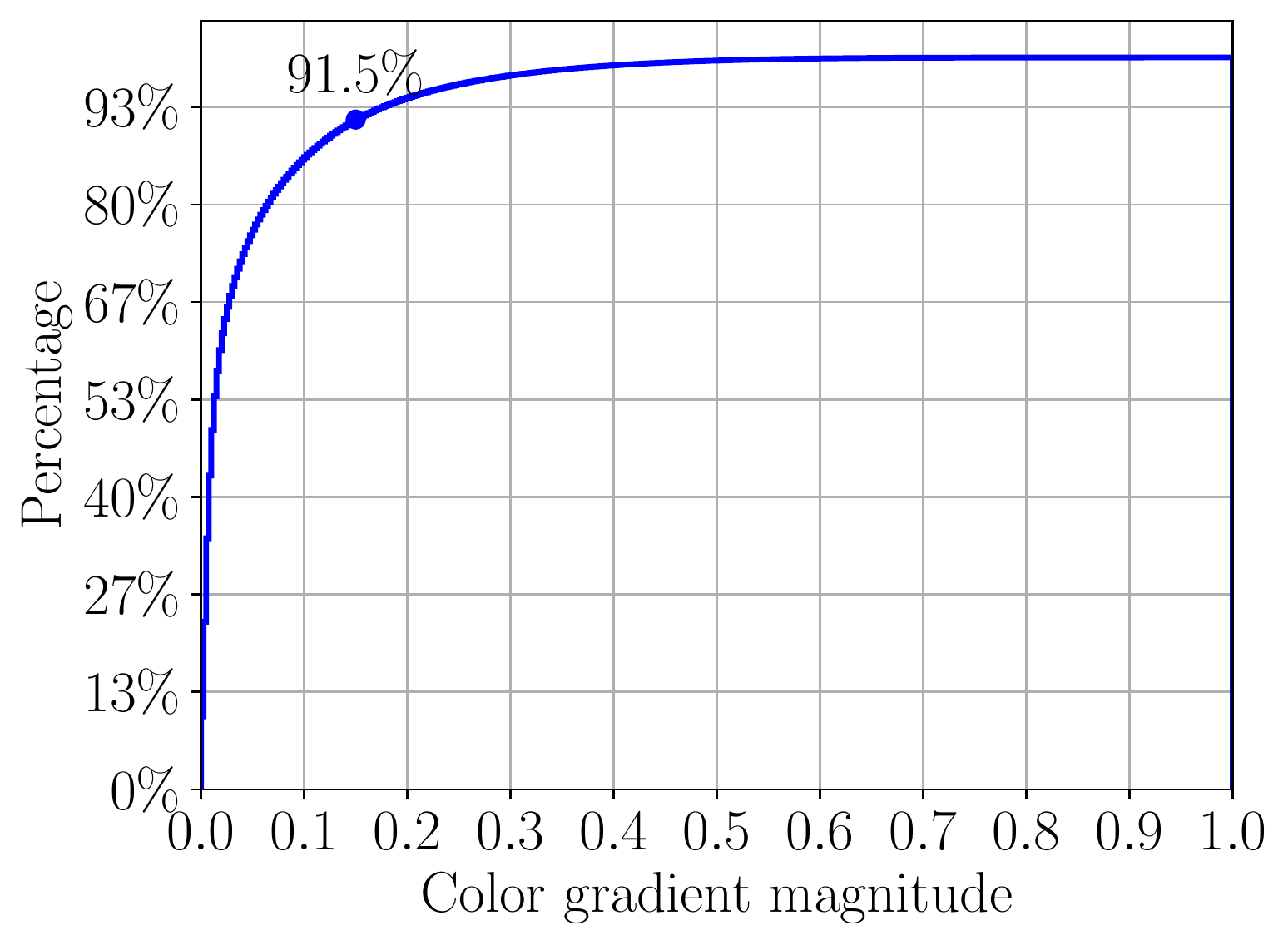}
\subcaption{\texttt{freiburg1-desk}}
\label{fig:dynamic_r_stat_fr1}
\end{subfigure}
\\[3mm]
\begin{subfigure}{0.50\linewidth}
\includegraphics[width=\linewidth]{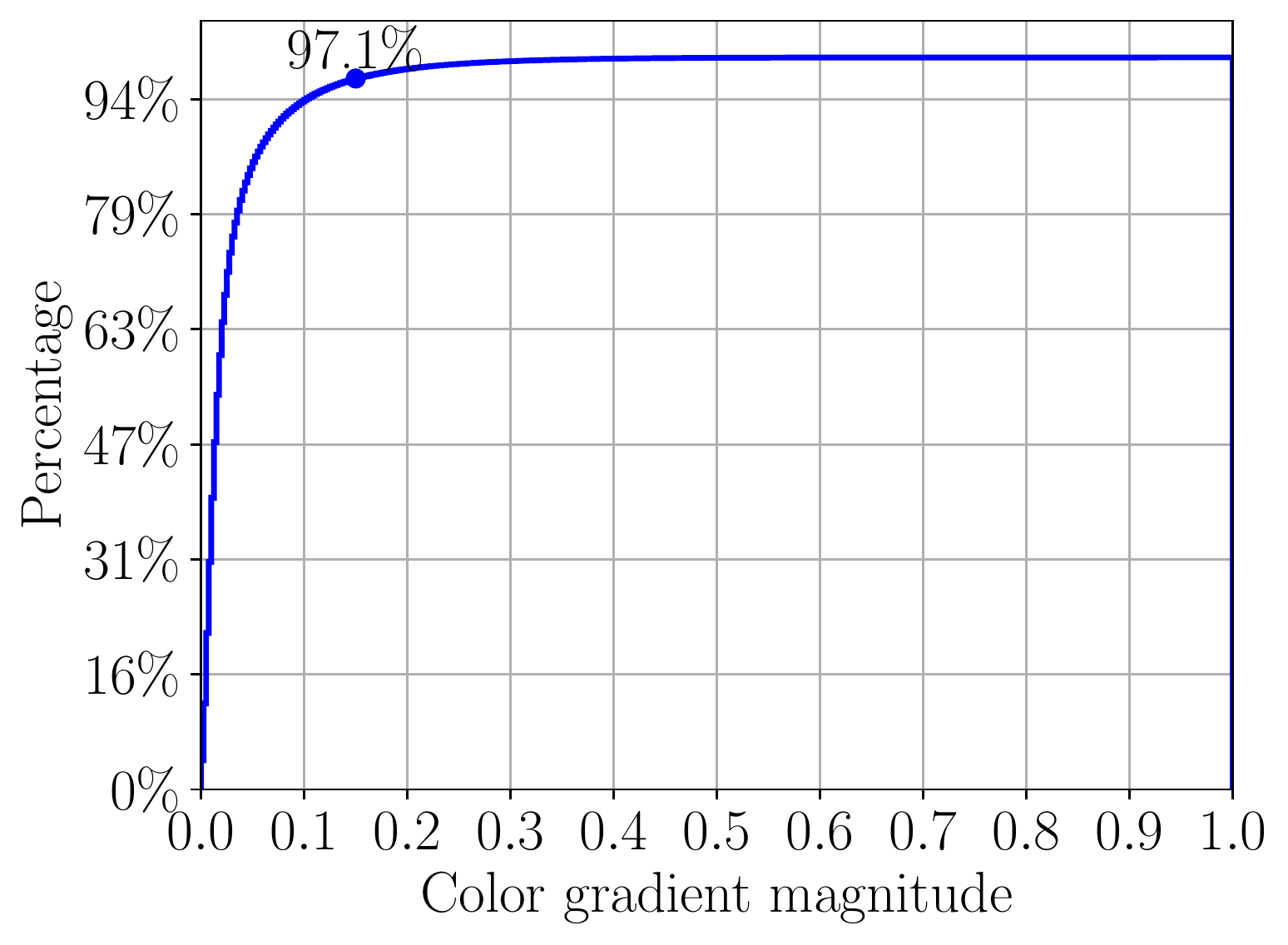}
\subcaption{\texttt{scene0000\_00}}
\label{fig:dynamic_r_stat_scene0000}
\end{subfigure}
&
\begin{subfigure}{0.50\linewidth}
\includegraphics[width=\linewidth]{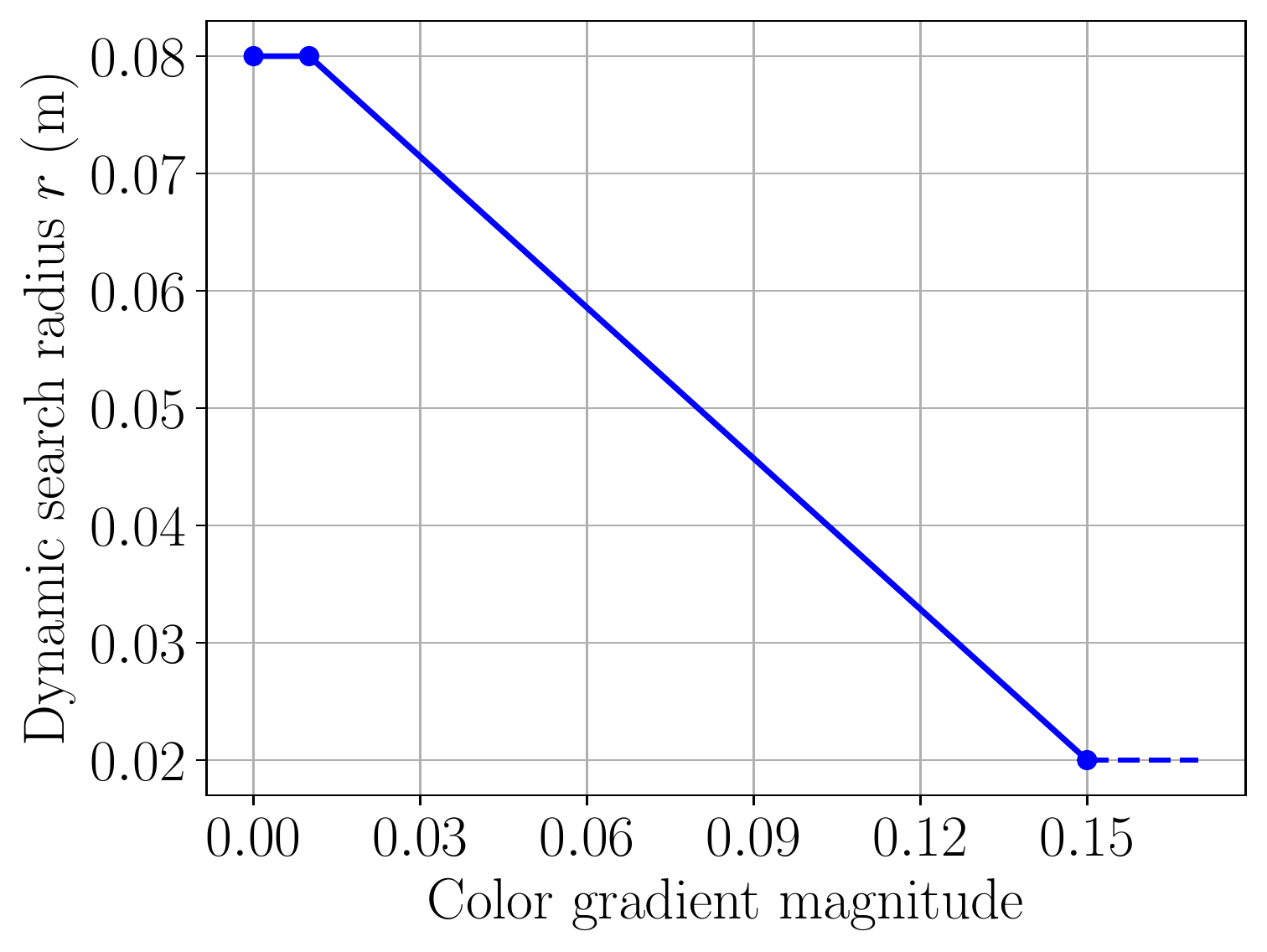}
\subcaption{Search radius $r(u, v)$.}
\label{fig:dynamic_r_interp1d_thre0.15}
\end{subfigure}
\end{tabular}
}
\vspace{2mm}
\caption{\textbf{Color Gradient Magnitude Histograms and Search Radius.} The cumulative histograms (a-c) show the percentage of pixels below a certain gradient magnitude. (d)  Search radius $r(u, v)$ as a function of the gradient magnitude at pixel $(u, v)$.}
\label{fig:dynamic_r_stat}
\end{figure}

\boldparagraph{Design Choices Dynamic Resolution Strategy.}
We empirically set the upper bound for the color gradient magnitude threshold to $g_u=0.15$ for all evaluated datasets. Based on the pre-calculated cumulative gradient magnitude histograms shown in \cref{fig:dynamic_r_stat_room0,fig:dynamic_r_stat_fr1,fig:dynamic_r_stat_scene0000} (depicting \texttt{room 0}, \texttt{freiburg1-desk}, and \texttt{scene0000\_00}), we observe that approximately less than 10\% of all pixels exceed the upper threshold. The threshold $g_u=0.15$ strikes a good balance between resolving highly textured regions and model compression. The cumulative histograms in \cref{fig:dynamic_r_stat_room0,fig:dynamic_r_stat_fr1,fig:dynamic_r_stat_scene0000} also reveal that the majority of pixels have close to zero gradient magnitude. We use a lower bound $g_l = 0.01$ for all datasets. The search radius $r(u, v)$ as a function of the color gradient magnitude at pixel $(u, v)$ is shown in \cref{fig:dynamic_r_interp1d_thre0.15}.

\boldparagraph{Exposure Network Architecture.}
For the exposure compensation network $G_{\phi}$, we use one hidden layer with 128 neurons followed by a $\mathrm{softplus}$ activation. The input latent vector is 8-dimensional and the output is 12-dimensional, which is reshaped into a 3$\times$3 affine matrix and a 3$\times$1 translation vector. The network $G_{\phi}$ and latent vector are jointly optimized both during mapping and tracking. The latent vector is put in shared memory and if the current frame is used for mapping, the latent vector is first optimized during tracking then refined during mapping. We did not explore other optimization strategies which could potentially improve performance.

\section{Implementation Details}
\label{sec:imp_details}
We use PyTorch 1.12 and Python 3.10 to implement the pipeline. 
Training is done with the Adam optimizer and the default hyperparameters \emph{betas} $= (0.9, 0.999)$, \emph{eps} = $1e$-$08$ and \emph{weight\_decay} = 0. The results are gathered using various Nvidia GPUs, all with a maximum memory of 12 GB. The learning rate for tracking is 0.002 on Replica and TUM-RGBD and 0.0005 on ScanNet. We use a learning rate of 0.03 for the initial geometry only optimization stage and 0.005 during color and geometry optimization stage.
 \cref{tab:dataset-params} describes other dataset-specific hyperparameters such as the mapping window size which describes how many frames (current frame and selected keyframes) are used during mapping. We also follow~\cite{zhu2022nice} and use a simple keyframe selection strategy which adds frames to the keyframe database at regular intervals (see also \cref{tab:dataset-params}). 
 
\begin{table}[tb]
  \footnotesize
  \setlength{\tabcolsep}{6pt}
  \renewcommand{\arraystretch}{1.05}
  \newcommand{\ccg}{}
  \begin{tabular}{lcccccc}
    \toprule
    \multirow{2}{*}{\ccg Dataset}& \ccg Map   & \ccg Keyframe & \ccg Map     & \ccg Track & \ccg Map \\ 
    & \ccg Every & \ccg Every    & \ccg Window  & \ccg Iter. & \ccg Iter. \\
    \midrule
    Replica~\cite{straub2019replica}   & 5 & 20 & 12 &  40 & 300 \\
    TUM-RGBD~\cite{Sturm2012ASystems}  & 2 & 50 & 10 & 200 & 150 \\
    ScanNet~\cite{Dai2017ScanNet}      & 5 & 10 & 20 & 100 & 300 \\
    \bottomrule
  \end{tabular}
  \caption{\textbf{Parameter Configurations on Tested Datasets.} Map Every: how often (in frames) mapping is done. Map Window: How many keyframes that are sampled to overlap with the current viewing frustum for mapping. Iter.: Iterations (optimization steps).}
  \label{tab:dataset-params}
\end{table}

\section{Evaluation Metrics}
\label{sec:metrics}
\boldparagraph{Mapping.}
We use the following five metrics to quantify the reconstruction performance. 
We compare the ground truth mesh to the predicted mesh.
The F-score is defined as the harmonic mean between Precision (P) and Recall (R), $F = 2\frac{PR}{P+R}$. 
Precision is defined as the percentage of points on the predicted mesh which lie within some distance $\tau$ from a point on the ground truth mesh.
Vice versa, Recall is defined as the percentage of points on the ground truth mesh which lie within the same distance $\tau$ from a point on the predicted mesh. In all our experiments, we use a distance threshold $\tau = 0.01$ m. Before the Precision and Recall are computed, the input meshes are aligned with the iterative closest point (ICP) algorithm.
We use the evaluation script provided by the authors of~\cite{Sandstrom2022LearningFusion}\footnote{\url{https://github.com/tfy14esa/evaluate_3d_reconstruction_lib}}. 
Finally, we report the depth L1 metric which renders depth maps from randomly sampled view points from the reconstructed and ground truth meshes. The depth maps are then compared and the L1 error is reported and averaged over 1000 sampled views. We use the evaluation code provided by~\cite{zhu2022nice}.

\boldparagraph{Tracking.} We use the absolute trajectory error (ATE) RMSE~\cite{Sturm2012ASystems} to compare tracking error across methods. This computes the translation difference between the estimated trajectory and the ground truth. Before evaluating the ATE RMSE, we align the trajectories with Horn's closed form solution~\cite{horn1988closed}.

\section{More Experiments}
\label{sec:exp}

\begin{figure}[t]
\centering
{\footnotesize
\setlength{\tabcolsep}{1pt}
\renewcommand{\arraystretch}{1}
\newcommand{\sz}{0.5}
\begin{subfigure}[t]{\linewidth}
\begin{tabular}{cc}
\includegraphics[width=0.5\linewidth]{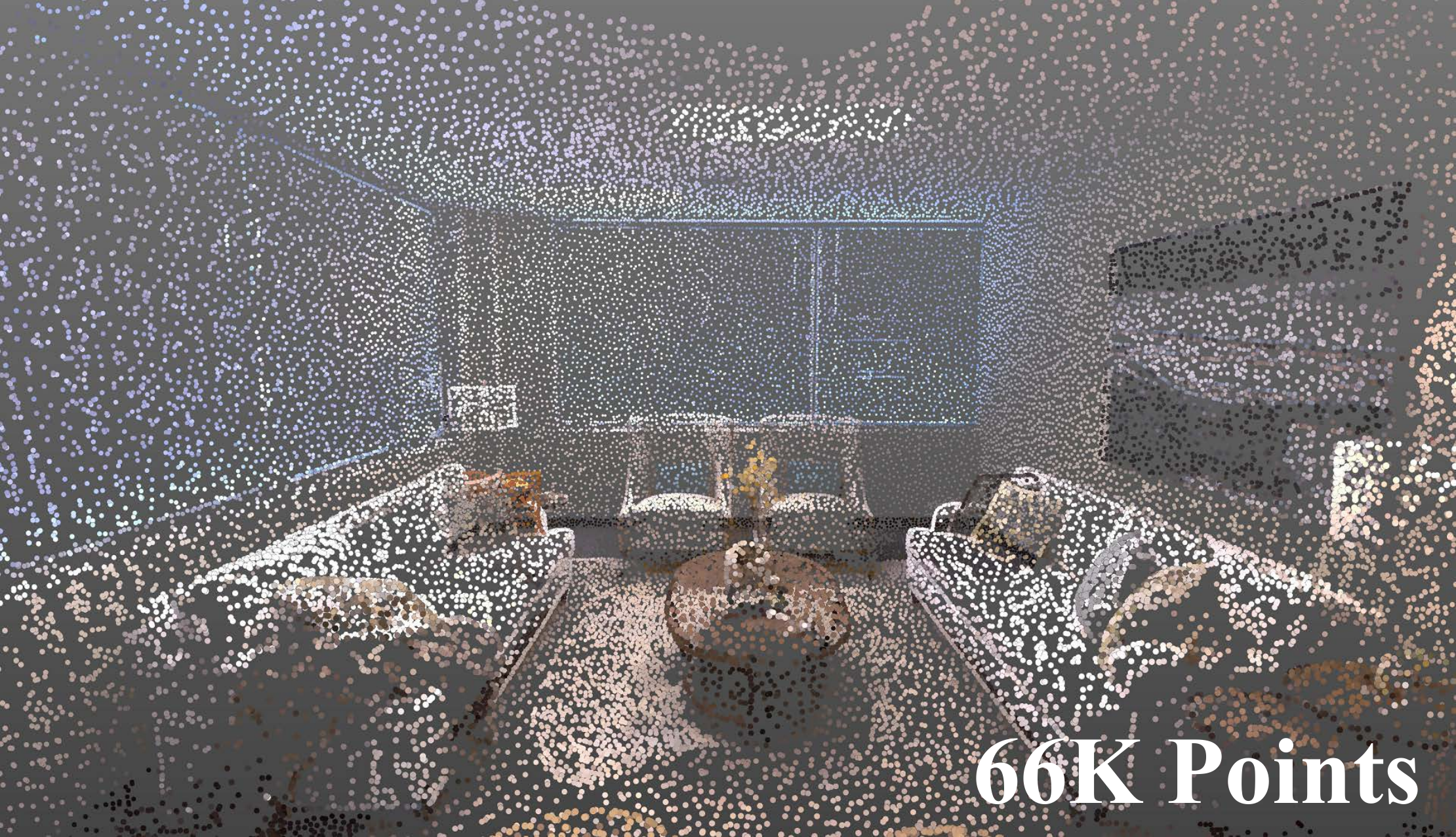}
&
\includegraphics[width=0.5\linewidth]{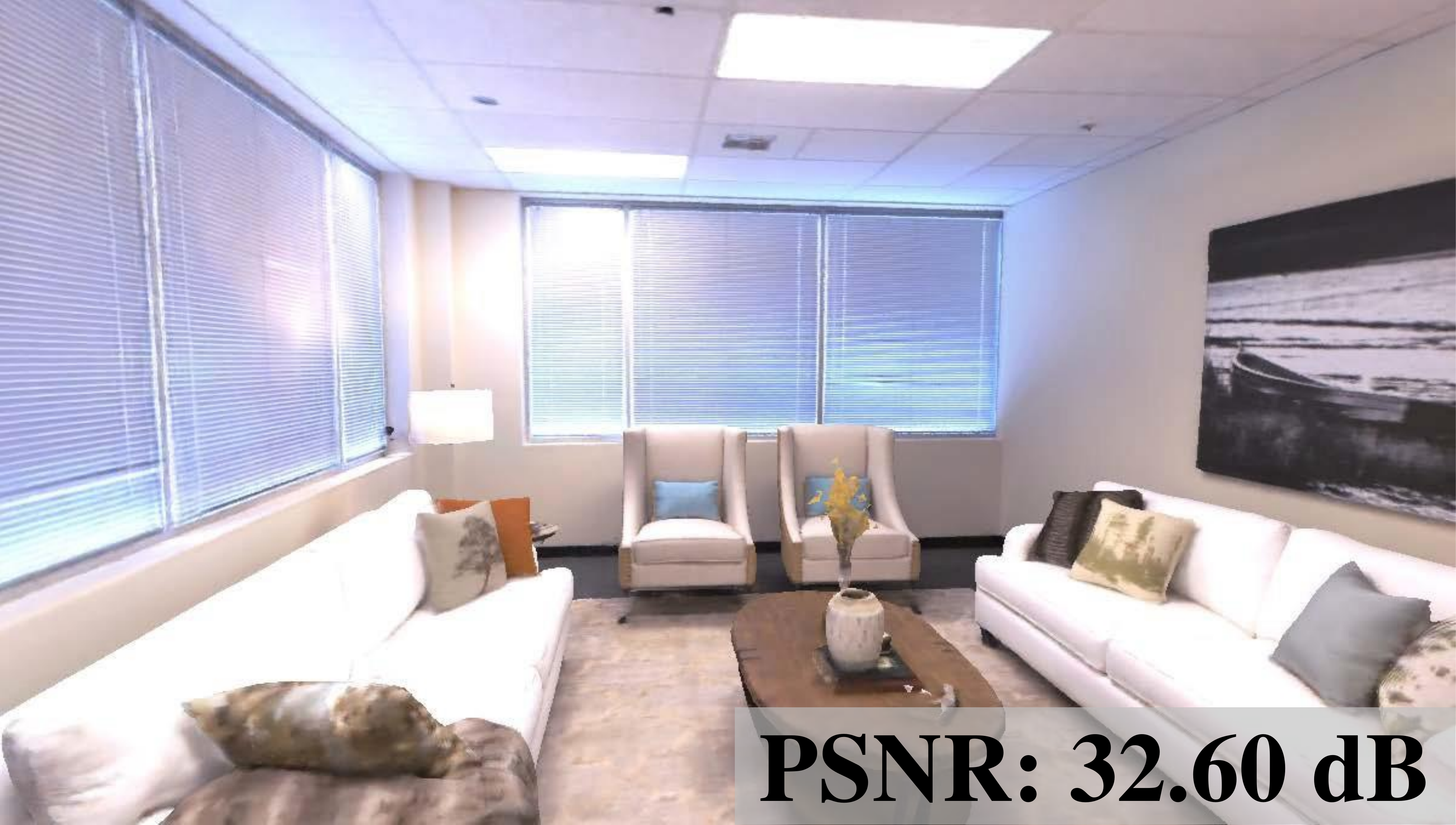}
\end{tabular}
\vspace{-2mm}
\subcaption{\quad $r_l=r_{u}=4cm$}
\vspace{2mm}
\label{fig:pcl_density_a}
\end{subfigure}
\vspace{2mm}
\begin{subfigure}[t]{\linewidth}
\begin{tabular}{cc}
\includegraphics[width=0.5\linewidth]{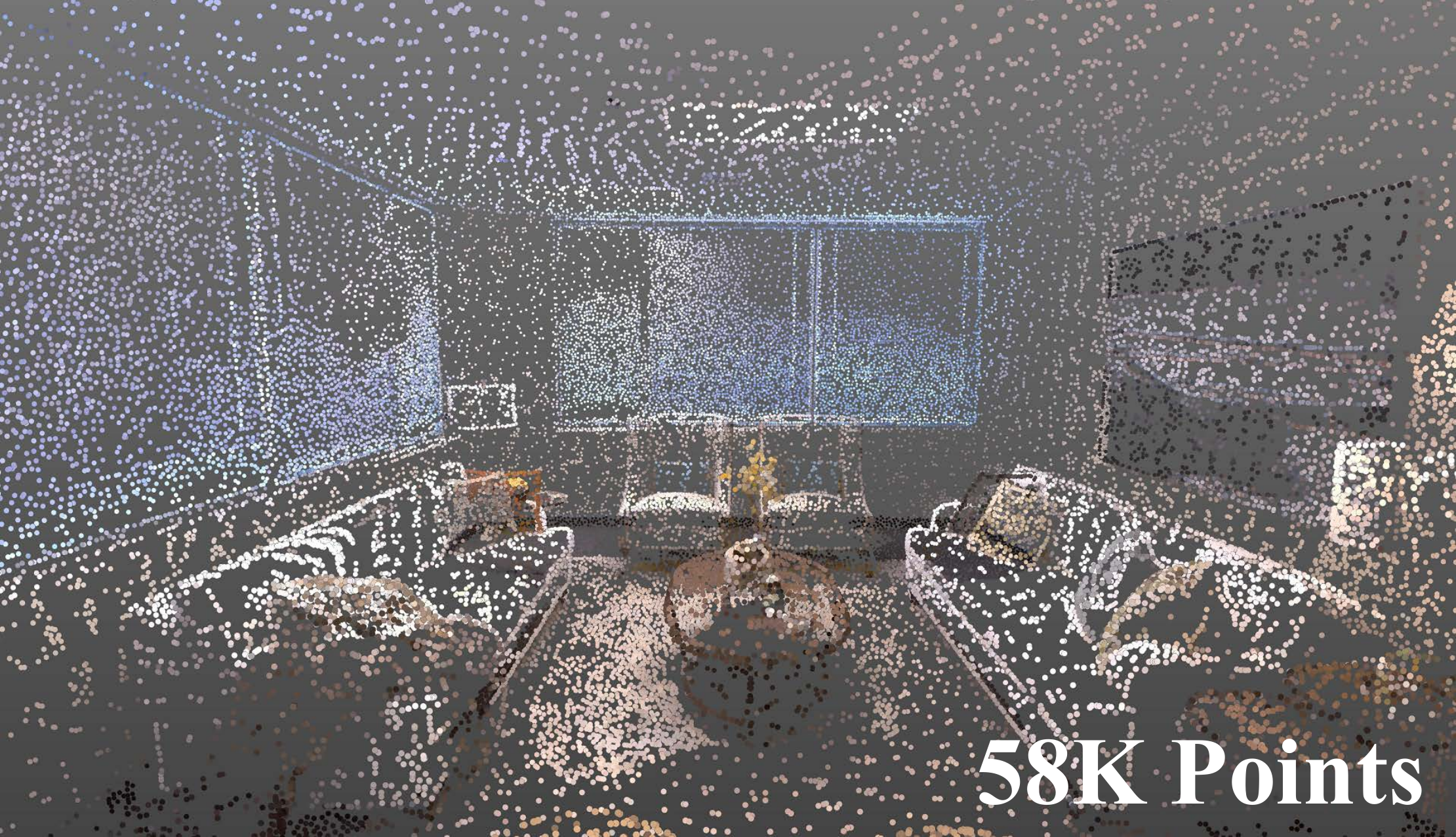}
&
\includegraphics[width=0.5\linewidth]{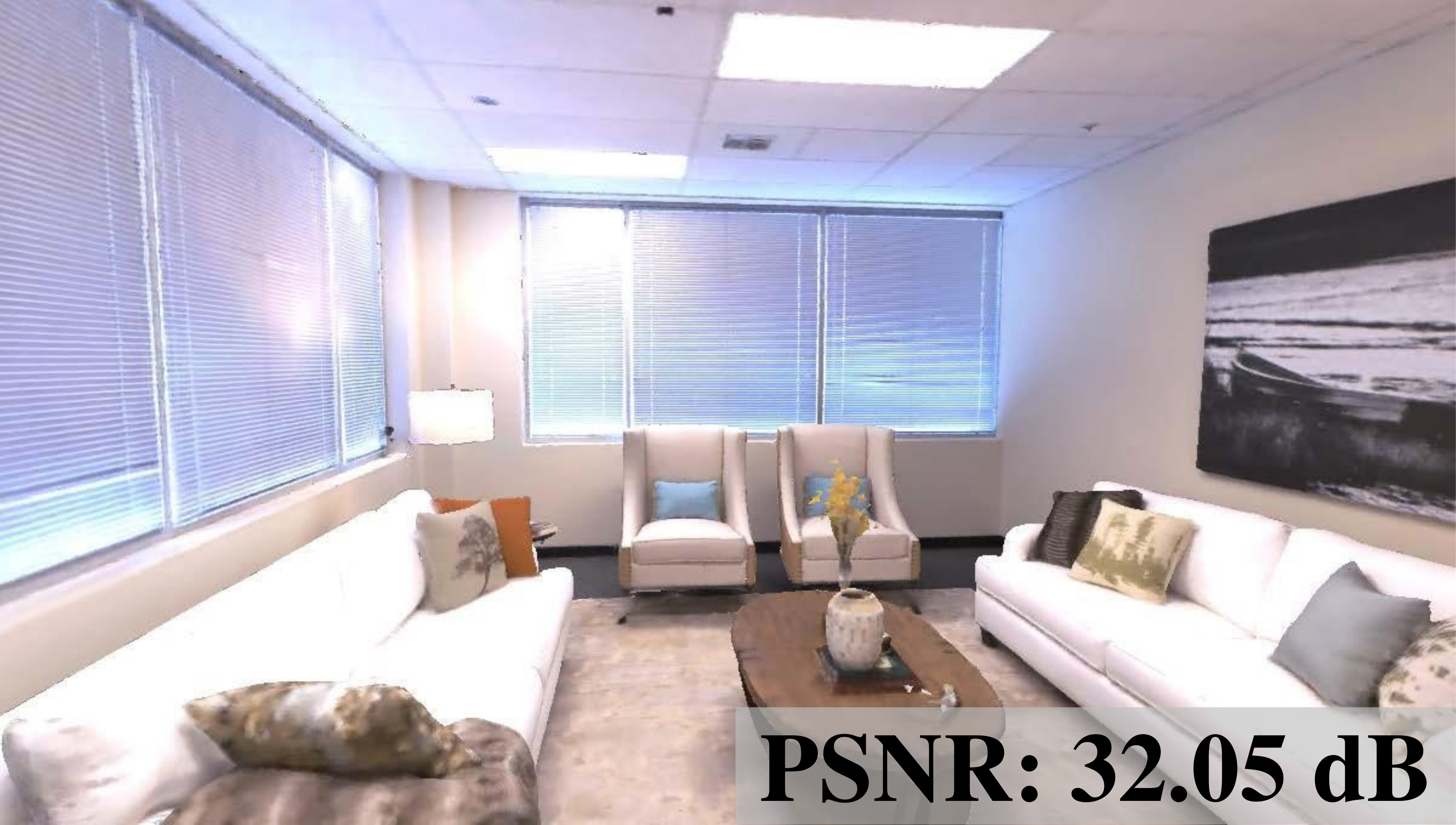}
\end{tabular}
\vspace{-2mm}
\subcaption{\quad $r_l=r_{u}=8cm$}
\label{fig:pcl_density_b}
\end{subfigure}
\vspace{2mm}
\begin{subfigure}[t]{\linewidth}
\begin{tabular}{cc}
\includegraphics[width=0.5\linewidth]{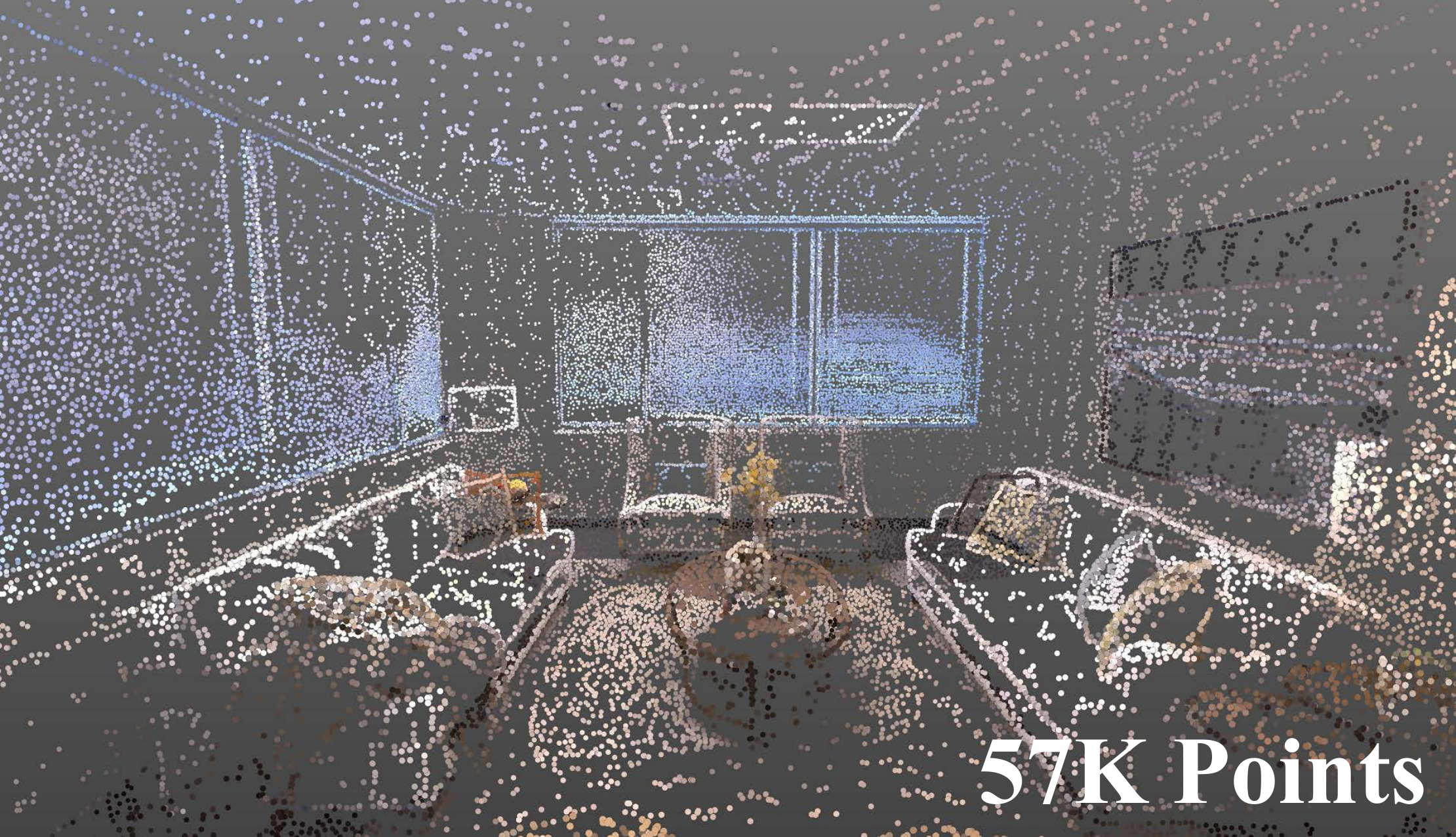}
&
\includegraphics[width=0.5\linewidth]{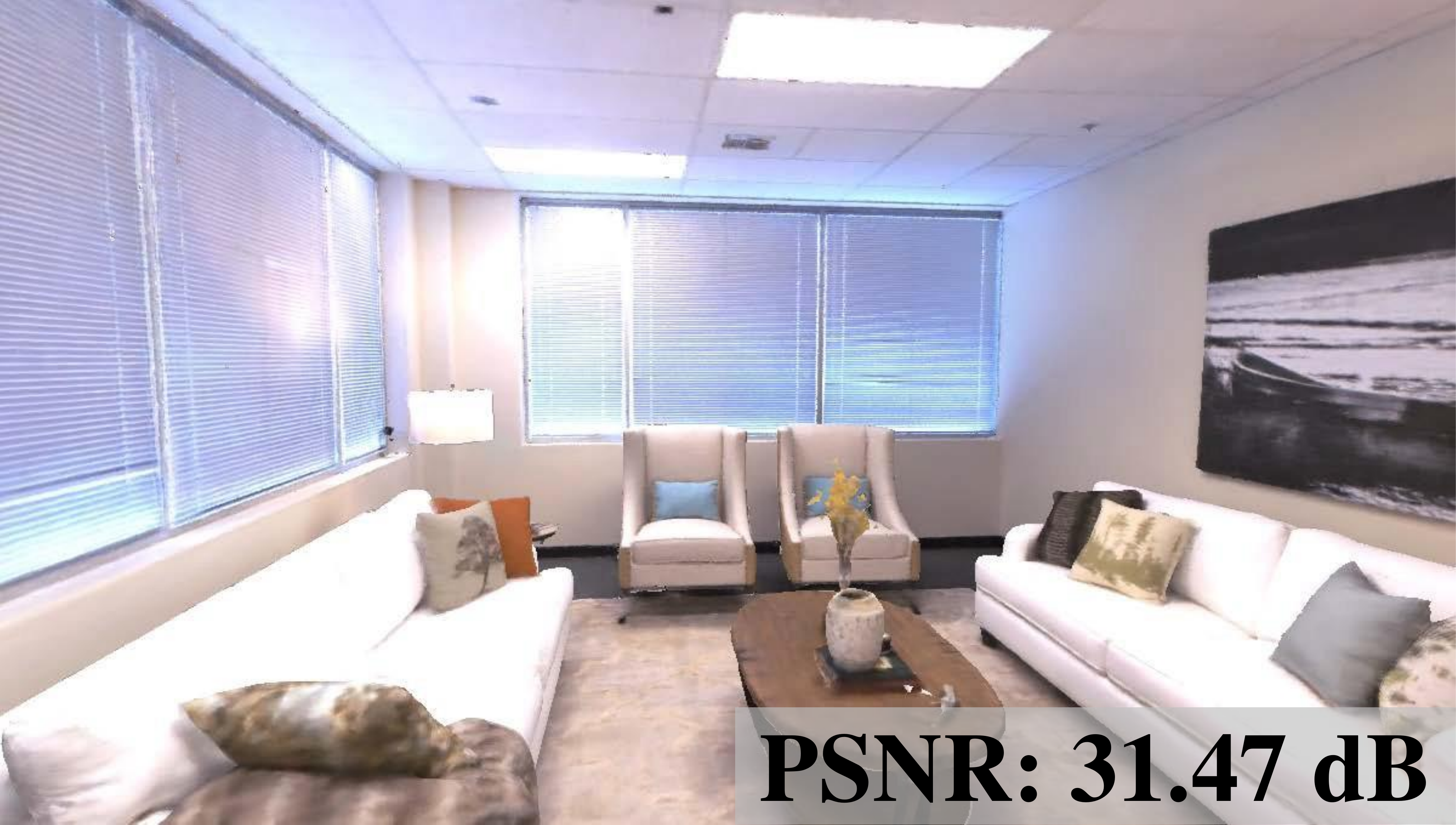}
\end{tabular}
\vspace{-2mm}
\subcaption{\quad $r_l=r_{u}=12cm$}
\label{fig:pcl_density_c}
\end{subfigure}
\vspace{2mm}
\begin{subfigure}[t]{\linewidth}
\begin{tabular}{cc}
\includegraphics[width=0.5\linewidth]{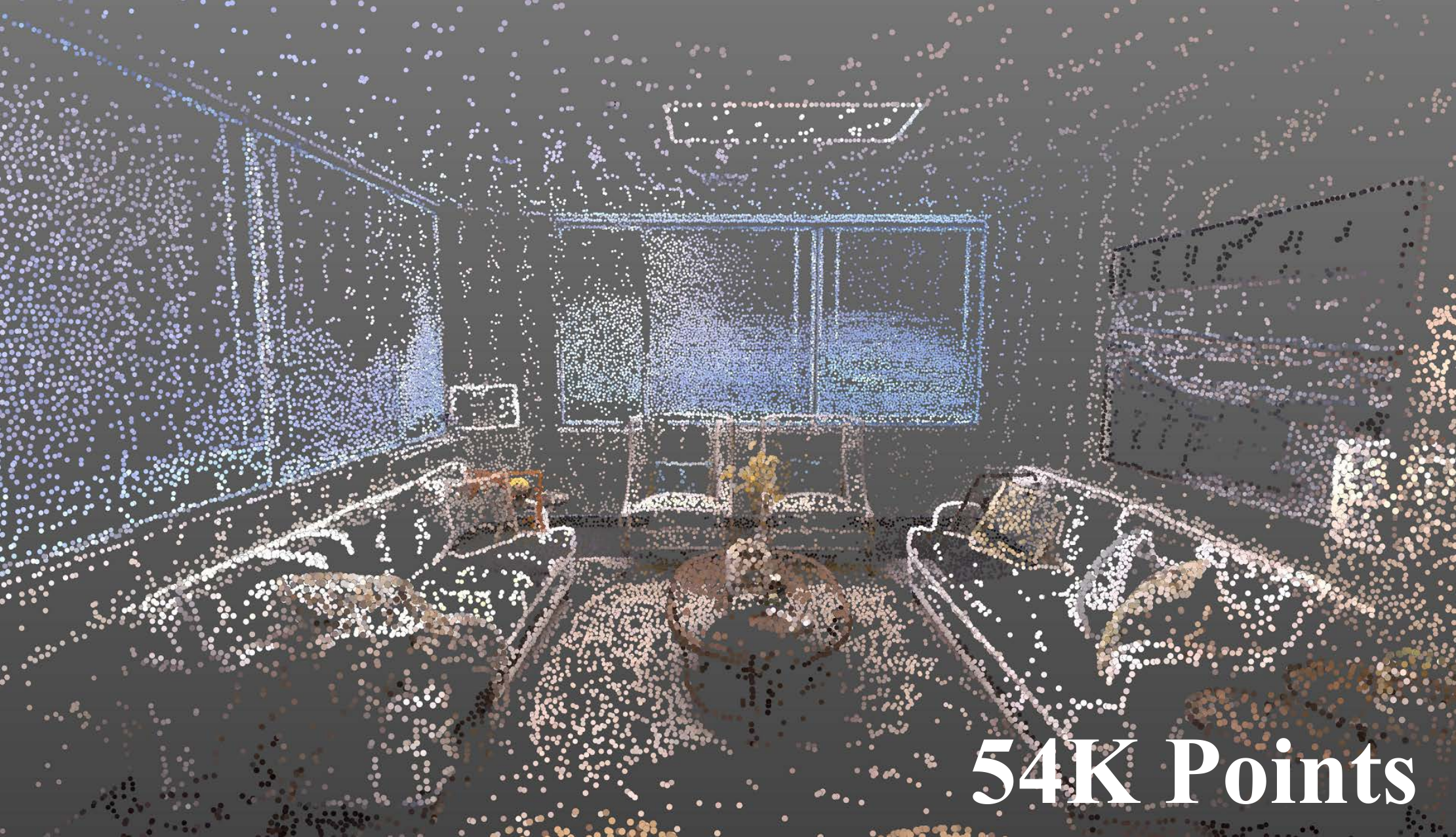}
&
\includegraphics[width=0.5\linewidth]{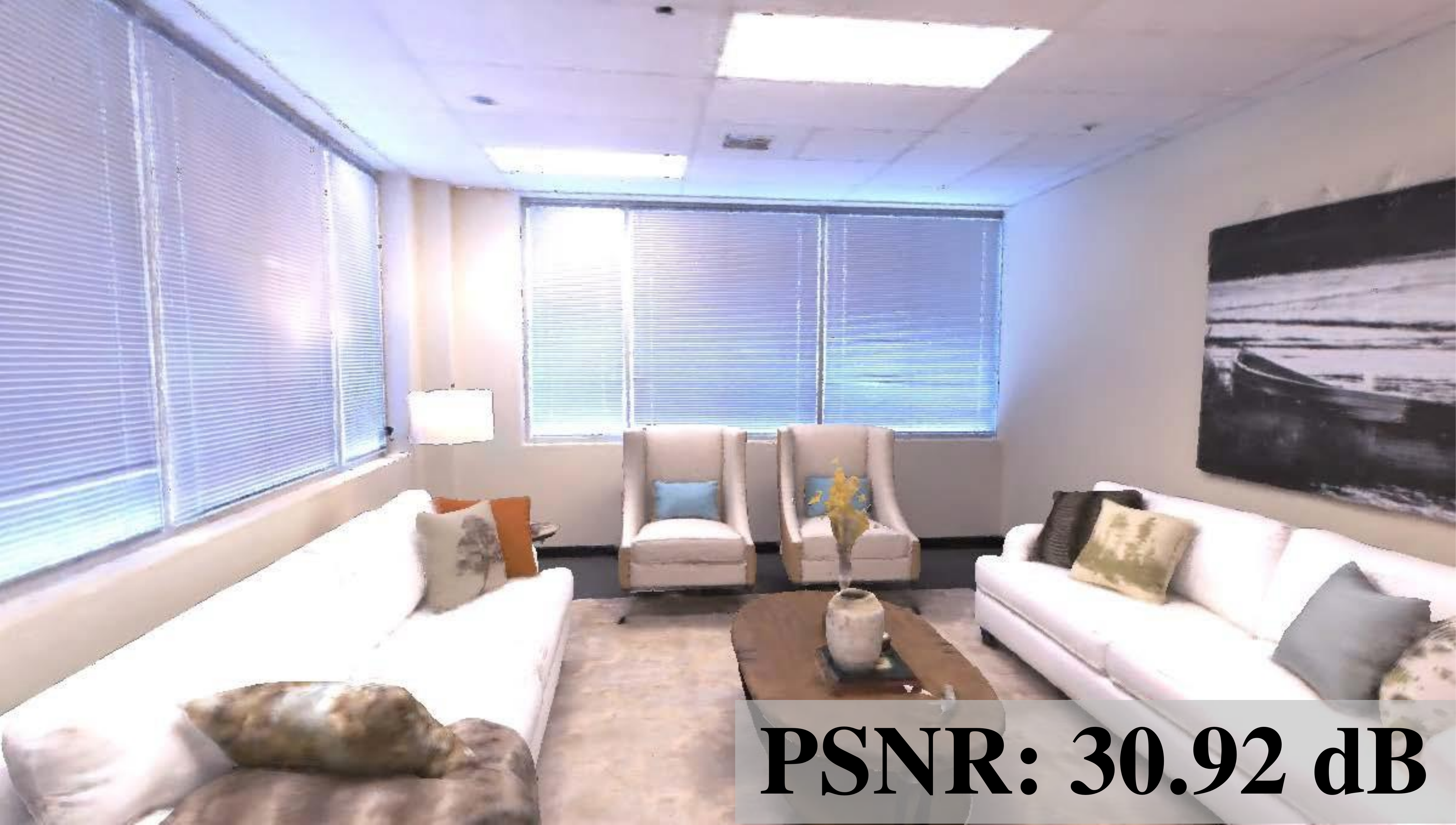}
\end{tabular}
\vspace{-2mm}
\subcaption{\quad $r_l=r_{u}=16cm$}
\label{fig:pcl_density_d}
\end{subfigure}
}
\caption{\textbf{Dynamic Search Radius Visualization.} With the dynamic point density enabled (\cref{fig:pcl_density_b,fig:pcl_density_c,fig:pcl_density_d}), we use less points than without the dynamic point density (\cref{fig:pcl_density_a}) while preserving high point densities in texture-rich regions, such as the window blinds.}
\label{fig:pcl_density_comparison}
\end{figure}

\boldparagraph{Dynamic Search Radius Visualization.} In the main paper, we ablate how the tracking, reconstruction and rendering performance metrics vary as the upper bound $r_u$ of the search radius is changed. In \cref{fig:pcl_density_comparison} we show qualitative examples of the surface point cloud for some selected values $r_u$ on $\texttt{room 0}$. \cref{fig:pcl_density_a} shows the point cloud without dynamic resolution using fixed $r_l = r_{u} = 4cm$ for all points. In \cref{fig:pcl_density_b,fig:pcl_density_c,fig:pcl_density_d} we enable dynamic resolution and show how the point density varies across the scene for different values of $r_u$. We use $r_l=2cm$ for all these experiments. The total number of points decreases from 66K to 54K. This is due to the sparsification of the point density in regions with little texture information. It is clear that using a dynamic search radius preserves rich textures, while effectively sparsifying the point density in textureless regions such as the sofas and walls.

\boldparagraph{Additional Qualitative Reconstructions.} In \cref{fig:replica_recon} we show additional reconstructions from the Replica dataset where our method is compared to NICE-SLAM~\cite{zhu2022nice} and Vox-Fusion~\cite{yang2022vox}.

\boldparagraph{Additional Qualitative Renderings.} In \cref{fig:replica_rendering} we show additional renderings from the Replica dataset where our method is compared to NICE-SLAM~\cite{zhu2022nice} and Vox-Fusion~\cite{yang2022vox}.

\boldparagraph{Evaluating Depth Error on Rendered Depth Maps.}
\cref{tab:depth-l1-render} shows additional results when the depth L1 error is evaluated directly on the rendered depth maps from the neural point cloud. This is in contrast to the main paper where we report the depth L1 on the predicted mesh from randomly sampled views. Compared to the mesh depth L1 metric, we report one order of magnitude smaller error from our rendered depth maps along the estimated trajectory.

\begin{table}[tb]
  \footnotesize
  \setlength{\tabcolsep}{4pt}
  \renewcommand{\arraystretch}{1.05}
  \newcommand{\ccg}{}
  \begin{tabular}{lcccccc}
    \toprule
    Metric &\texttt{off 0} & \texttt{off 1}& \texttt{off 2} & \texttt{off 3}& \texttt{off 4} \\
    \midrule
    Mesh Depth L1    & 0.30 & 0.61 & 0.53 &  0.54 & 0.45 \\
    Rendering Depth L1  & \textbf{0.037} & \textbf{0.025} & \textbf{0.054} & \textbf{0.082} & \textbf{0.061} \\
    \bottomrule
  \end{tabular}
  \caption{\textbf{Depth L1 Error [cm] on Replica~\cite{straub2019replica}.} The table reports the depth L1 error for the rendered depth map and for the reconstructed mesh (after TSDF fusion and Marching cubes).
  The results in the main paper only report the depth L1 error for the mesh. }
  \label{tab:depth-l1-render}
\end{table}

\boldparagraph{Adaptive Mapping Ablation.} As mentioned in the implementation details in the main paper, the number of mapping iterations is computed as $m_i = m_i^dn/300$, where $m_i^d$ is the default mapping iterations and $n$ is the number of added points for the frame at hand. By default, we clip $m_i$ to lie within $[0.95m_i^d,2m_i^d]$. We further decrease the lower bound from 0.95 to [0.9, 0.05] on the \texttt{office 0} scene. The resulting average mapping iterations per frame and associated per frame mapping runtimes are presented in \cref{tab:ablation-min-iter-ratio}. We find that we can speed up the mapping phase by a factor of four compared to the results reported in the main paper.
\begin{table}[htb!]
  \footnotesize
  \setlength{\tabcolsep}{1.6pt}
  \renewcommand{\arraystretch}{1.05}
  \newcommand{\ccg}{}
  \begin{tabular}{lcccccccccc}
    \toprule
    \texttt{Lower Bound} &0.9 & 0.8& 0.7& 0.6& 0.5&0.4&0.3&0.2&0.1&0.05 \\
    \midrule
    Avg. Iter./Frame &281 &253 &225 &199 &172 &147 &123 &102&78&72\\
    Mapping/Frame [s] &9.22 &8.30	&7.39 &6.53 & 5.65& 4.83 &4.04 &3.35&2.56&2.36 \\
    \bottomrule
  \end{tabular}
  \caption{\textbf{Adaptive Mapping Ablation.} For various choices of the lower bound, Average Mapping Iterations, and Per-frame Mapping Runtime using adaptive iterations.} 
  \label{tab:ablation-min-iter-ratio}
\end{table}

\begin{figure}[tb]
\centering
{\footnotesize
\setlength{\tabcolsep}{1pt}
\renewcommand{\arraystretch}{1}
\newcommand{\sz}{0.5}
\begin{tabular}{cc}
\begin{subfigure}{0.50\linewidth}
\includegraphics[width=\linewidth]{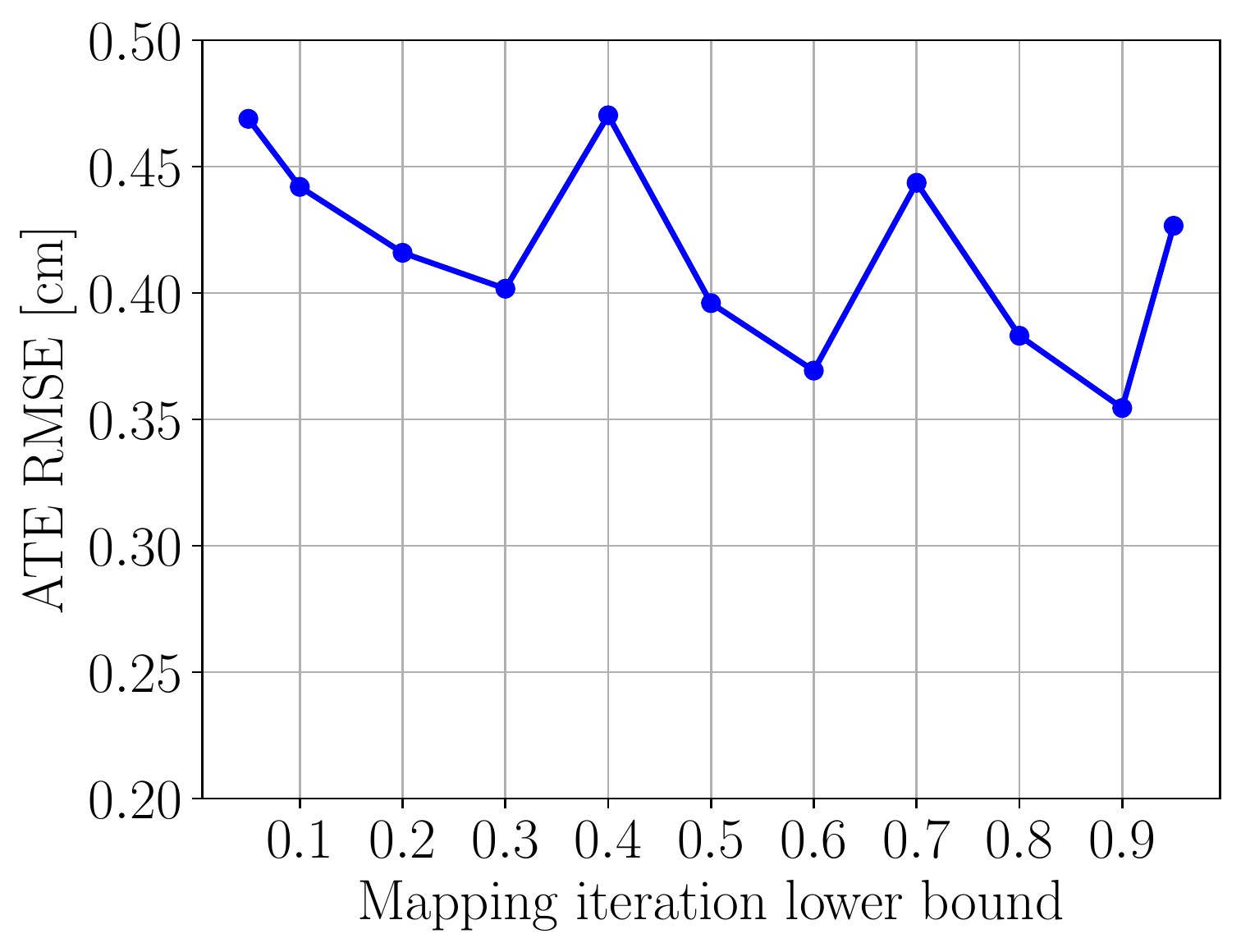}
\subcaption{ATE RMSE}
\label{fig:min_iter_ratio_a}
\end{subfigure}
&
\begin{subfigure}{0.50\linewidth}
\includegraphics[width=\linewidth]{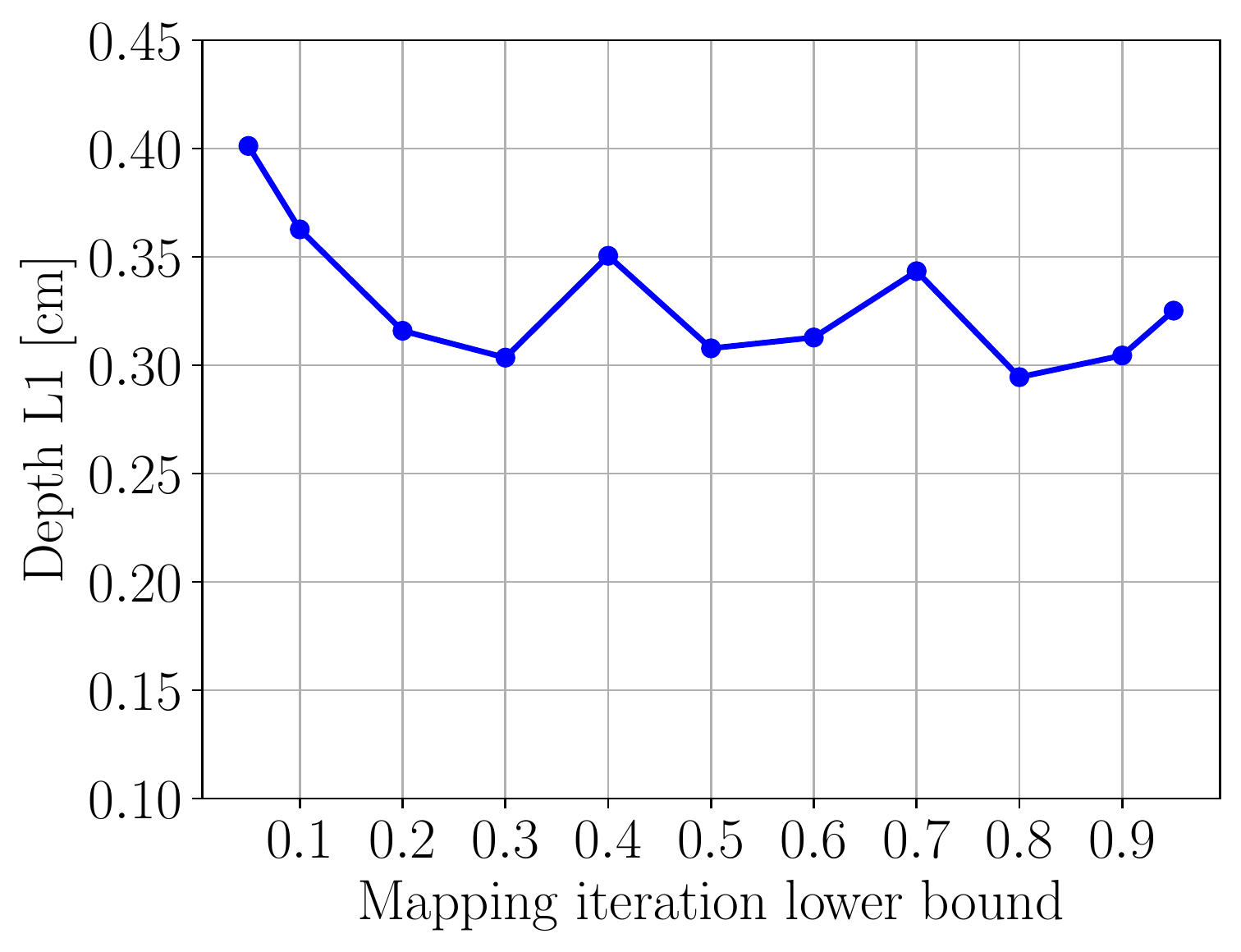}
\subcaption{Depth L1}
\label{fig:min_iter_ratio_b}
\end{subfigure}
\\[3mm]
\begin{subfigure}{0.50\linewidth}
\includegraphics[width=\linewidth]{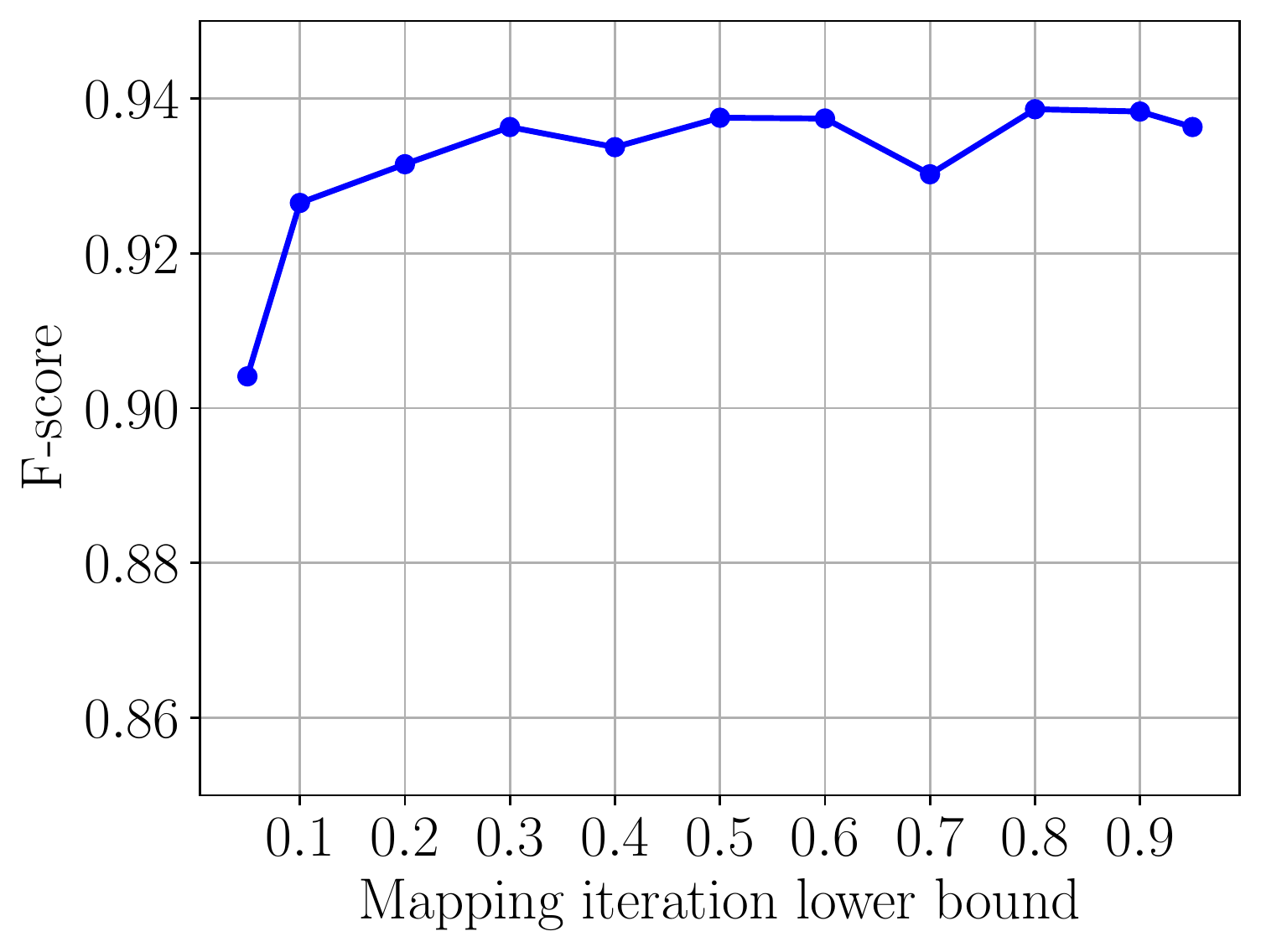}
\subcaption{F-score}
\label{fig:min_iter_ratio_c}
\end{subfigure}
&
\begin{subfigure}{0.50\linewidth}
\includegraphics[width=\linewidth]{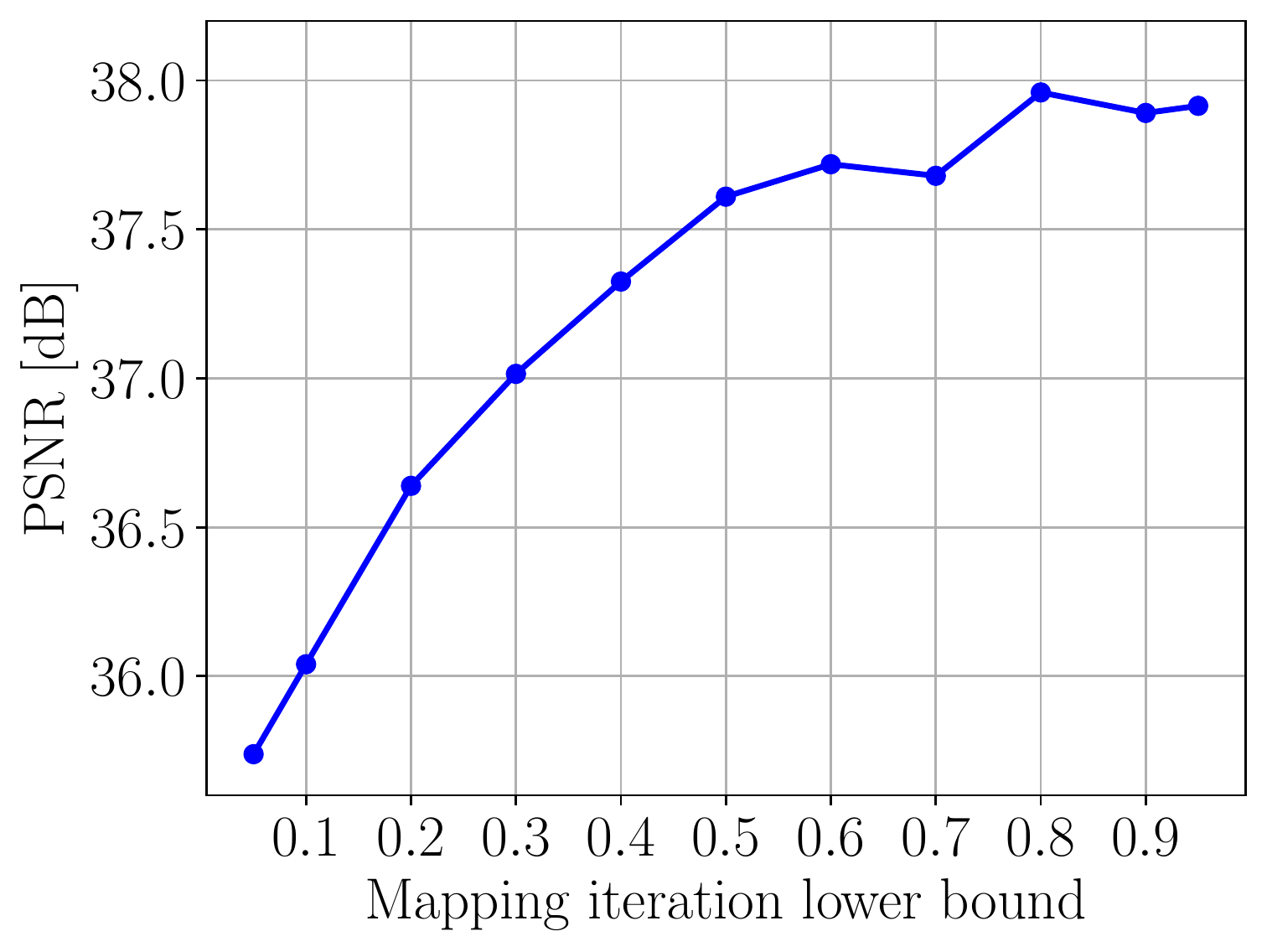}
\subcaption{PSNR}
\label{fig:min_iter_ratio_d}
\end{subfigure}
\end{tabular}
}
\vspace{2mm}
\caption{\textbf{Adaptive Mapping Ablation.} We report the rendering, tracking and reconstruction accuracy for different lower bound values and find that only the rendering quality is marginally worse as fewer mapping iterations are used.}
\label{fig:ablation_min_iter_ratio}
\end{figure}

\noindent \cref{fig:ablation_min_iter_ratio} summarizes the rendering, tracking and reconstruction metrics for different lower bound values. All metrics are virtually unchanged until the lower bound drops to 0.8. When the lower bound 0.05 is used, the per frame mapping speed is increased by 406\% compared to the default case, while the tracking accuracy only degrades by 10\%, depth L1 by 23\%, F-score by 3\% and PSNR by 6\%. This suggests the effectiveness of our adaptive mapping iteration strategy.

\boldparagraph{Additional ScanNet Results.} In \cref{tab:scannet}, we provide additional evaluation on four ScanNet scenes over the main paper and show competitive performance compared to the baseline methods. When taking the average over all scenes (the scenes in the main paper and the additional four scenes we select), we find that our method outperforms NICE-SLAM~\cite{zhu2022nice} and Vox-Fusion~\cite{yang2022vox}.

\begin{figure*}[ht]
\centering
{\footnotesize
\setlength{\tabcolsep}{5pt}
\renewcommand{\arraystretch}{2}
\newcommand{\sz}{0.26}
\begin{tabular}{lcccc}
\multirow{2}{*}[10pt]{\rotatebox[origin=c]{90}{\makecell{\bf Rendering Comparison on TUM-RGBD}}} & \rotatebox[origin=c]{90}{\texttt{freiburg2-xyz}} & 
\includegraphics[valign=c,width=\sz\linewidth]{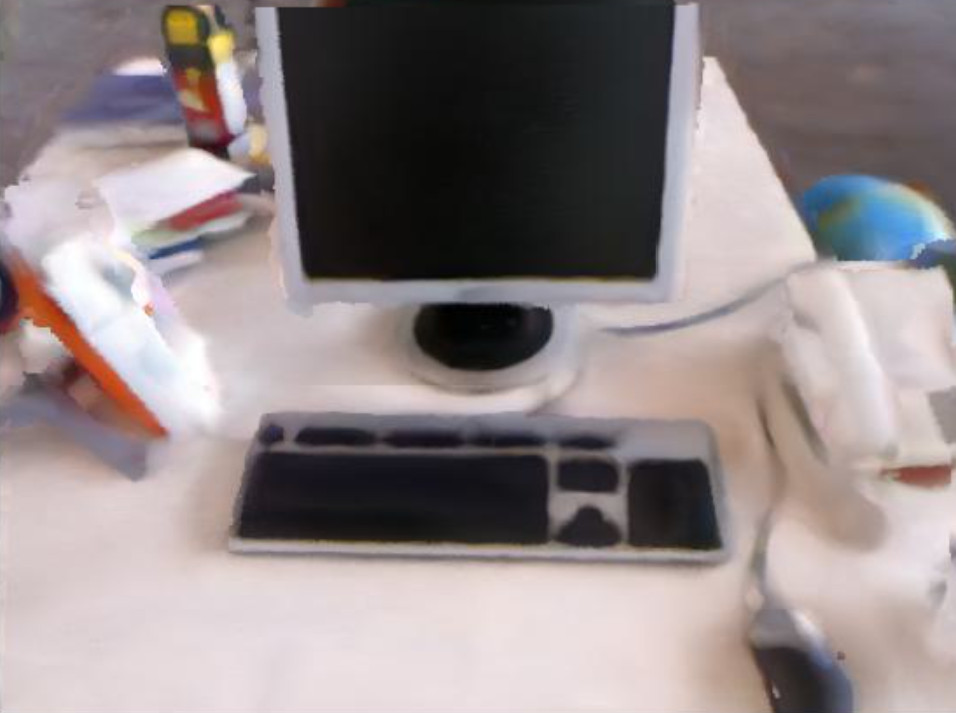} & 
\includegraphics[valign=c,width=\sz\linewidth]{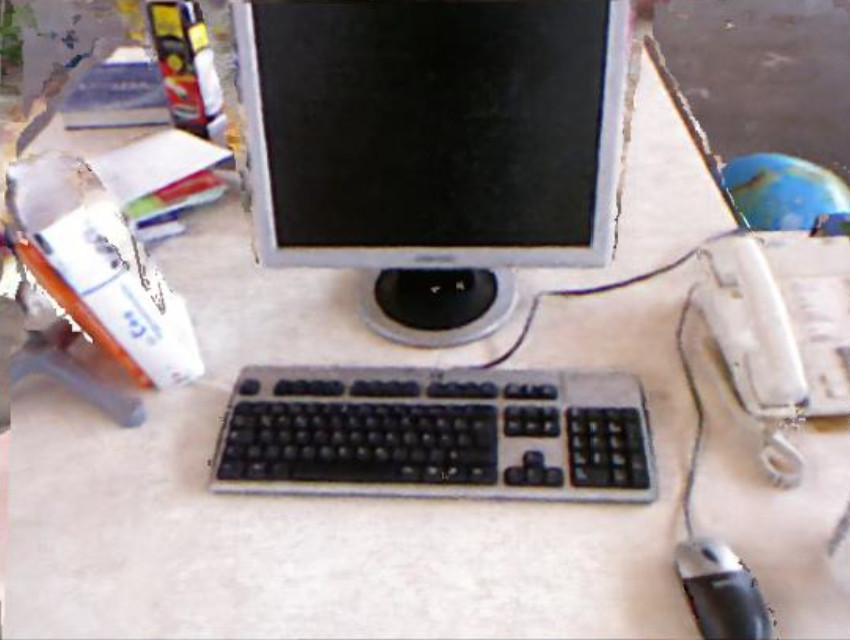} &
\includegraphics[valign=c,width=\sz\linewidth]{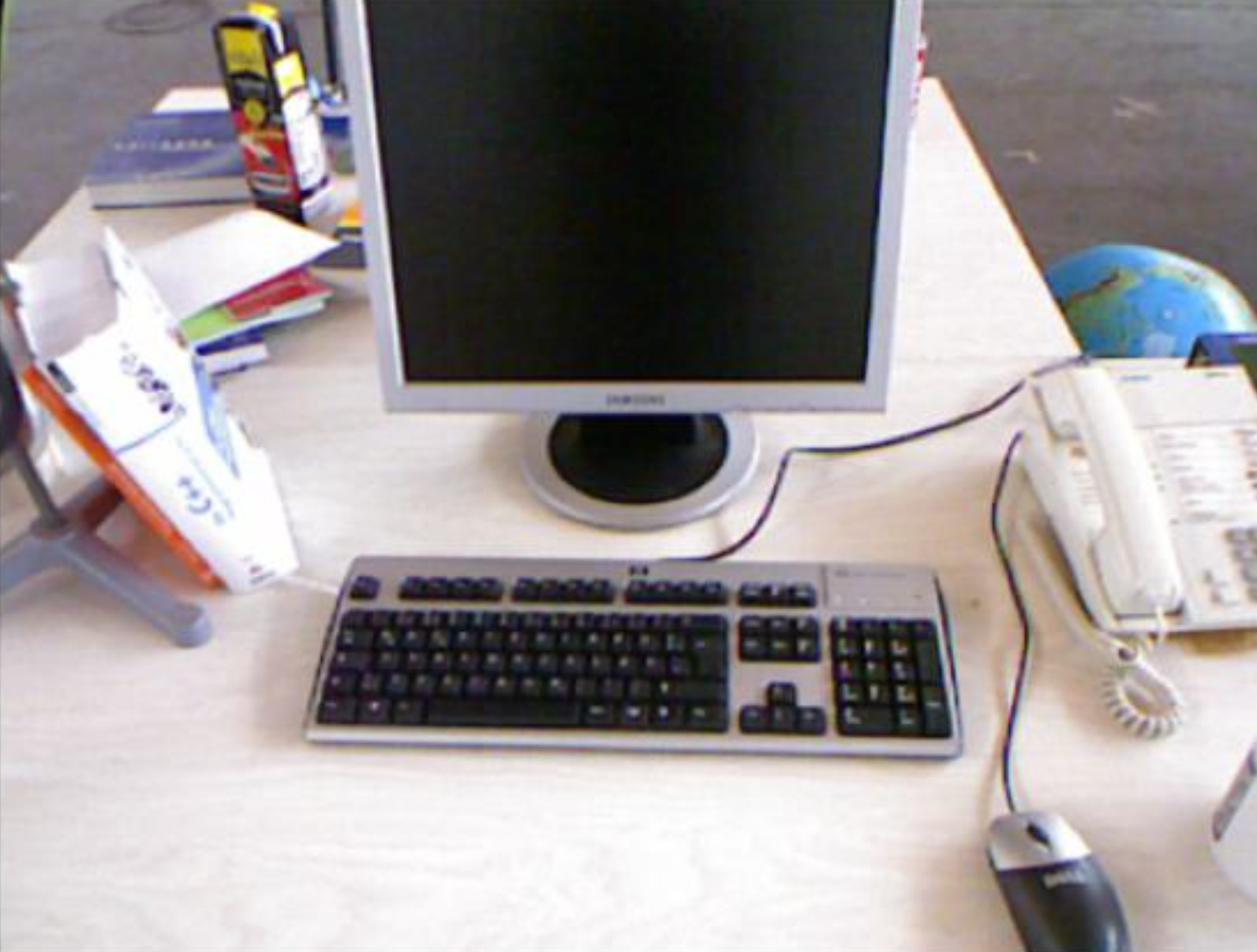} \\
& \rotatebox[origin=c]{90}{\texttt{freiburg3-office}} & 
\includegraphics[valign=c,width=\sz\linewidth]{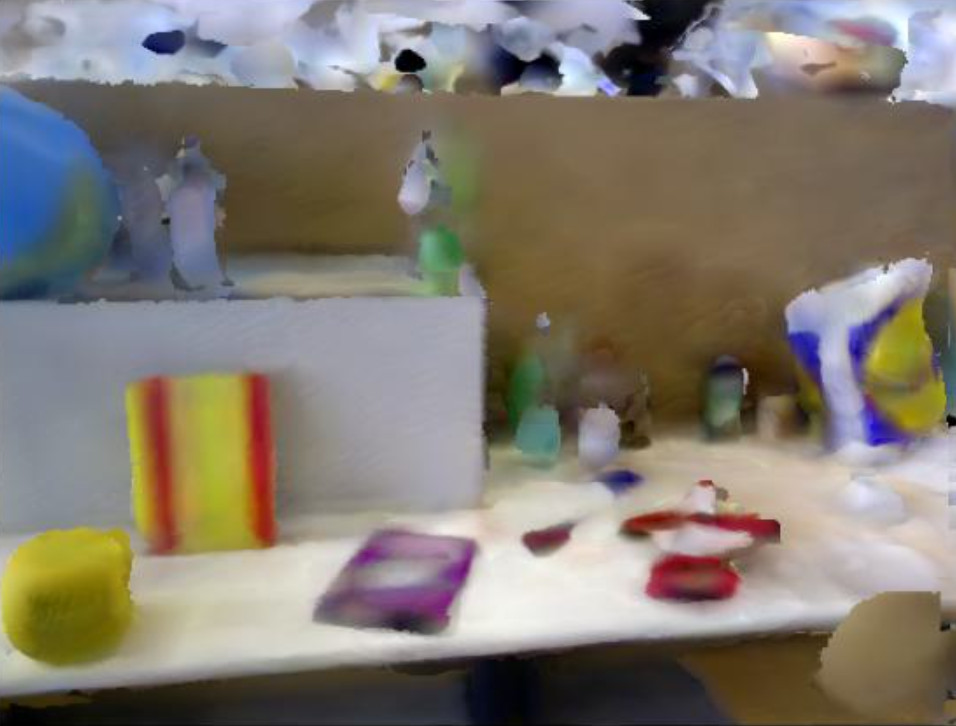} & 
\includegraphics[valign=c,width=\sz\linewidth]{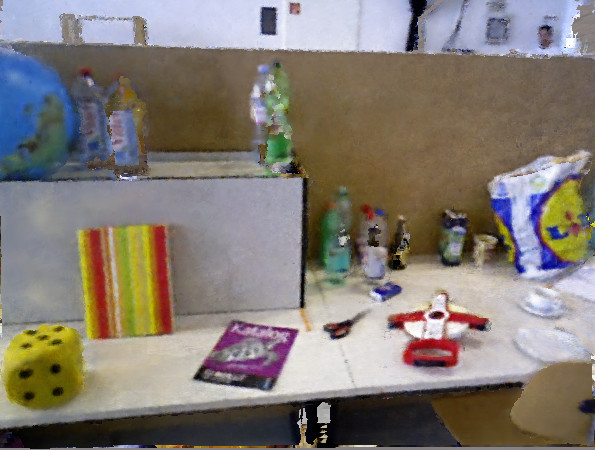} &
\includegraphics[valign=c,width=\sz\linewidth]{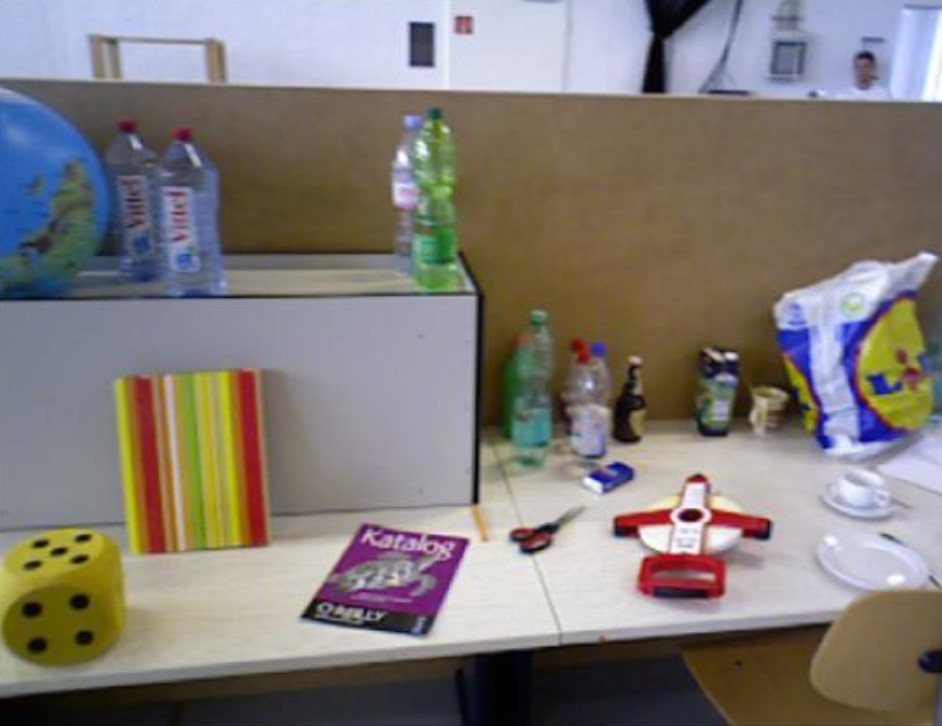} \\
& & NICE-SLAM~\cite{zhu2022nice} & \ours (ours) & Ground Truth \\
  \multirow{2}{*}[10pt]{\rotatebox[origin=c]{90}{\makecell{\bf Mesh-based Comparison on TUM-RGBD}}} & 
  \rotatebox[origin=c]{90}{\texttt{freiburg2-xyz}} & 
  \includegraphics[valign=c,width=\sz\linewidth]{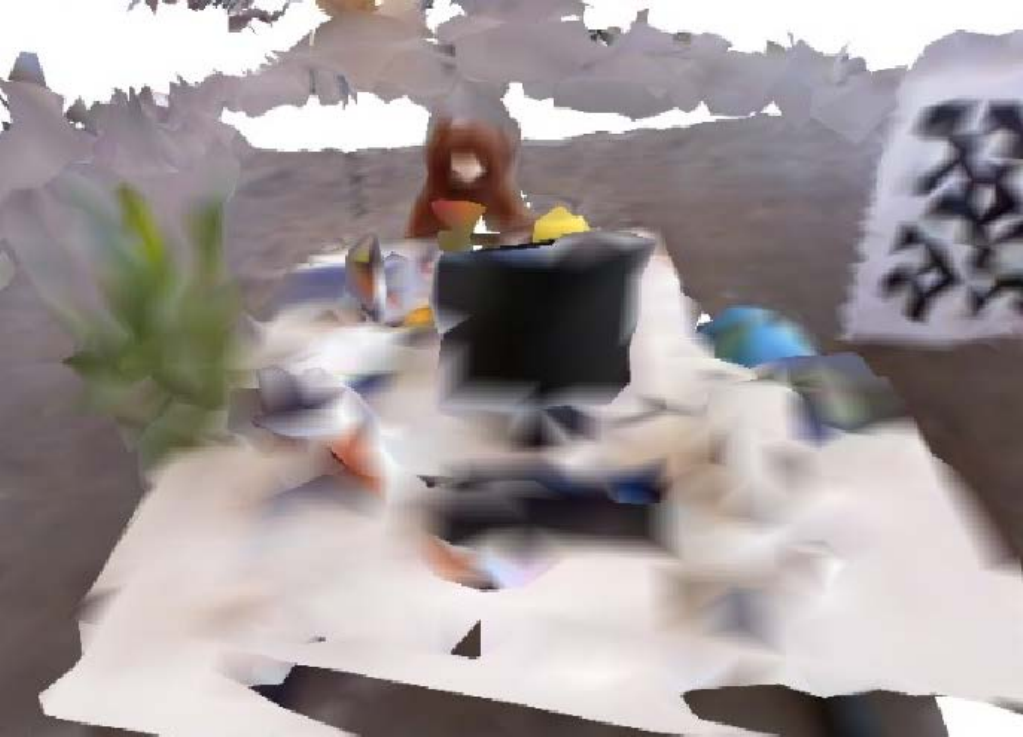} & 
  \includegraphics[valign=c,width=\sz\linewidth]{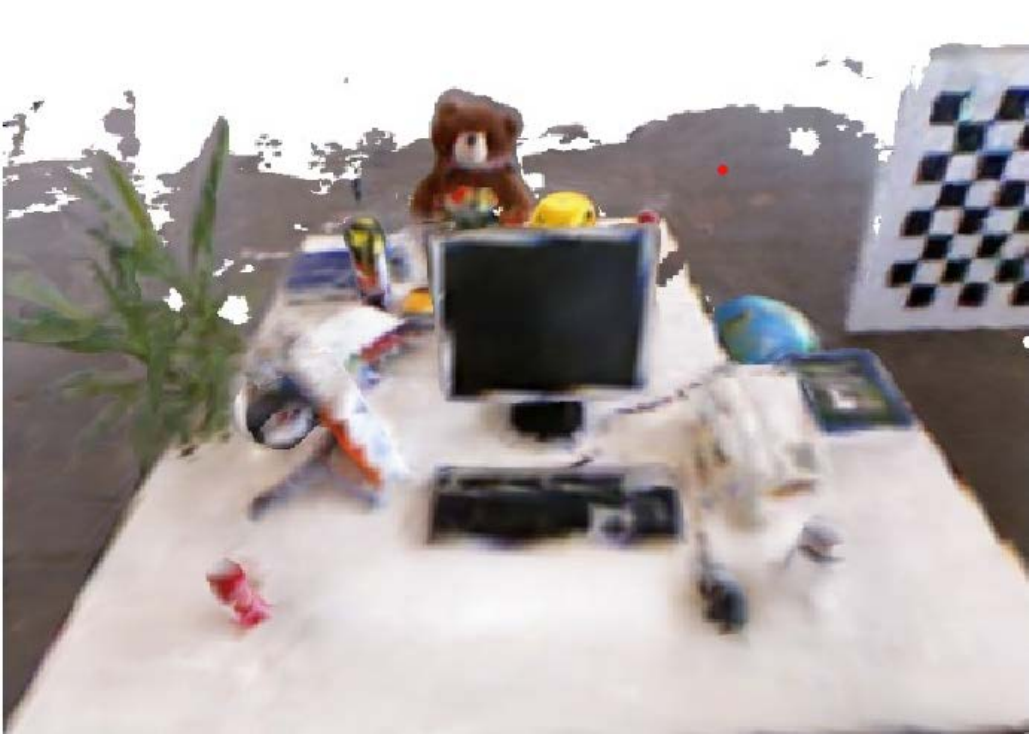} &
  \includegraphics[valign=c,width=\sz\linewidth]{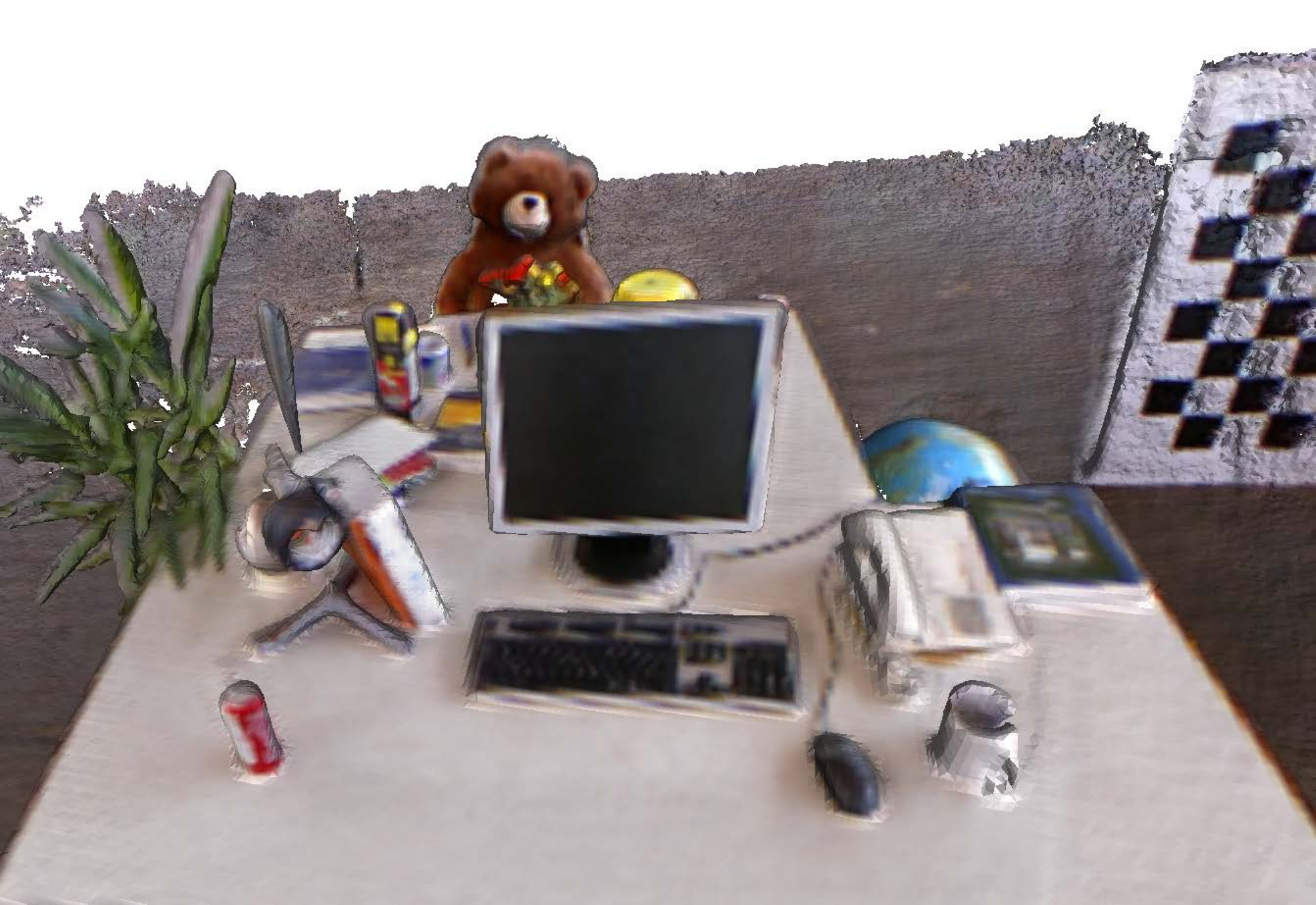} \\
 & \rotatebox[origin=c]{90}{\texttt{freiburg3-office}} & 
  \includegraphics[valign=c,width=\sz\linewidth]{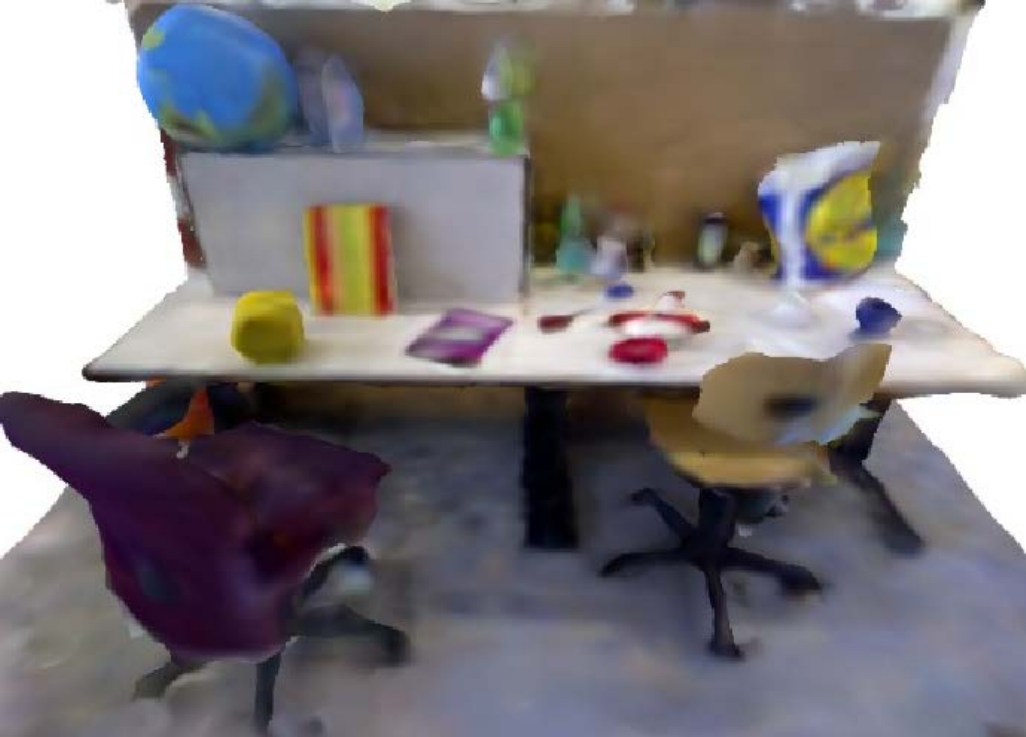} & 
  \includegraphics[valign=c,width=\sz\linewidth]{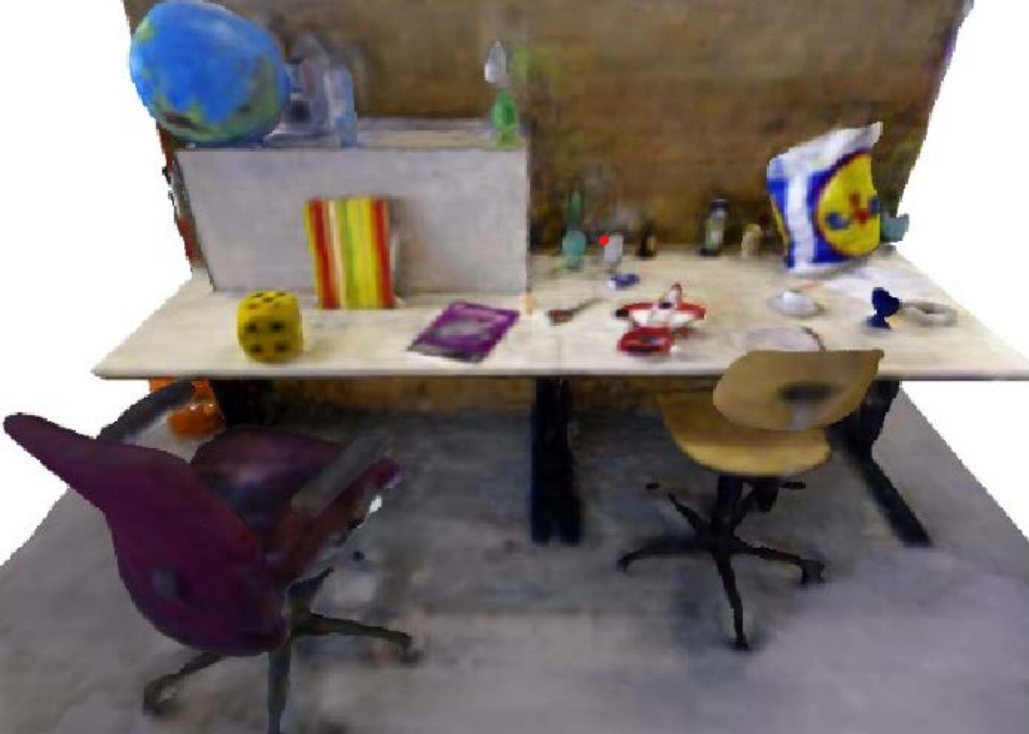} &
  \includegraphics[valign=c,width=\sz\linewidth]{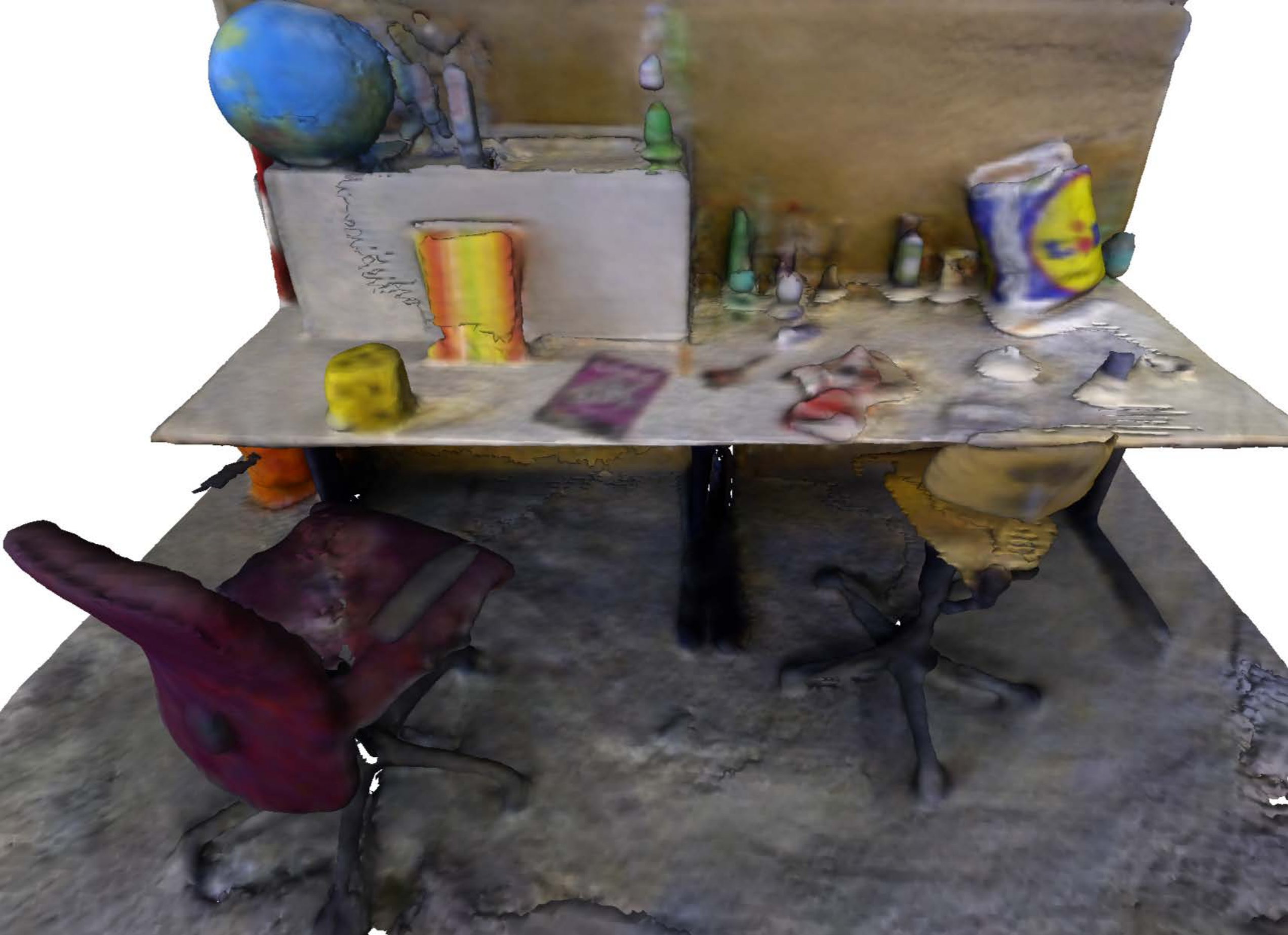} \\
 & & NICE-SLAM~\cite{zhu2022nice} & ESLAM~\cite{mahdi2022eslam} &\ours (ours) \\
& \rotatebox[origin=c]{90}{\texttt{Scene 0169}} & 
\includegraphics[trim={0 0 0 5cm},clip, valign=c,width=\sz\linewidth]{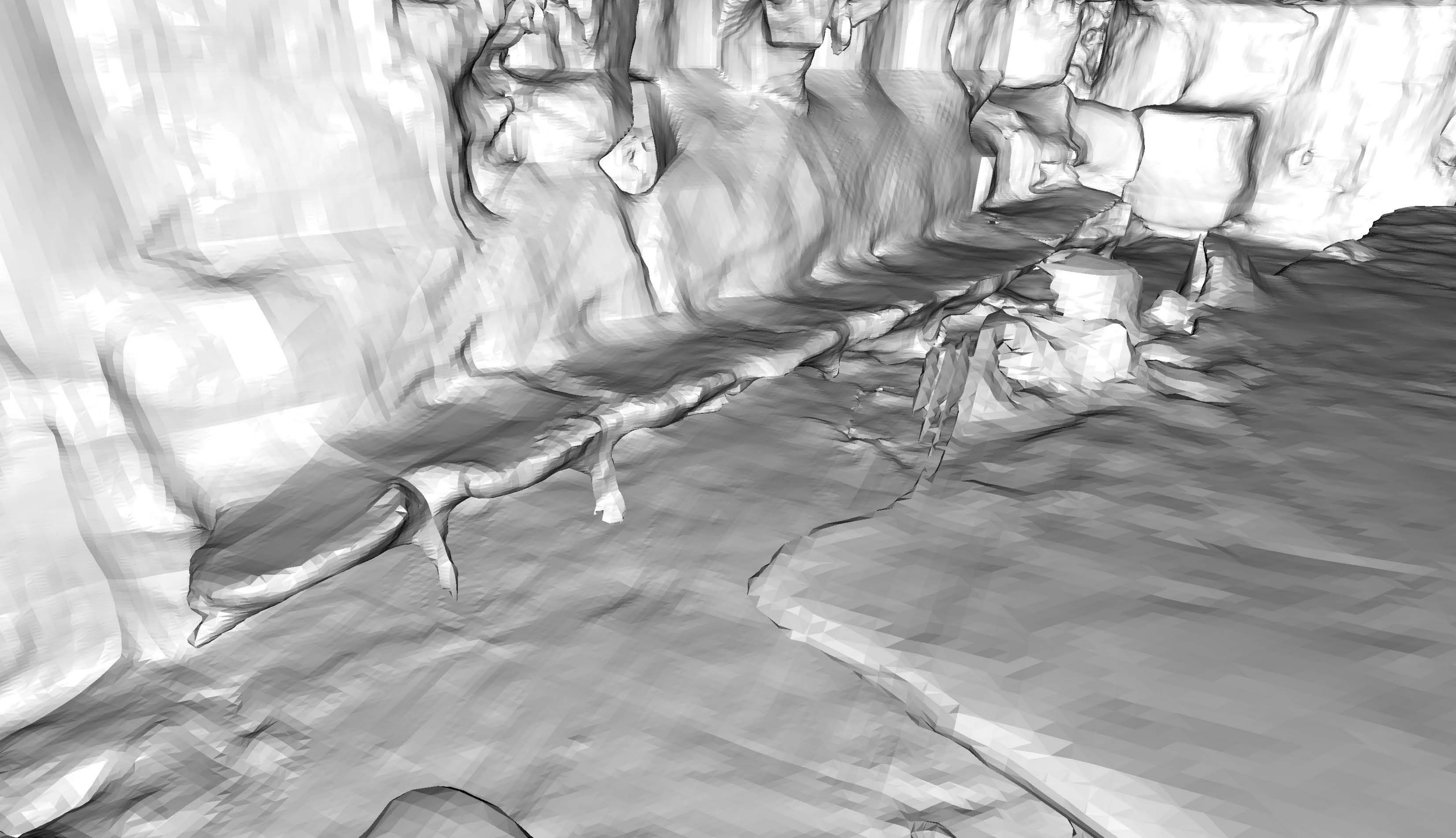} & 
\includegraphics[trim={0 0 0 5cm},clip,valign=c,width=\sz\linewidth]{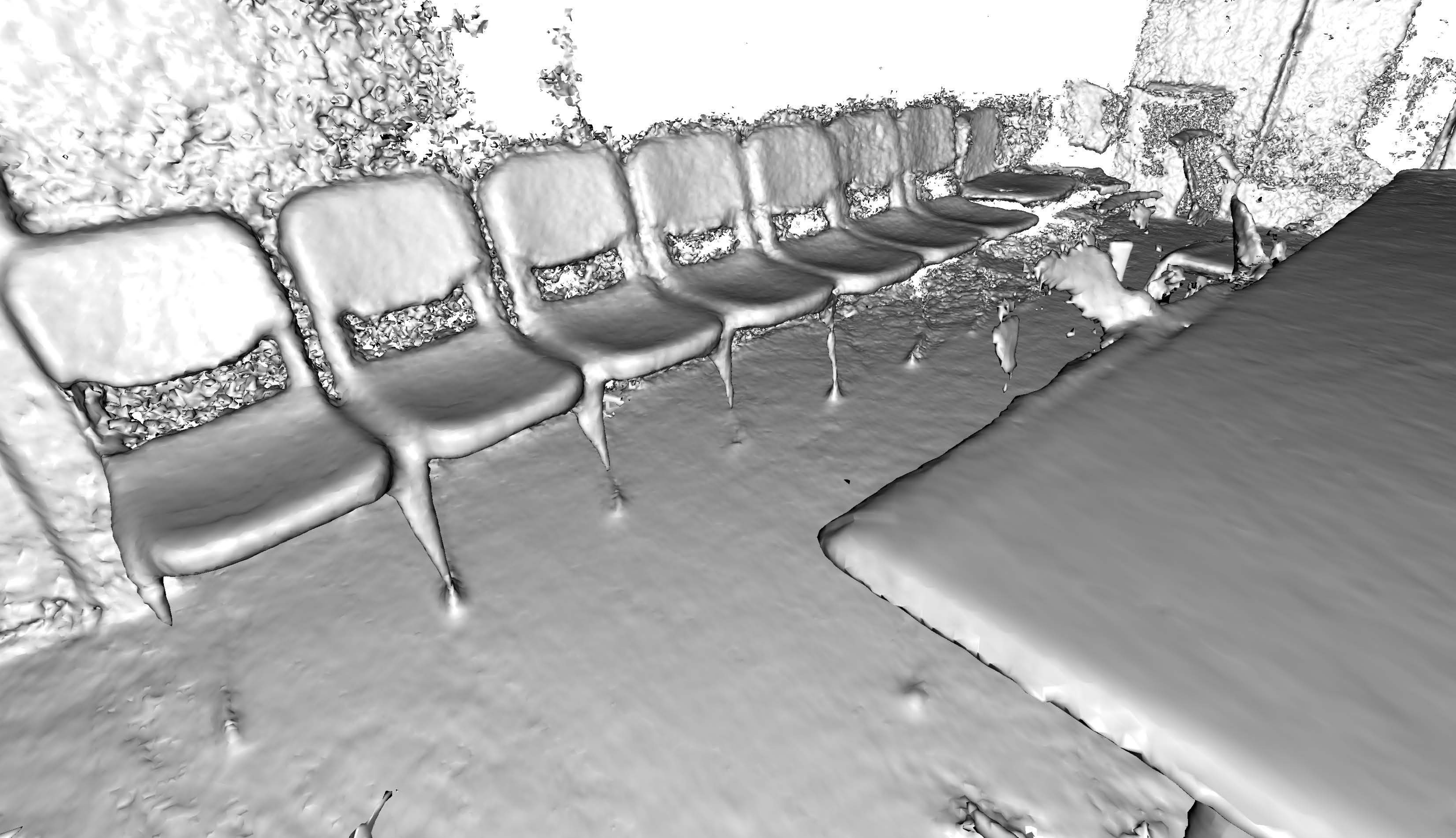} &
\includegraphics[trim={0 0 0 5cm},clip,valign=c,width=\sz\linewidth]{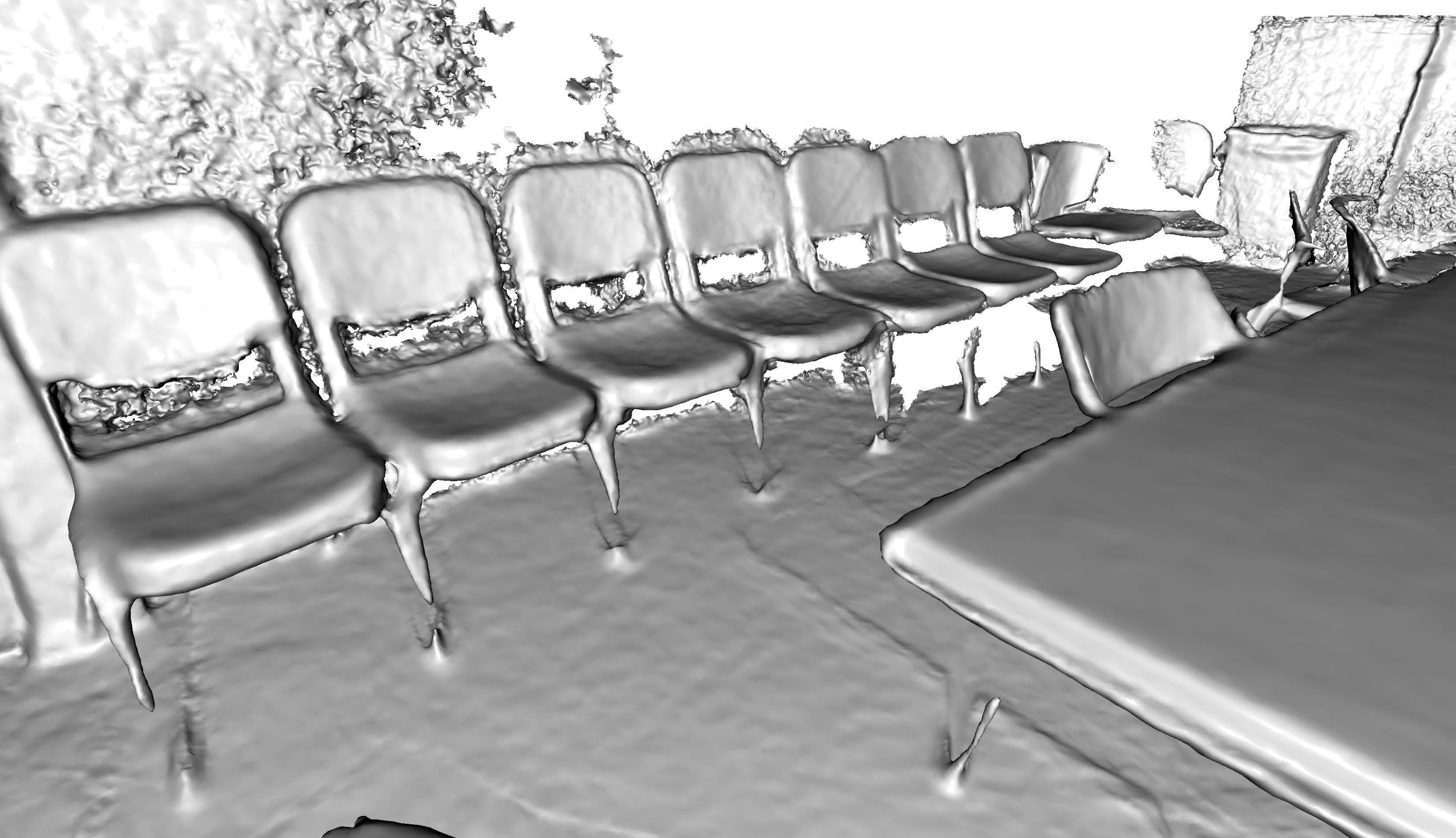} \\
\multirow{2}{*}[65pt]{\rotatebox[origin=c]{90}{\makecell{\bf Reconstruction on ScanNet}}} & \rotatebox[origin=c]{90}{\texttt{Scene 0207}} & 
\includegraphics[valign=c,width=\sz\linewidth]{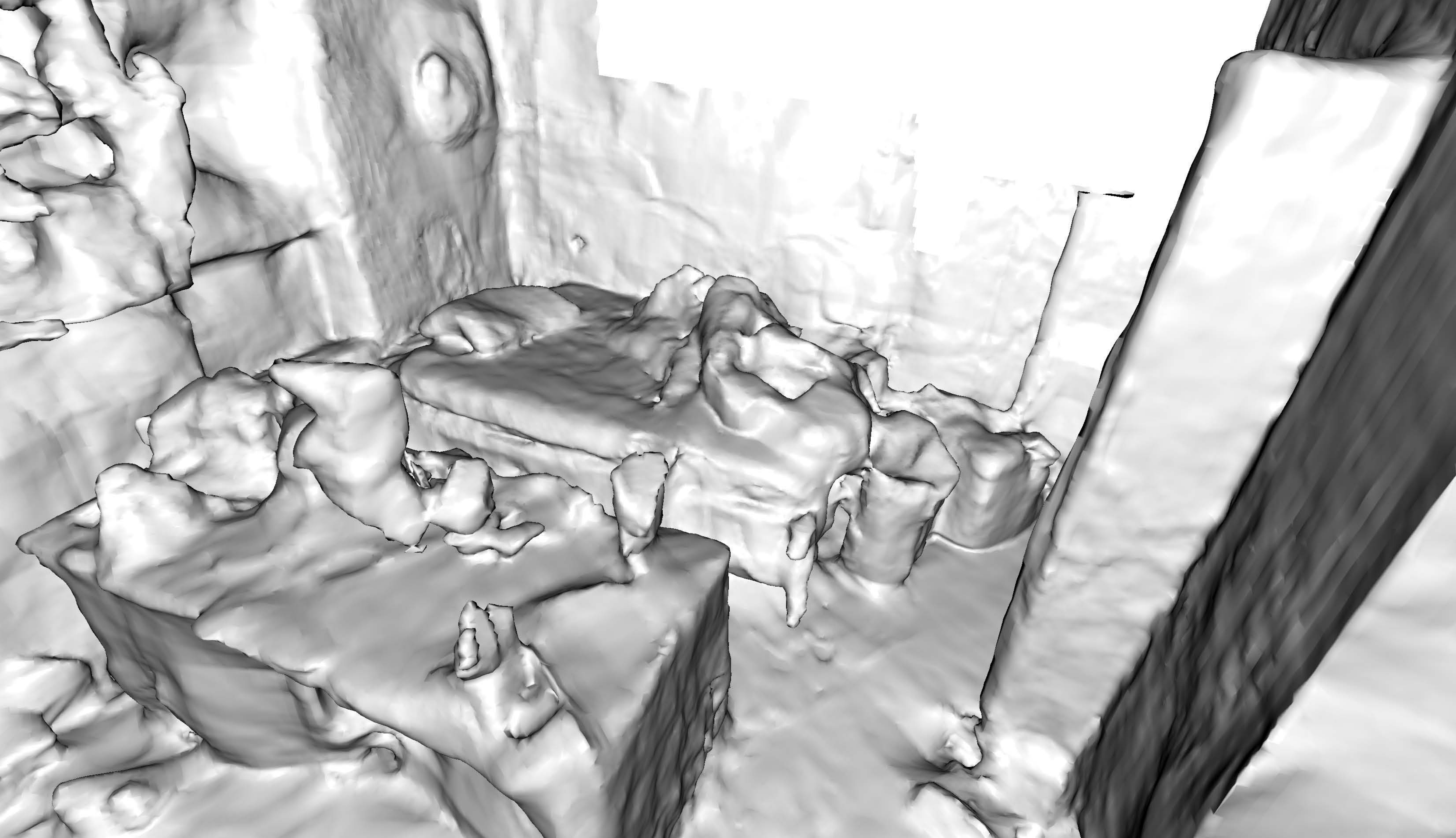} & 
\includegraphics[valign=c,width=\sz\linewidth]{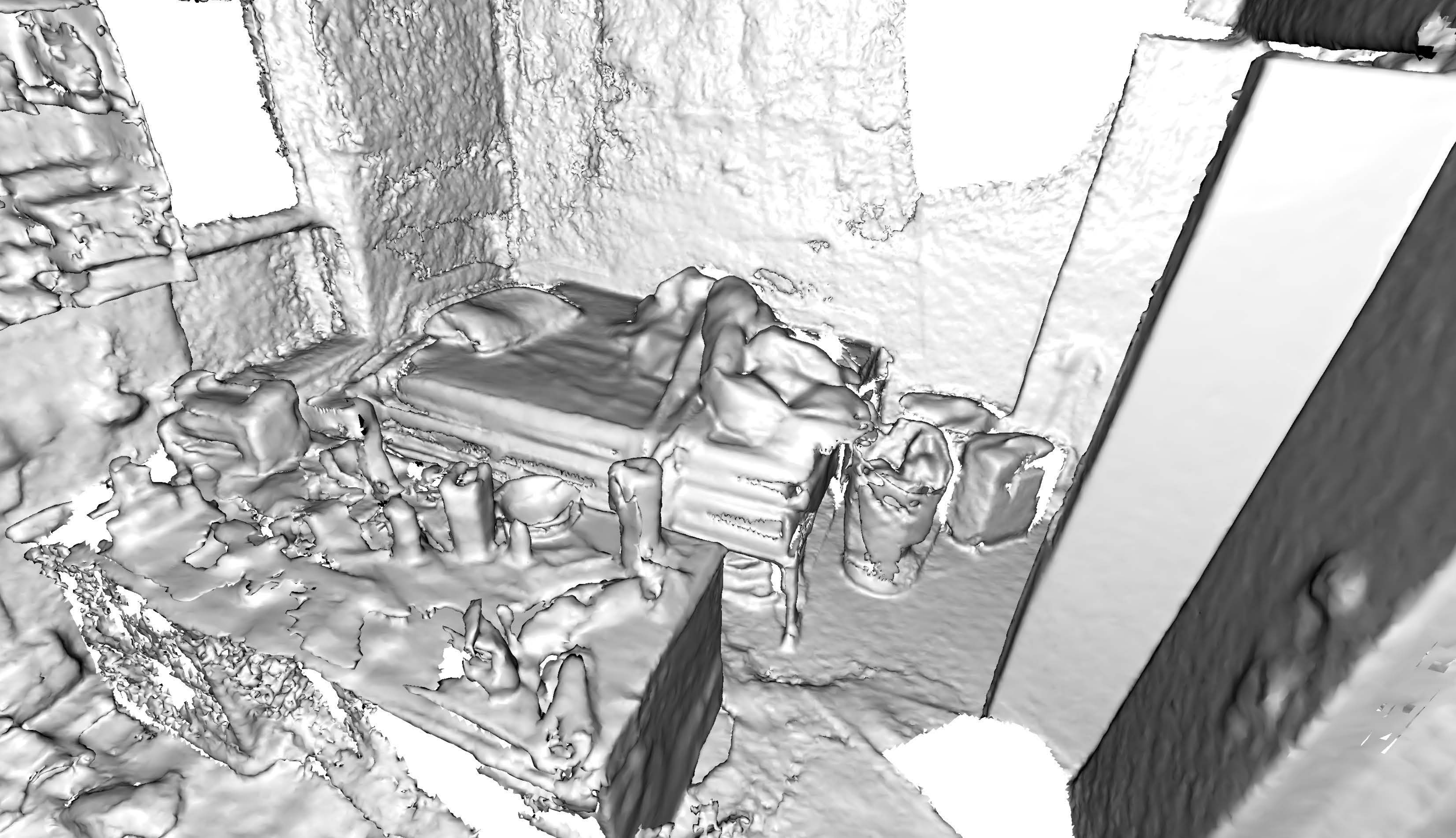} &
\includegraphics[valign=c,width=\sz\linewidth]{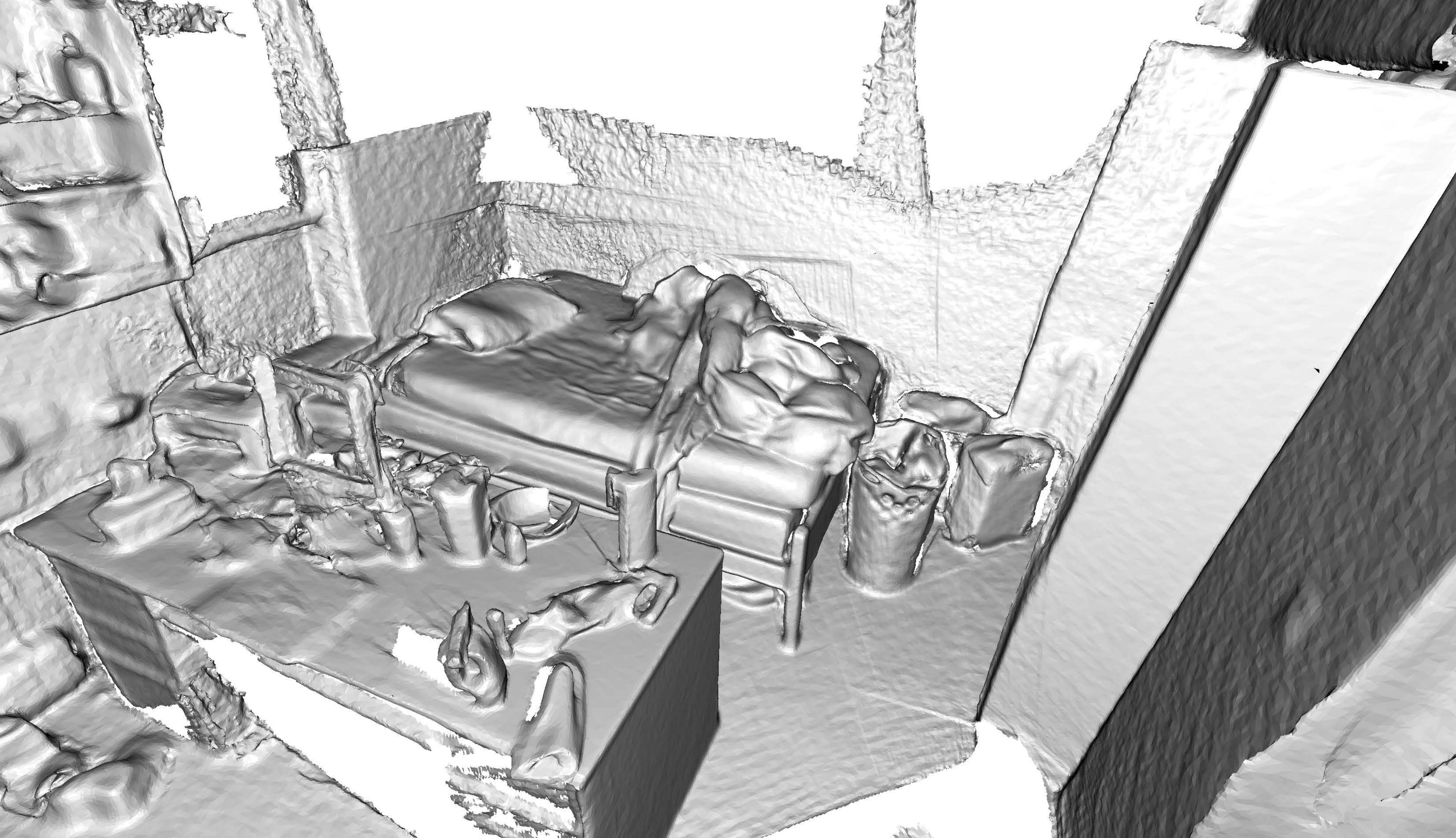} \\
& & NICE-SLAM~\cite{zhu2022nice} & \ours (ours) & Ground Truth
\end{tabular}
}
\caption{\textbf{Rendering and Reconstruction Comparisons}. We showcase, from top to bottom, the rendering performance on TUM-RGBD, the colored mesh and the phong shaded mesh. The results suggest that our method can produce high quality renderings, textured and untextured meshes.}
\label{fig:render_mesh_reconstruction}
\vspace{7pt}
\end{figure*}

\boldparagraph{Qualitative Results on TUM-RGBD and ScanNet.} We compare our method to NICE-SLAM~\cite{zhu2022nice} and ESLAM~\cite{mahdi2022eslam} in \cref{fig:render_mesh_reconstruction}. In the cases where ground truth is available, we also compare to that. We showcase, from top to bottom, the rendering performance on TUM-RGBD, the colored mesh and the phong shaded mesh. The results suggest that our method can produce high quality renderings, textured and untextured meshes.

\begin{table*}[tb]
\centering
\setlength{\tabcolsep}{2pt}
\resizebox{\linewidth}{!}
{
\begin{tabular}{llllllllllll}
\toprule
Method \textbackslash{} Scene & \texttt{0000\_00} & \texttt{0025\_02} & \texttt{0059\_00} & \texttt{0062\_00} & \texttt{0103\_00} & \texttt{0106\_00} & \texttt{0126\_00}  & \texttt{0169\_00} & \texttt{0181\_00} & \texttt{0207\_00} & Avg. \\ \midrule
NICE-SLAM~\cite{zhu2022nice} & \nd12.00 & \rd 10.11 &\nd 14.00 & \nd 4.59 & \fs \textbf{4.94} & \fs7.90 &\rd 21.80 & \fs 10.90 & \fs \textbf{13.40} & \fs6.20 & \nd 10.58  \\
Vox-Fusion$^*$~\cite{yang2022vox} & \rd 68.84 (16.55) & \nd 8.54  & \rd 24.18 &\rd 7.96 & \nd 5.26 & \nd 8.41 &\fs \textbf{5.77} &  \rd27.28 &  \rd 23.30 & \nd 9.41 &  \rd 18.90 (13.67) \\
\textbf{\ours (Ours)} & \fs 10.24 & \fs 8.05  & \fs 7.81  & \fs 3.75 & \rd 7.79 & \rd 8.65  & \nd 8.10  & \nd 22.16  & \nd 14.77 & \rd 9.54 & \fs 10.08 \\ 
 \bottomrule 
\end{tabular}
}
\caption{\textbf{ScanNet Tracking} We report the ATE RMSE ($\downarrow$ [cm]) as the average over three runs. For failed runs we report the average of only successful runs in parentheses. All methods work differently well on various scenes, but our method performs better on average. Best results are highlighted as \colorbox{colorFst}{\bf first}, \colorbox{colorSnd}{second}, and \colorbox{colorTrd}{third}.}
\label{tab:scannet}
\vspace{8pt}
\end{table*}

\begin{figure*}[tb]
\centering
{\footnotesize
\setlength{\tabcolsep}{1pt}
\renewcommand{\arraystretch}{1}
\newcommand{\sz}{0.244}
\begin{tabular}{ccccc}
\rotatebox[origin=c]{90}{\texttt{Office 2}} & 
\includegraphics[valign=c,width=\sz\linewidth]{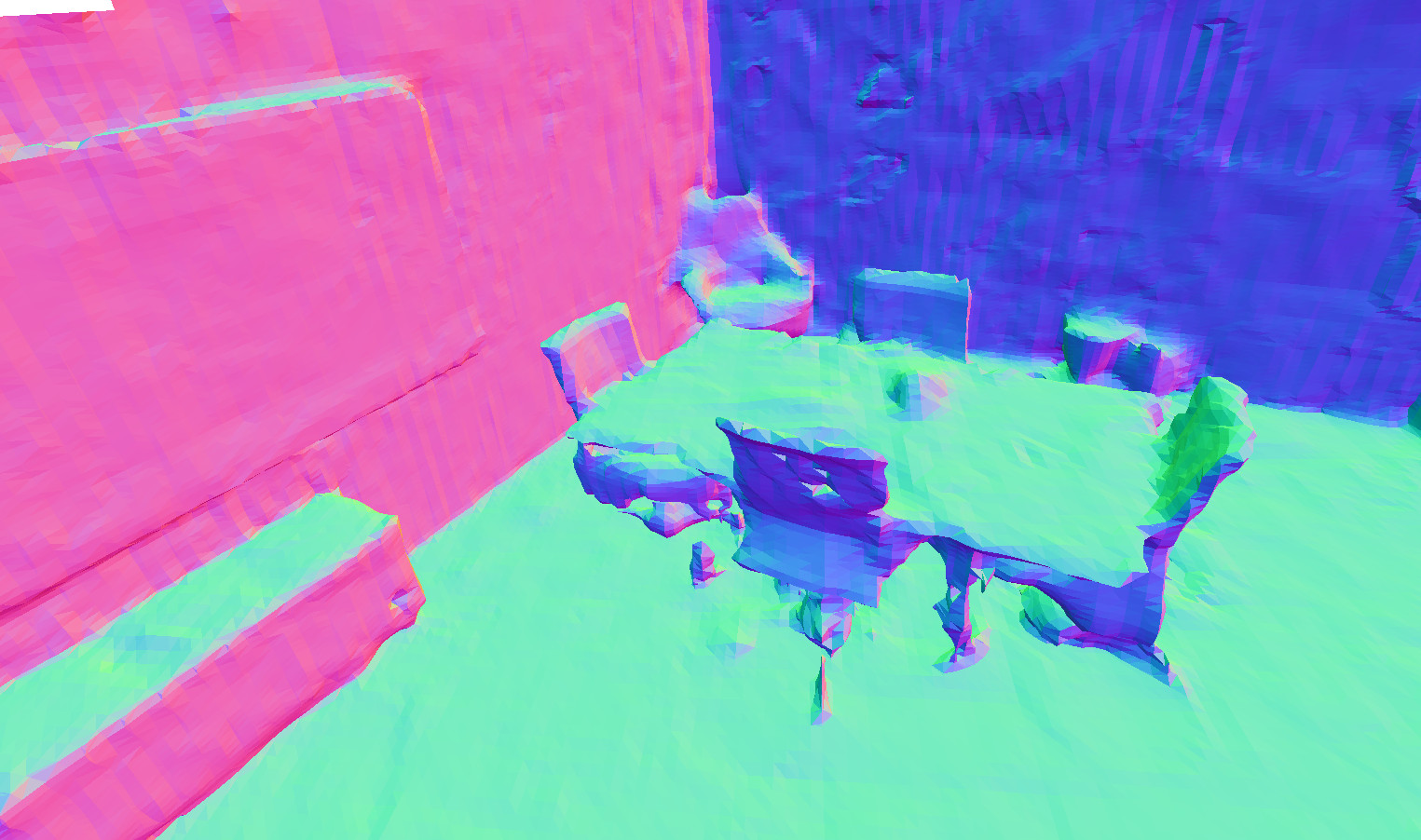} & 
\includegraphics[valign=c,width=\sz\linewidth]{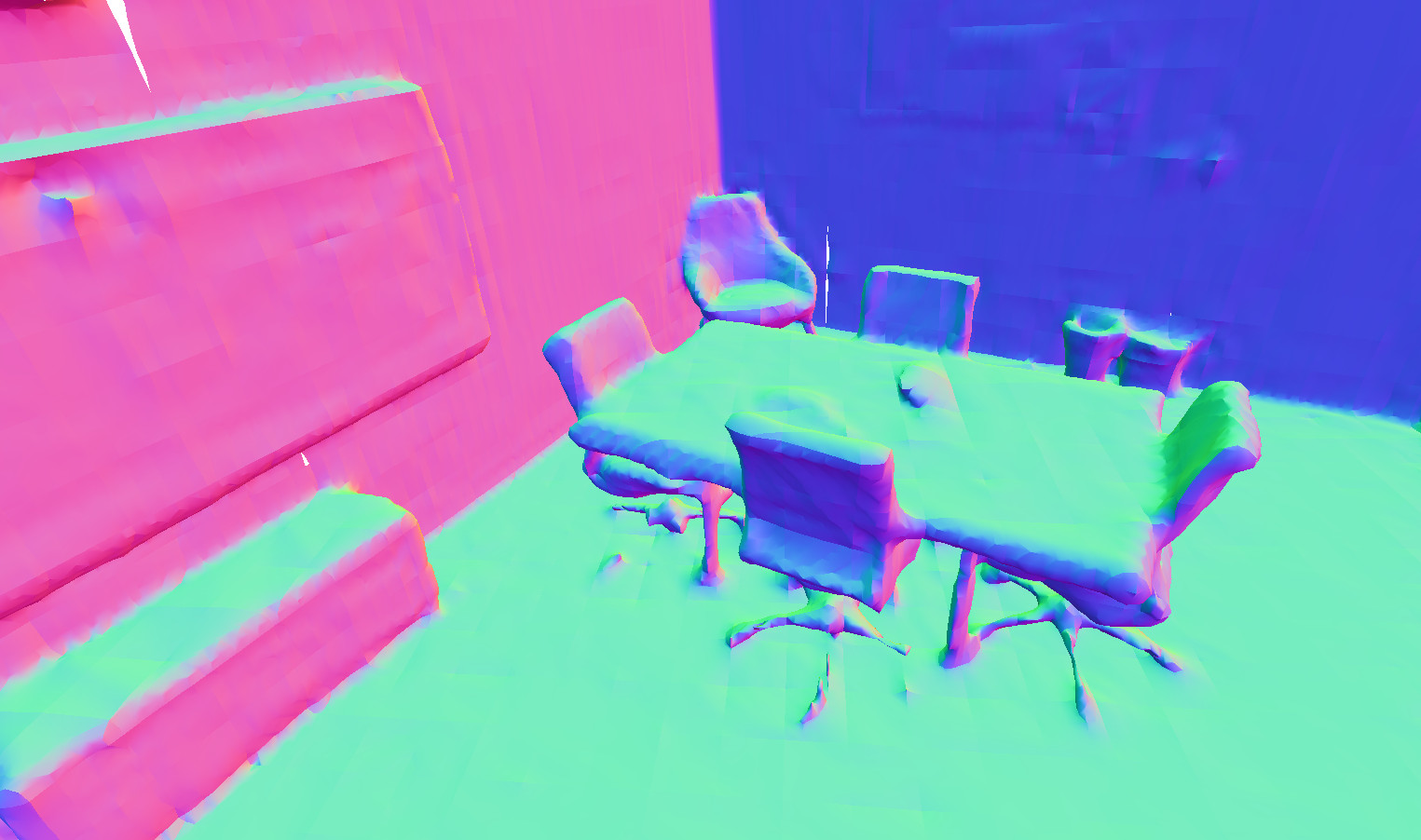} &
\includegraphics[valign=c,width=\sz\linewidth]{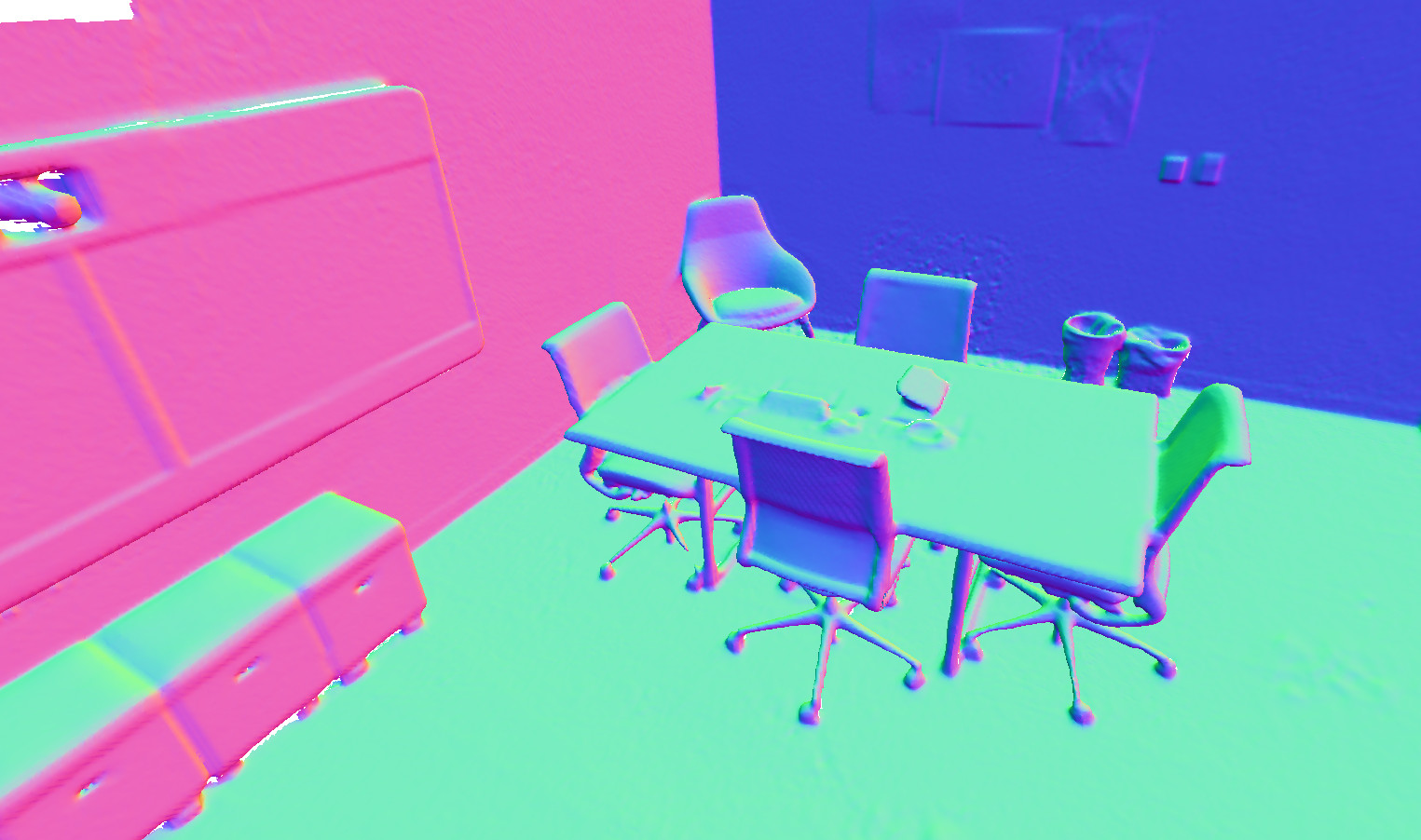} &
\includegraphics[valign=c,width=\sz\linewidth]{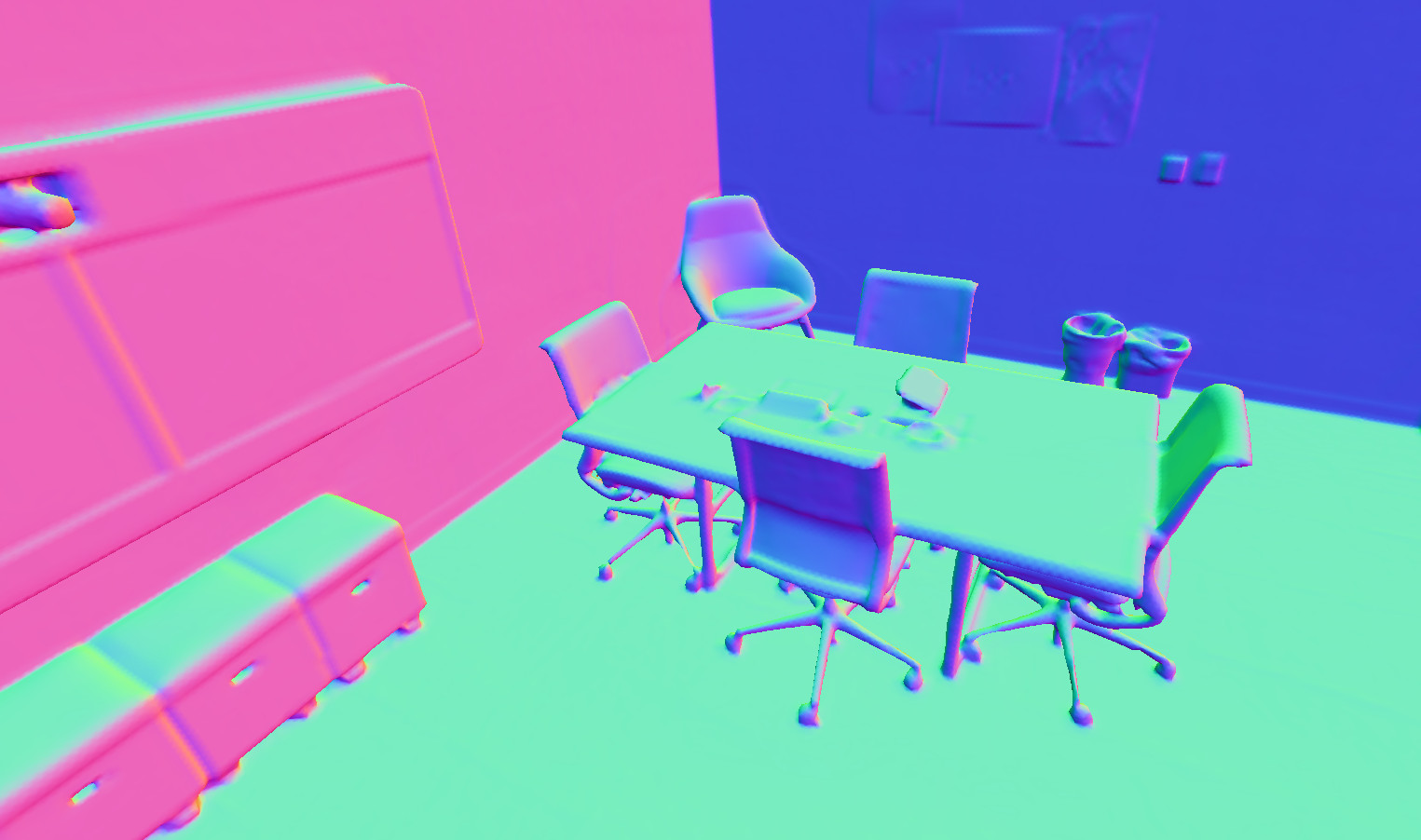} \\
\rotatebox[origin=c]{90}{\texttt{Office 4}} & 
\includegraphics[valign=c,width=\sz\linewidth]{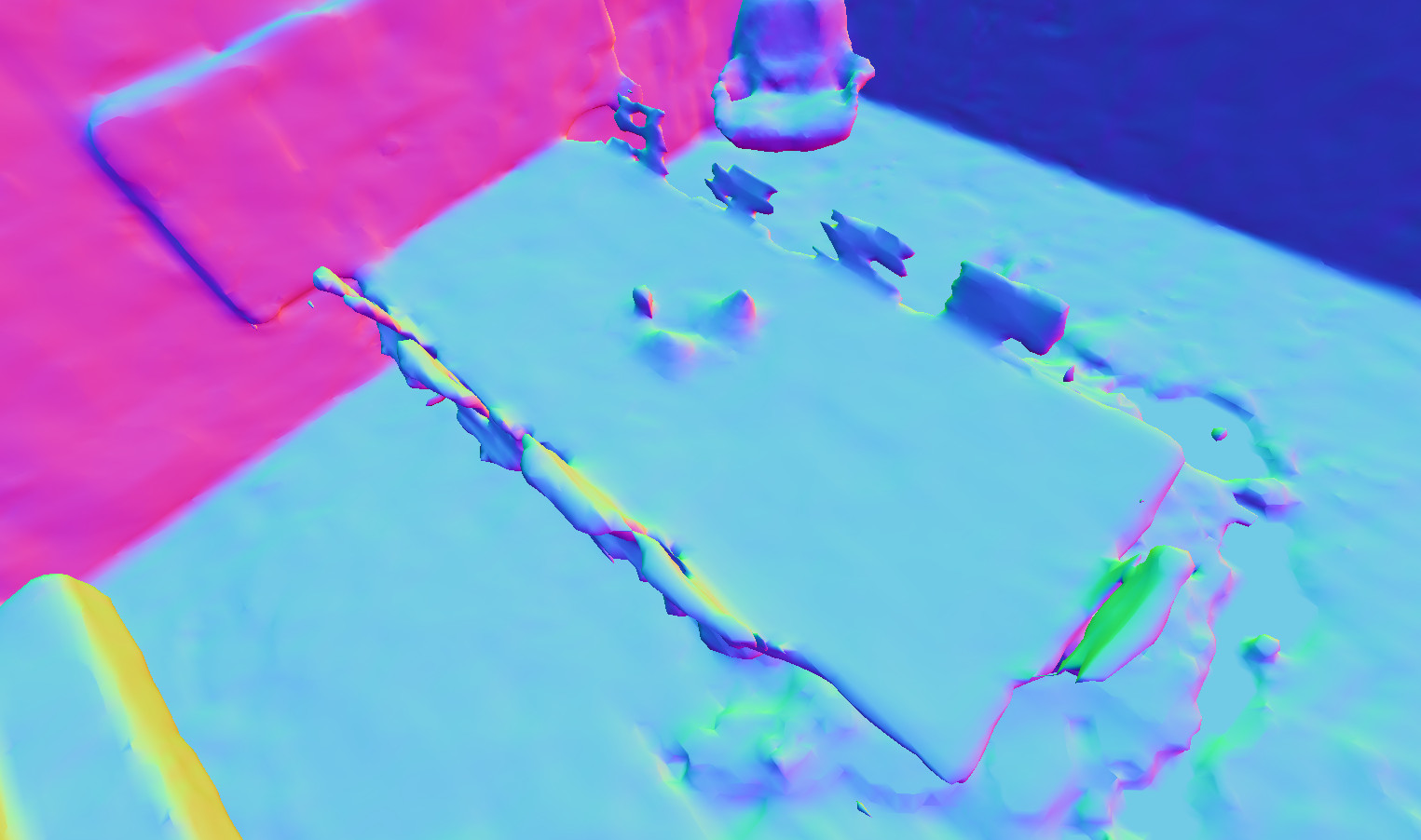} & 
\includegraphics[valign=c,width=\sz\linewidth]{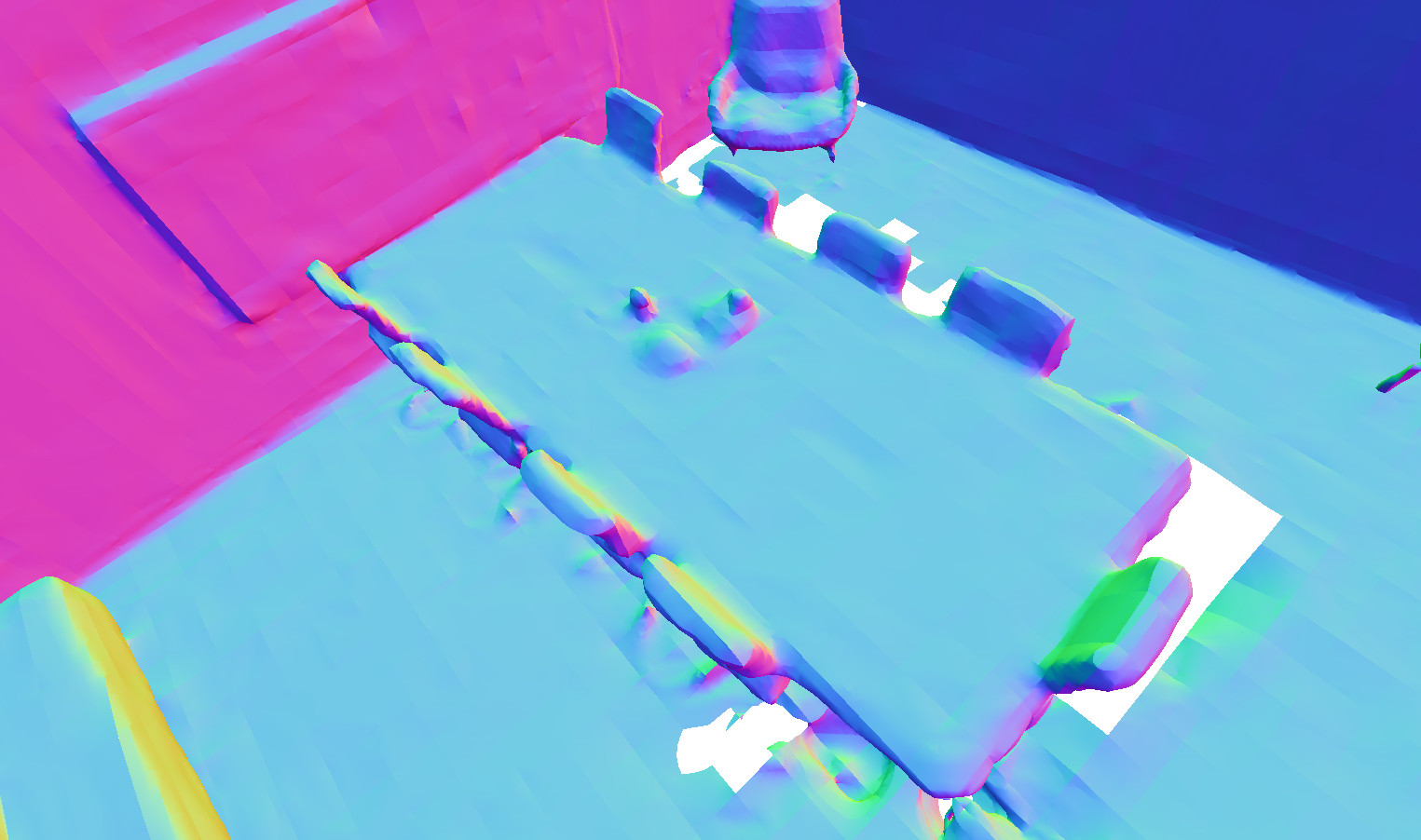} &
\includegraphics[valign=c,width=\sz\linewidth]{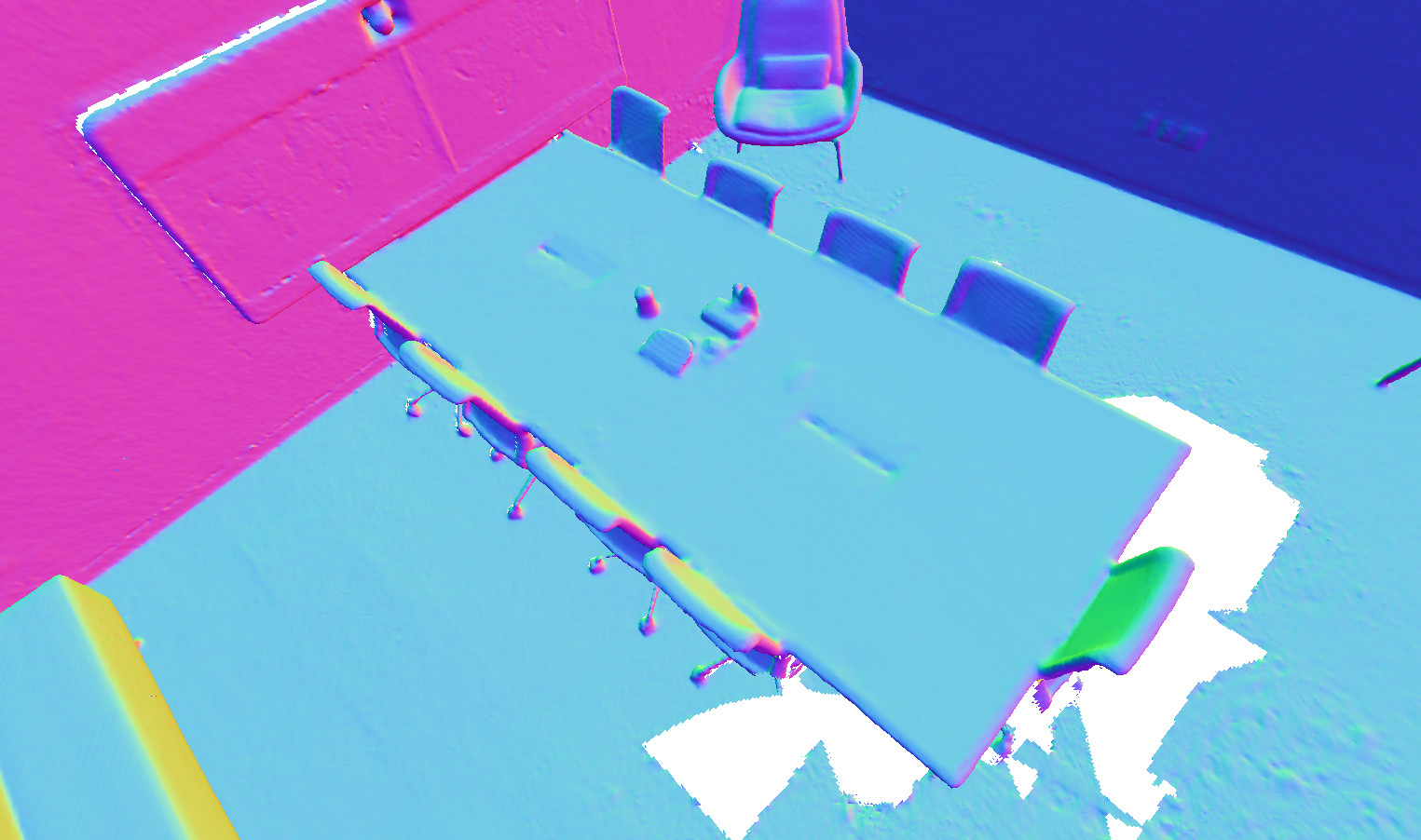} &
\includegraphics[valign=c,width=\sz\linewidth]{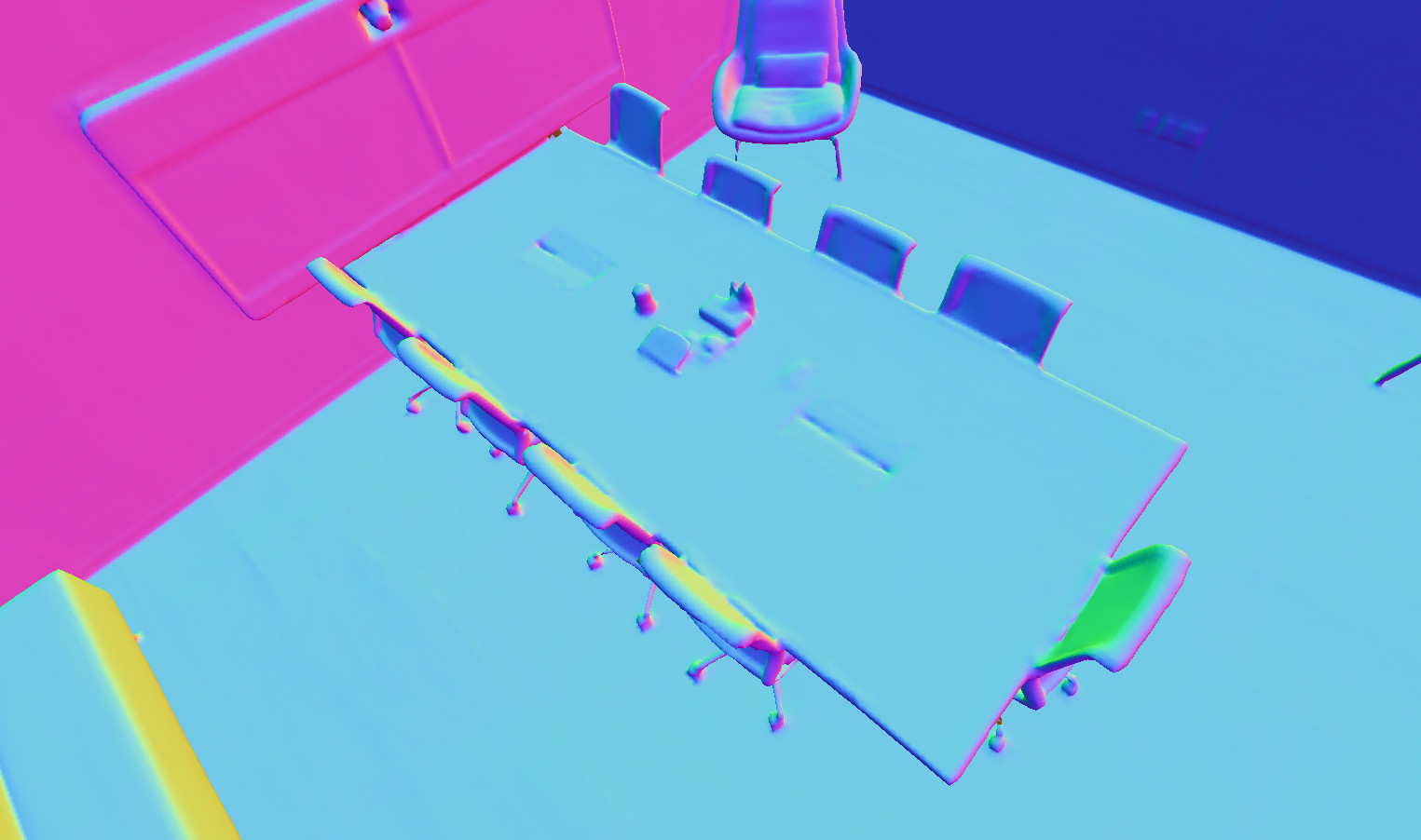} \\
\rotatebox[origin=c]{90}{\texttt{Room 1}} & 
\includegraphics[valign=c,width=\sz\linewidth]{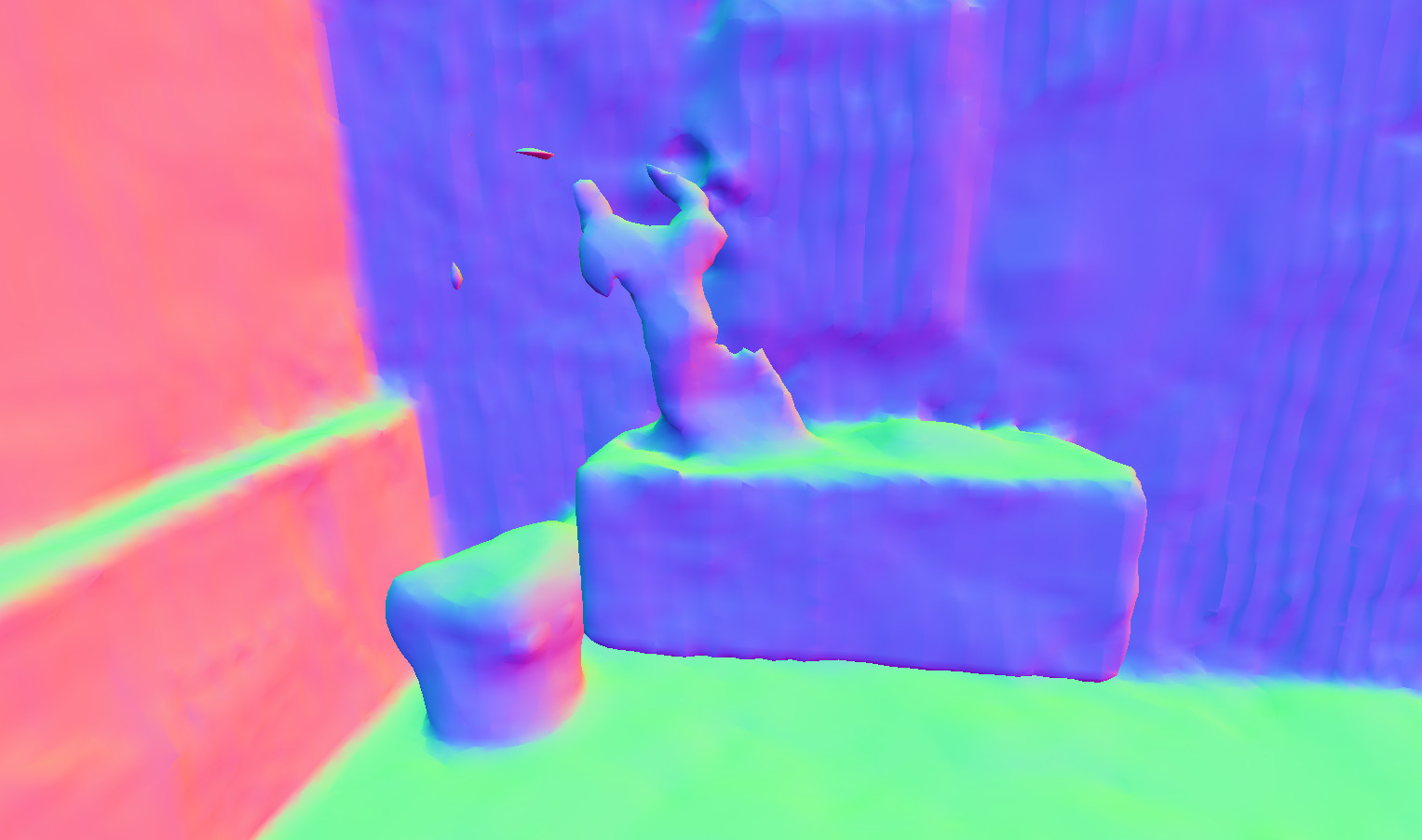} & 
\includegraphics[valign=c,width=\sz\linewidth]{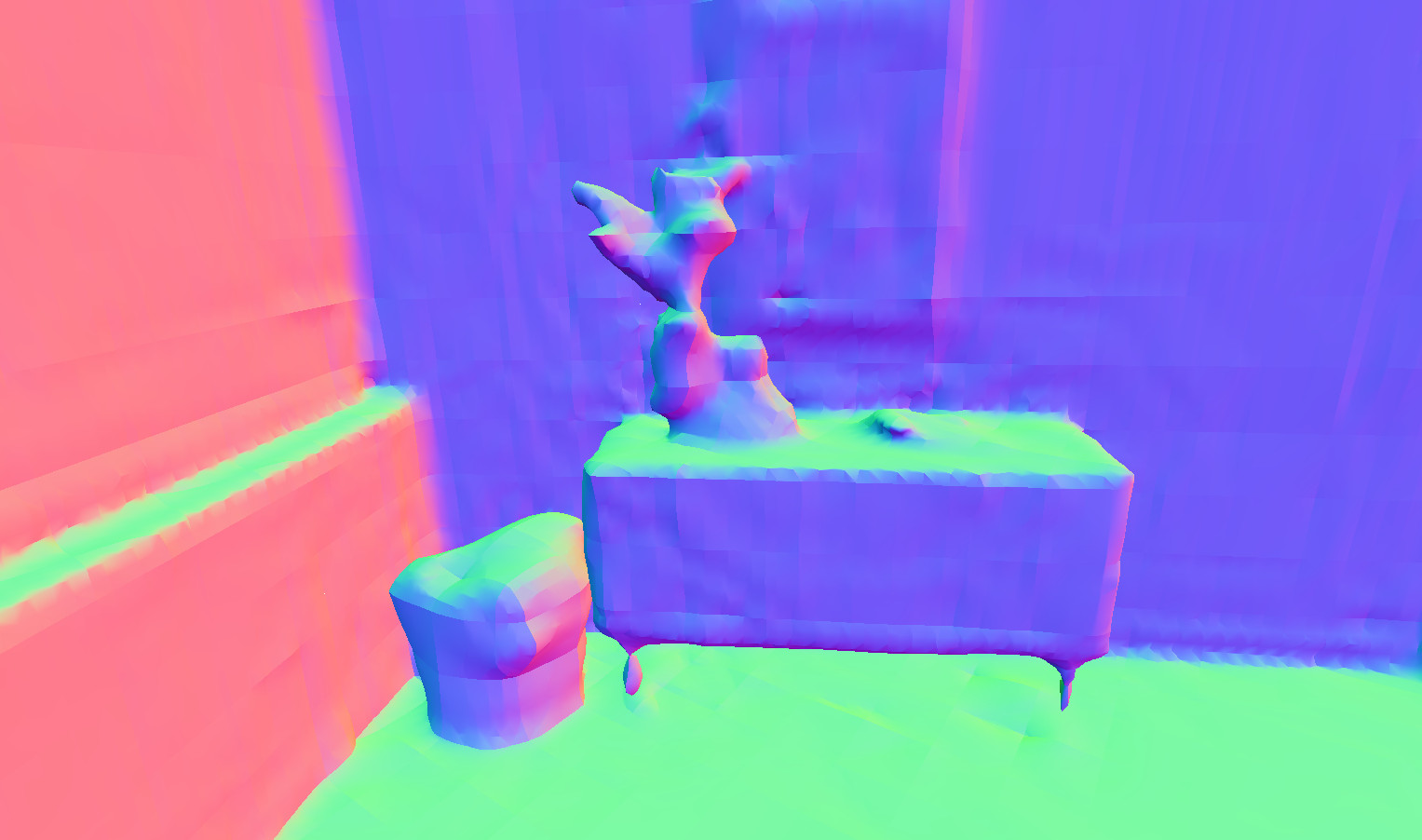} &
\includegraphics[valign=c,width=\sz\linewidth]{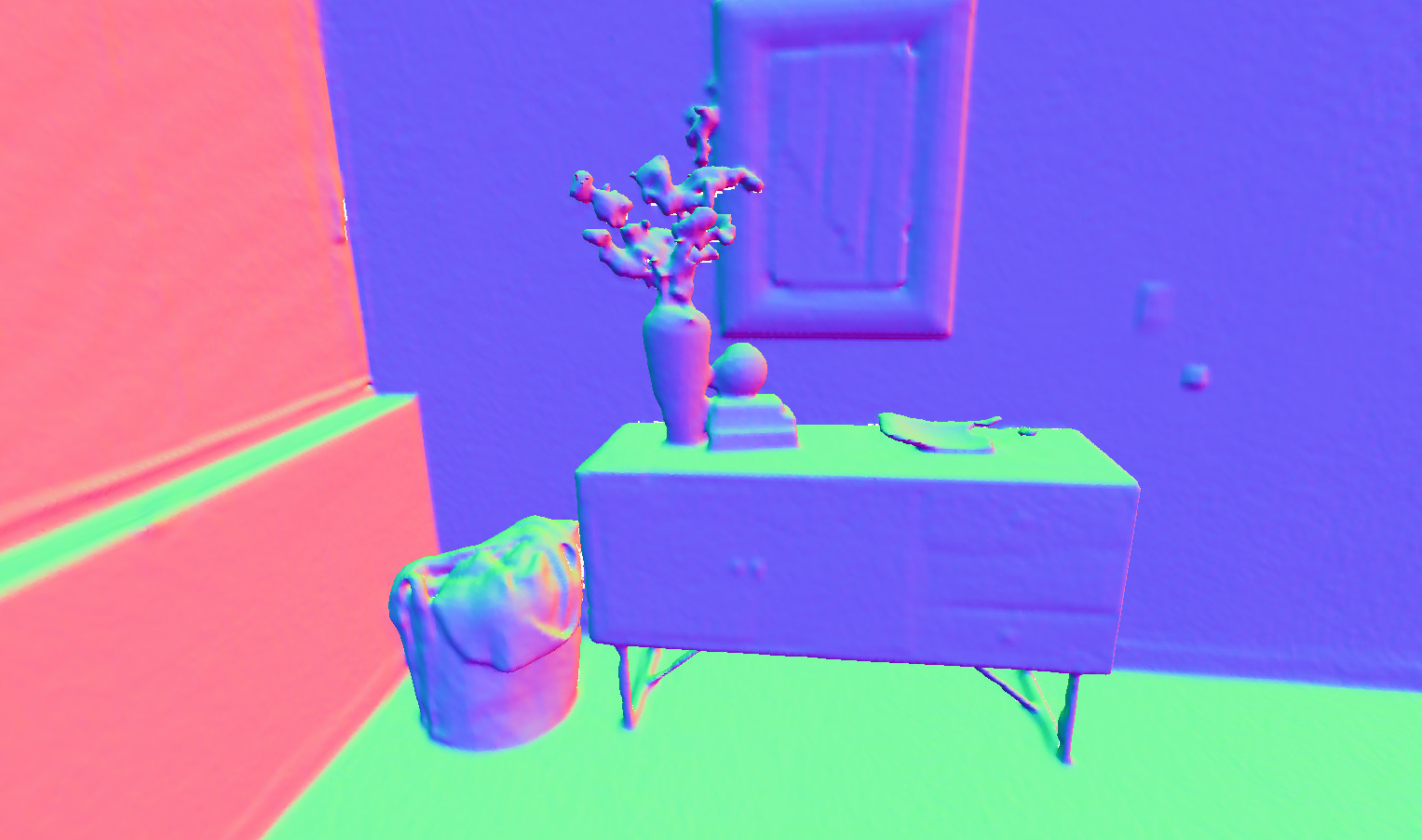} &
\includegraphics[valign=c,width=\sz\linewidth]{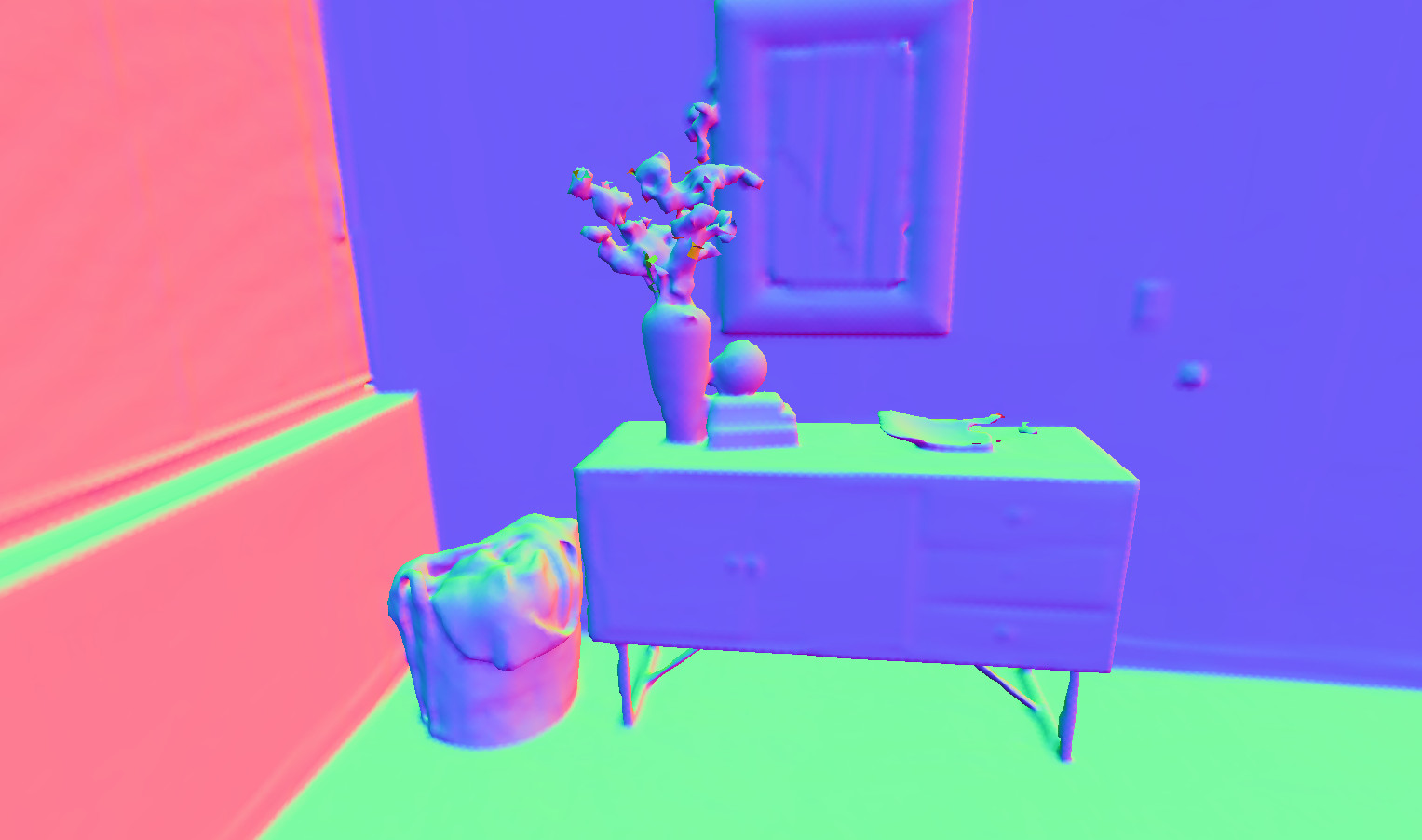} \\
\rotatebox[origin=c]{90}{\texttt{Room 2}} & 
\includegraphics[valign=c,width=\sz\linewidth]{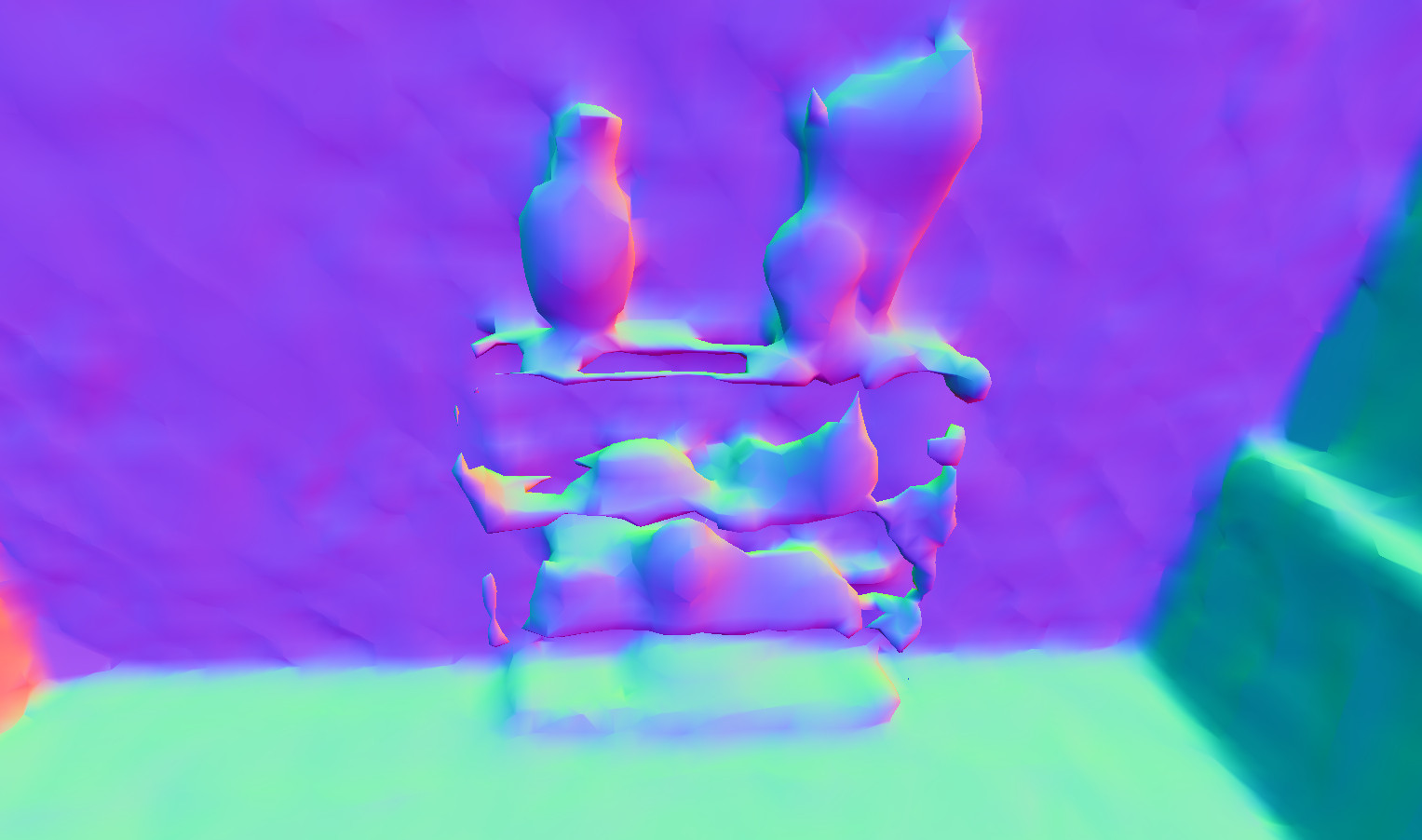} & 
\includegraphics[valign=c,width=\sz\linewidth]{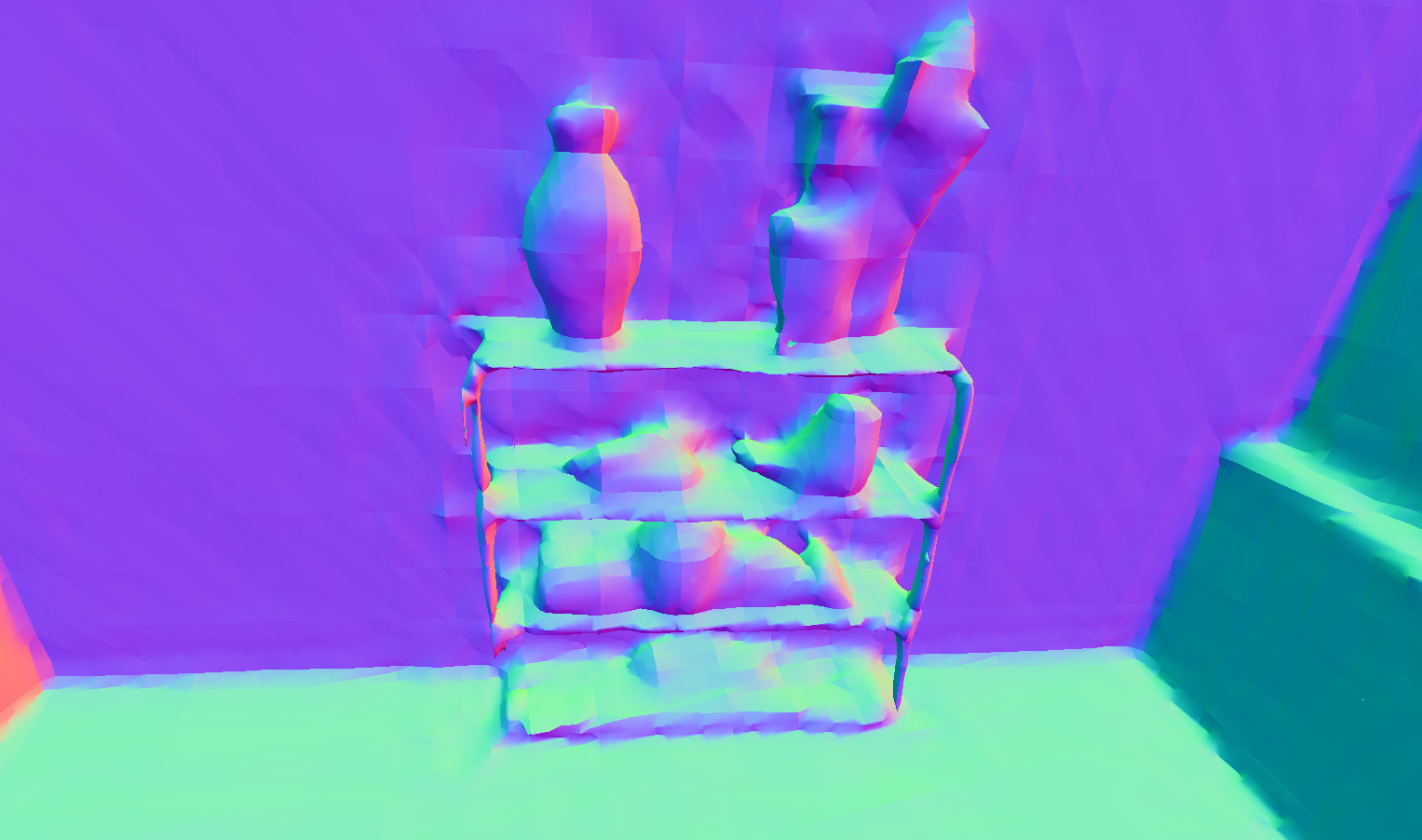} &
\includegraphics[valign=c,width=\sz\linewidth]{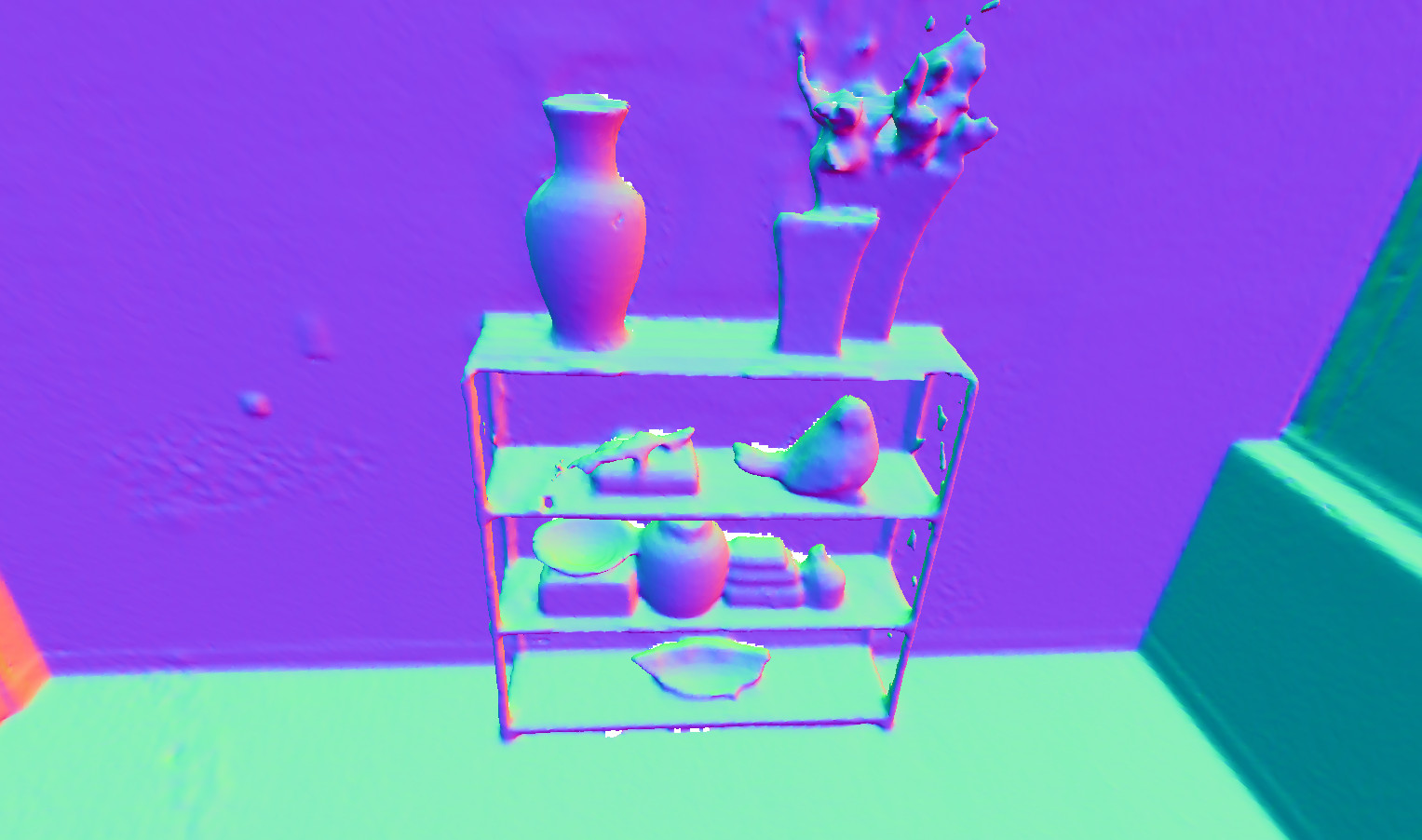} &
\includegraphics[valign=c,width=\sz\linewidth]{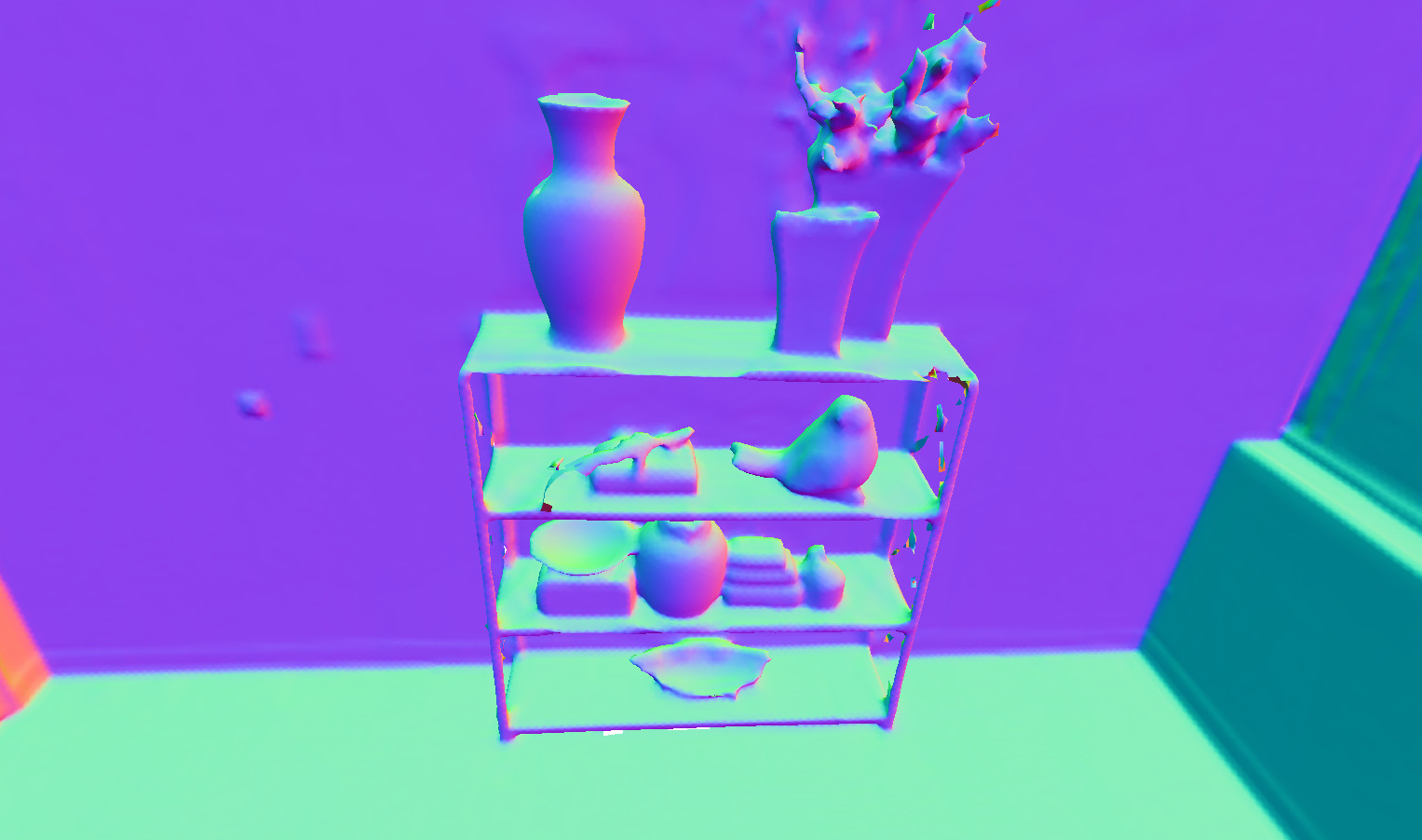} \\
 & NICE-SLAM~\cite{zhu2022nice} & Vox-Fusion$^{*}$~\cite{yang2022vox} & \ours (ours) & Ground Truth \\
\end{tabular}
}
\caption{\textbf{Reconstruction Performance on Replica~\cite{straub2019replica}}. \ours yields on average more precise reconstructions than existing methods. We use normal shading to highlight geometric changes better.}
\label{fig:replica_recon}
\end{figure*}

\begin{figure*}[tb]
\centering
{\footnotesize
\setlength{\tabcolsep}{1pt}
\renewcommand{\arraystretch}{1}
\newcommand{\sz}{0.23}
\begin{tabular}{ccccc}
\rotatebox[origin=c]{90}{\texttt{Office 2}} & 
\includegraphics[valign=c,width=\sz\linewidth]{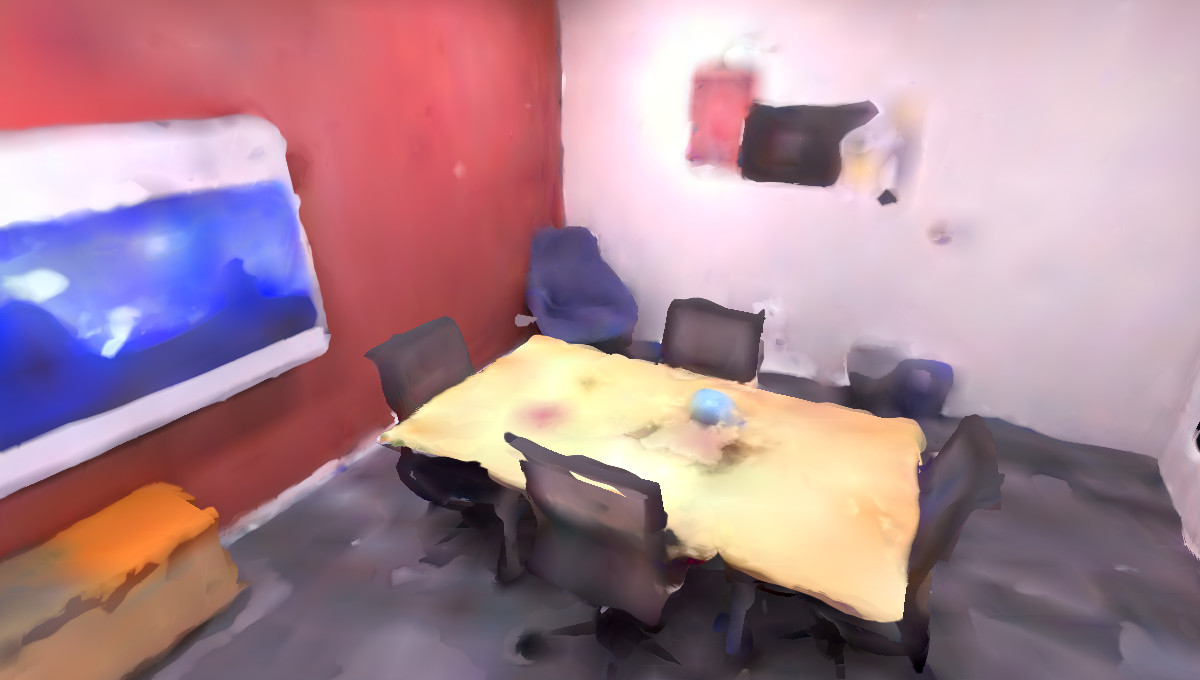} & 
\includegraphics[valign=c,width=\sz\linewidth]{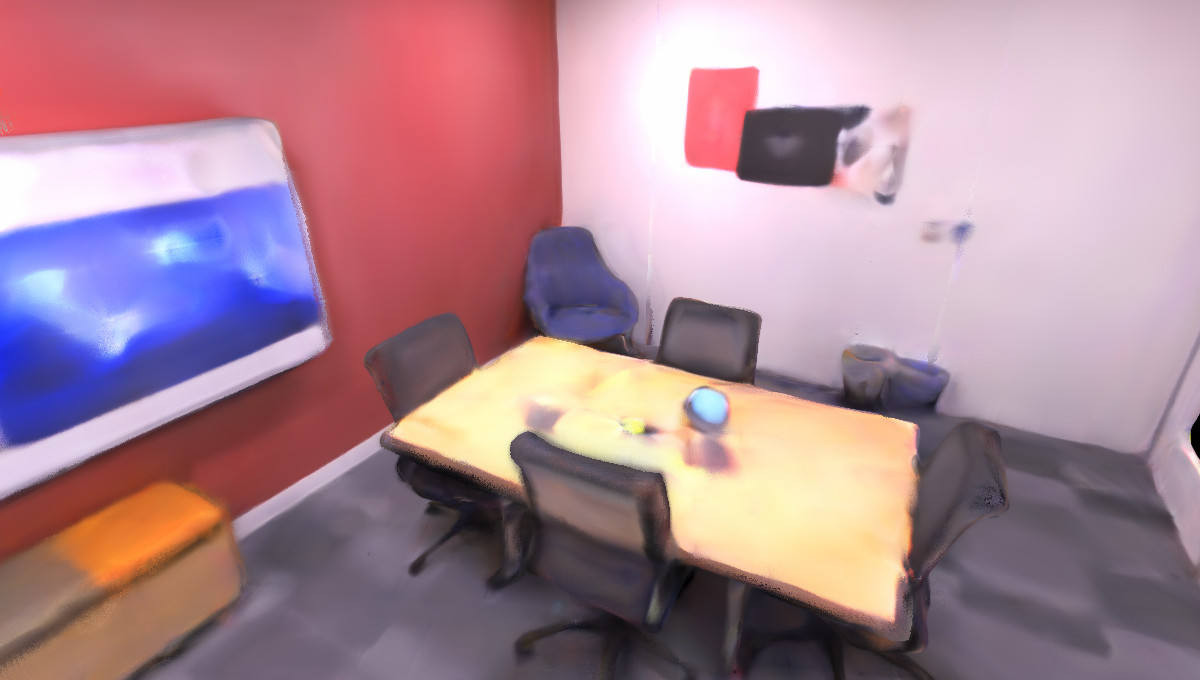} &
\includegraphics[valign=c,width=\sz\linewidth]{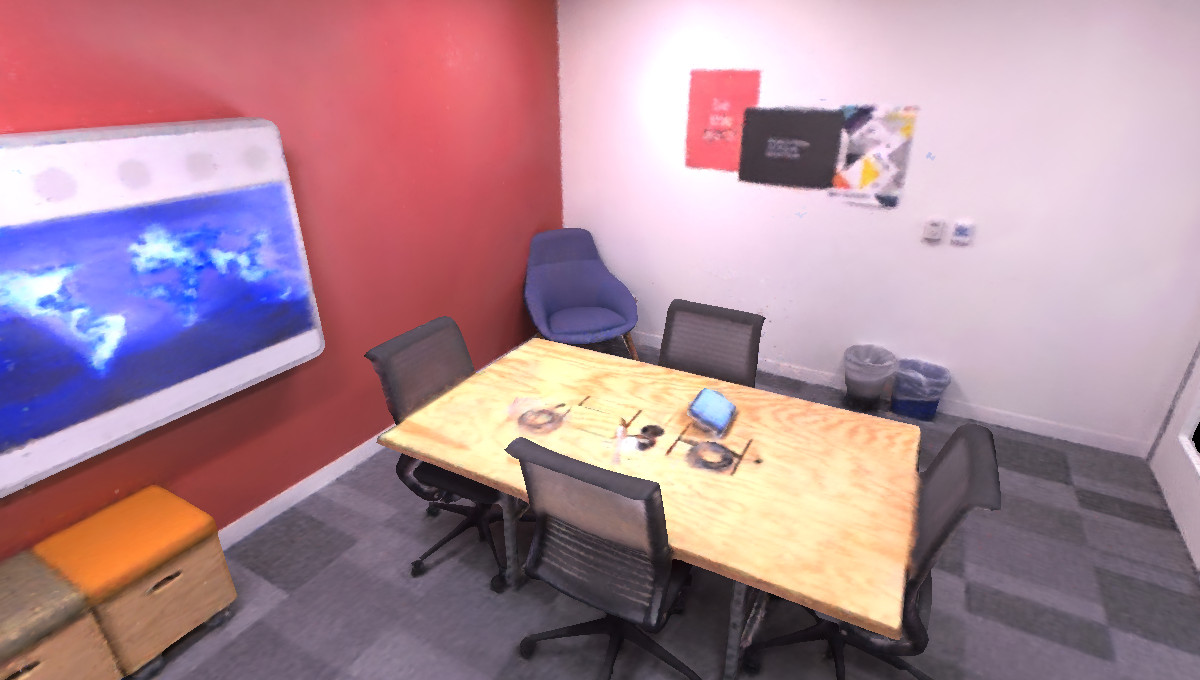} &
\includegraphics[valign=c,width=\sz\linewidth]{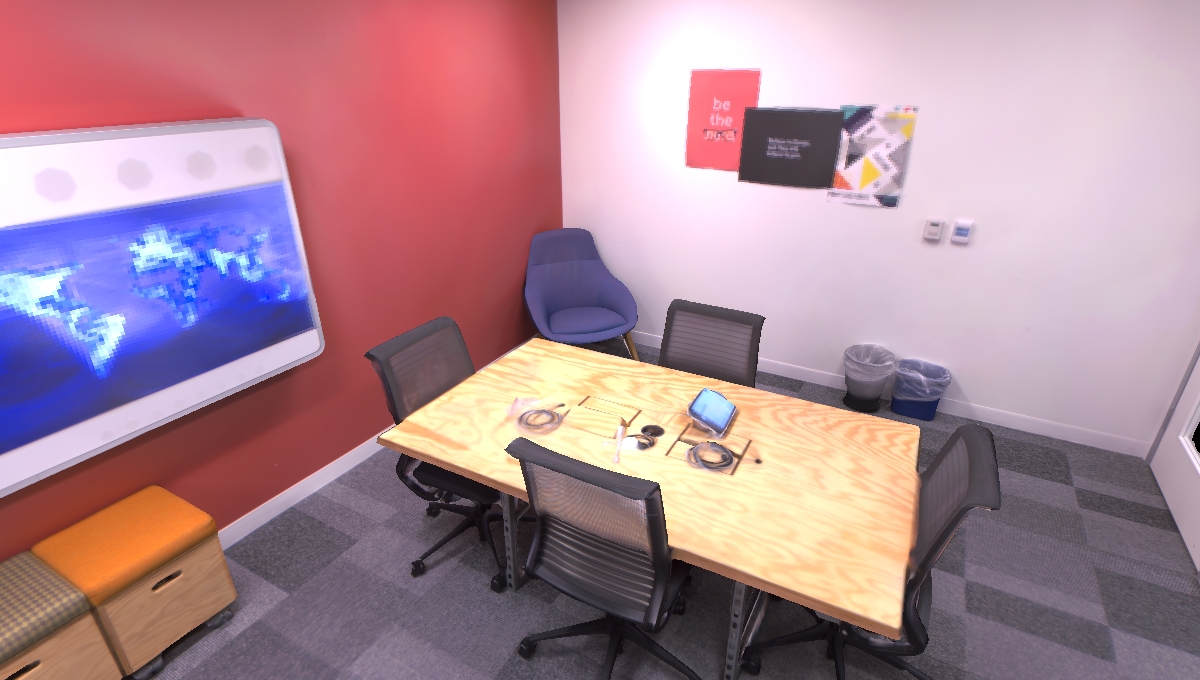} \\
\rotatebox[origin=c]{90}{\texttt{Office 3}} & 
\includegraphics[valign=c,width=\sz\linewidth]{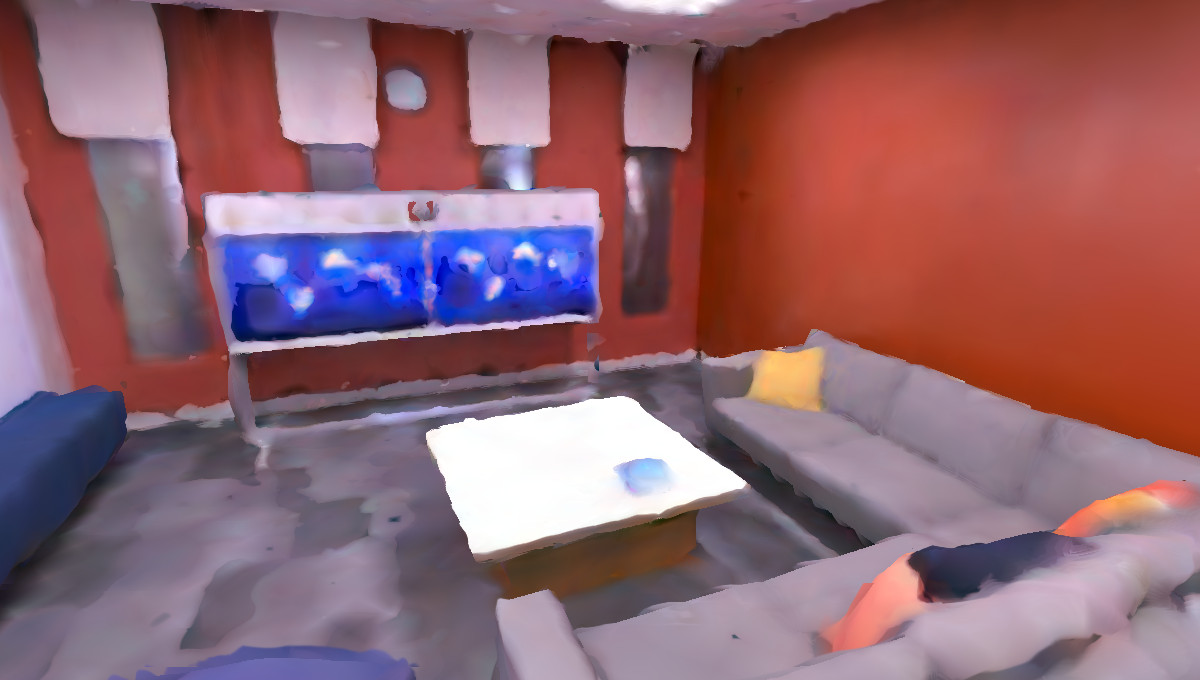} & 
\includegraphics[valign=c,width=\sz\linewidth]{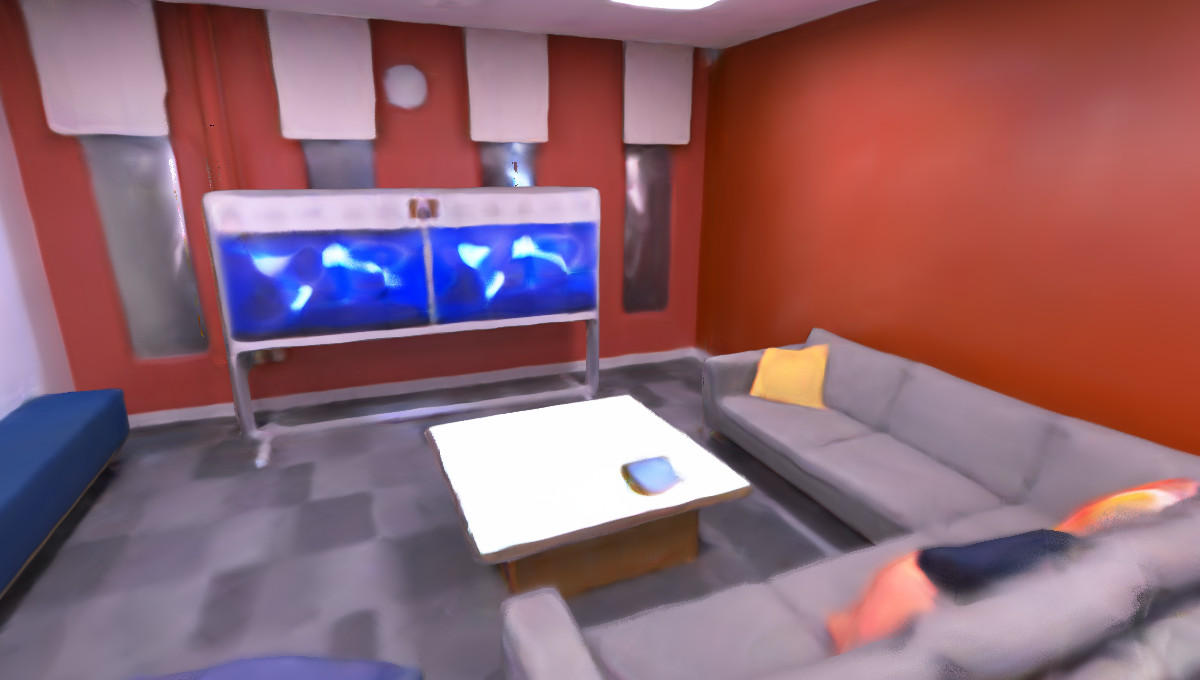} &
\includegraphics[valign=c,width=\sz\linewidth]{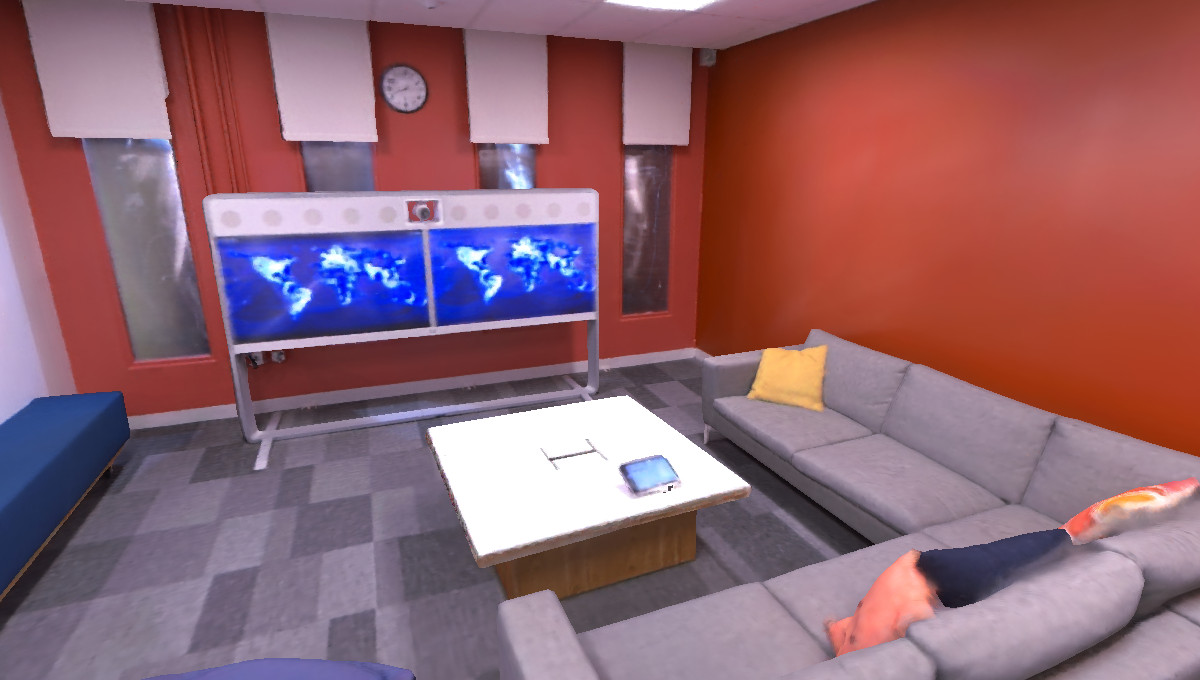} &
\includegraphics[valign=c,width=\sz\linewidth]{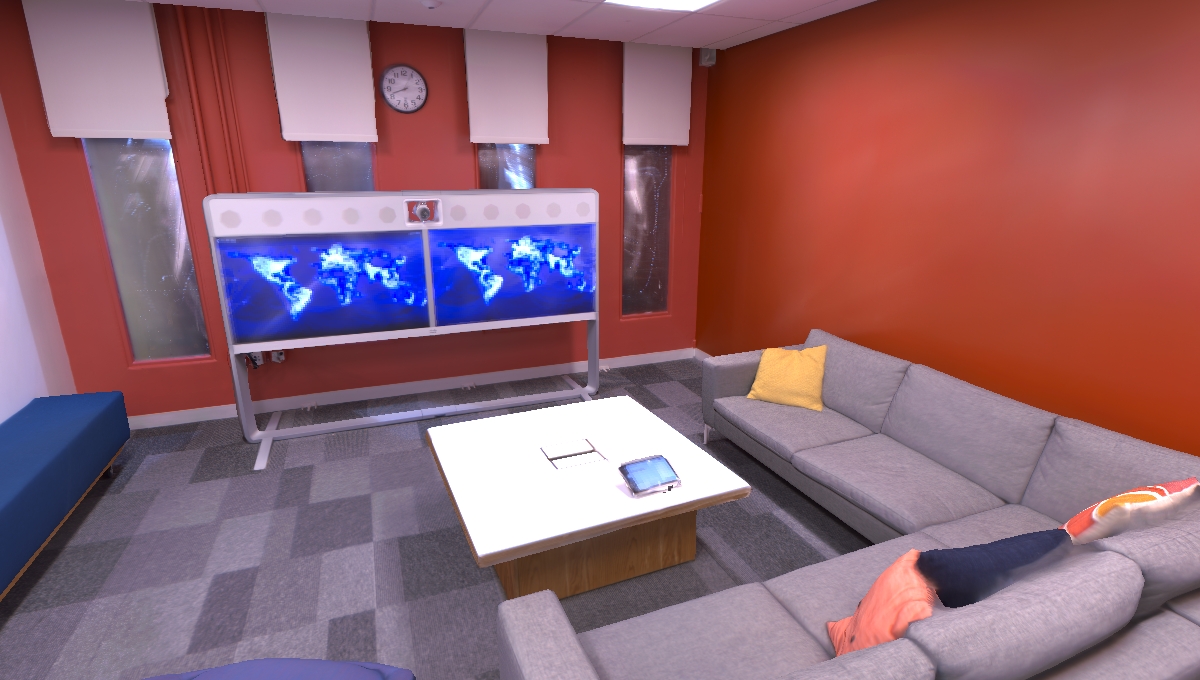} \\
\rotatebox[origin=c]{90}{\texttt{Office 4}} & 
\includegraphics[valign=c,width=\sz\linewidth]{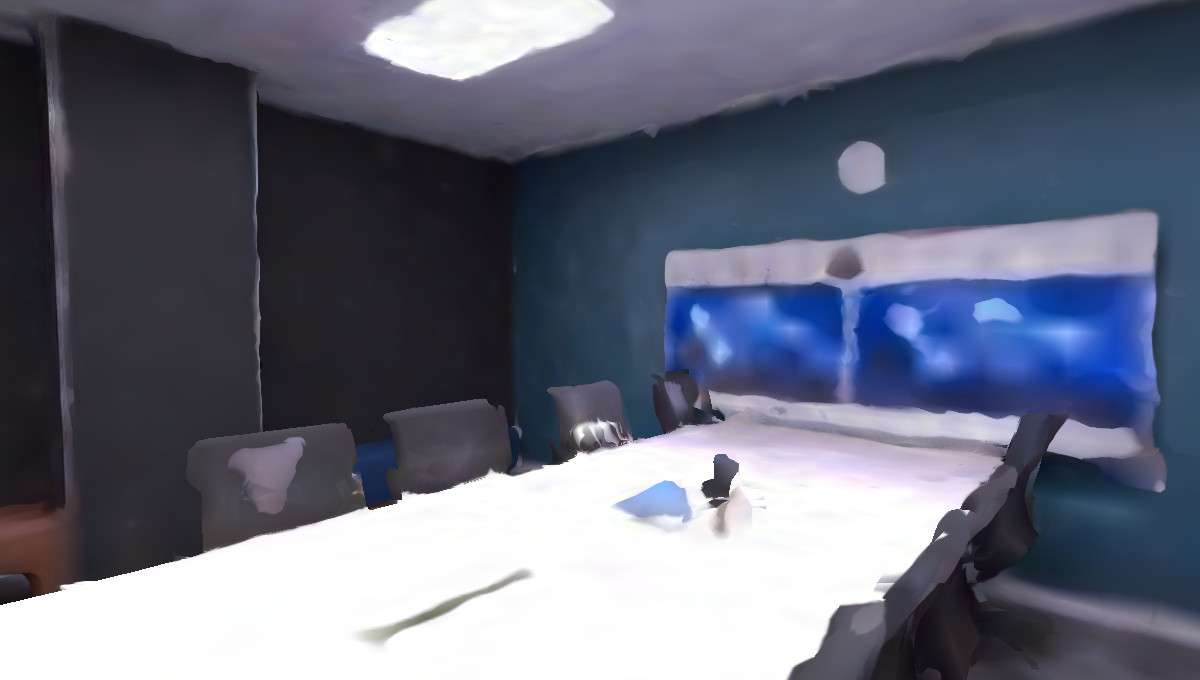} & 
\includegraphics[valign=c,width=\sz\linewidth]{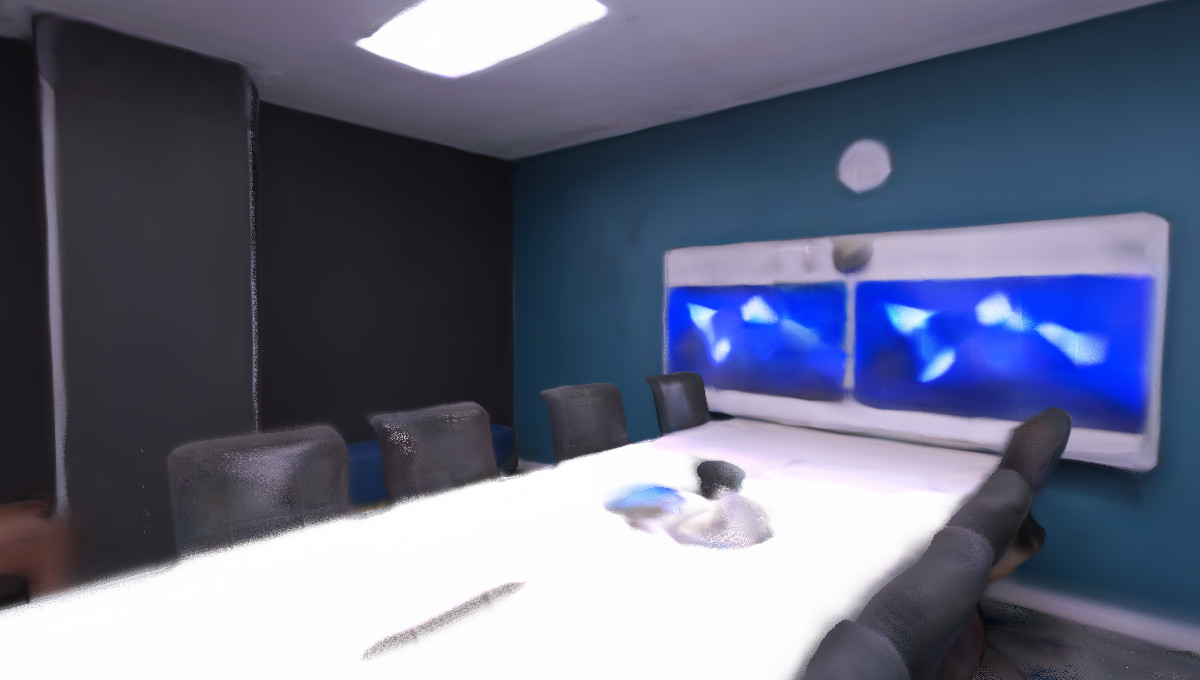} &
\includegraphics[valign=c,width=\sz\linewidth]{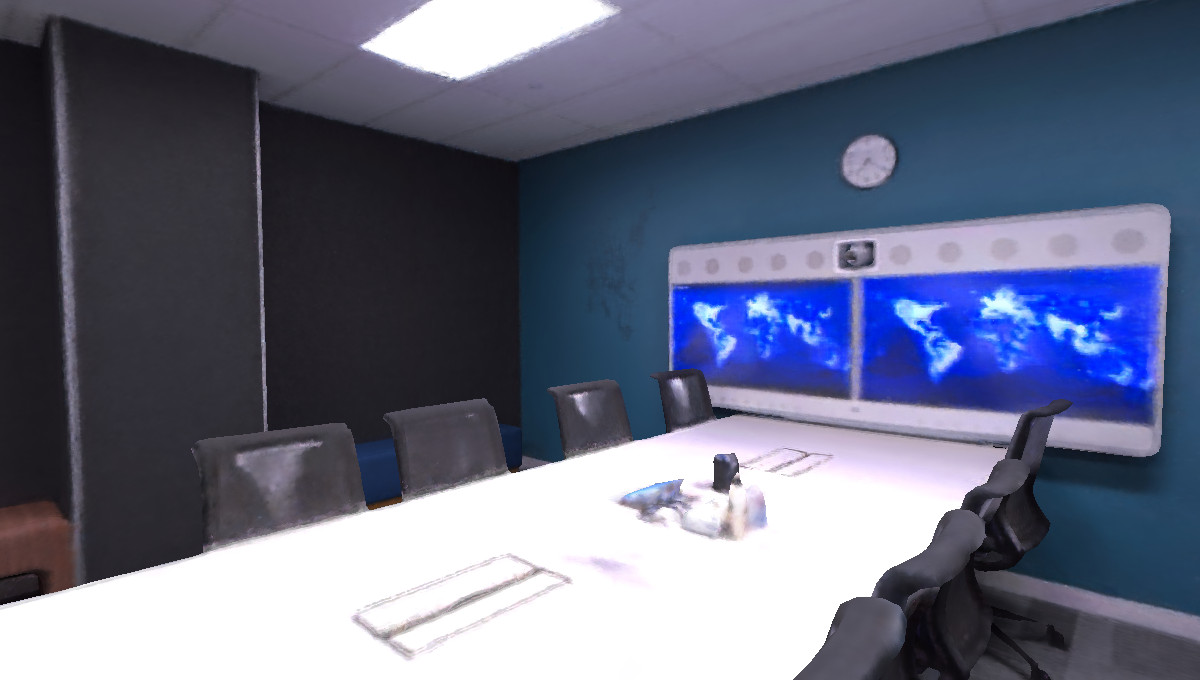} &
\includegraphics[valign=c,width=\sz\linewidth]{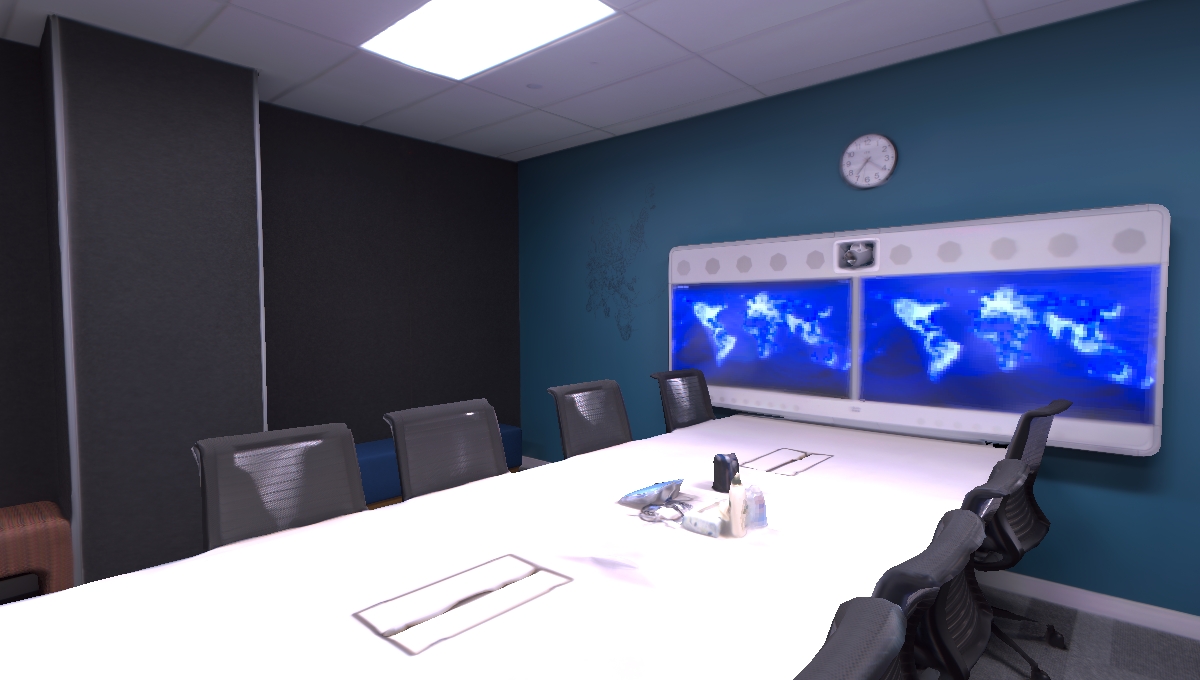} \\
 & NICE-SLAM~\cite{zhu2022nice} & Vox-Fusion$^{*}$~\cite{yang2022vox} & \ours (ours) & Ground Truth \\
\end{tabular}
}
\caption{\textbf{Rendering Performance on Replica~\cite{straub2019replica}}. Thanks to the adaptive density of the neural point cloud, \ours{} is able to encode more high-frequency details and to substantially increase the fidelity of the renderings.}
\label{fig:replica_rendering}
\end{figure*}

\clearpage

{\small
\balance
\bibliographystyle{ieee_fullname}
\bibliography{egbib}

\begin{thebibliography}{10}\itemsep=-1pt

\bibitem{azinovic2022neural}
Dejan Azinovi{\'c}, Ricardo Martin-Brualla, Dan~B Goldman, Matthias
  Nie{\ss}ner, and Justus Thies.
\newblock Neural rgb-d surface reconstruction.
\newblock In {\em IEEE/CVF Conference on Computer Vision and Pattern
  Recognition}, pages 6290--6301, 2022.

\bibitem{bian2022nope}
Wenjing Bian, Zirui Wang, Kejie Li, Jia-Wang Bian, and Victor~Adrian
  Prisacariu.
\newblock Nope-nerf: Optimising neural radiance field with no pose prior.
\newblock {\em arXiv preprint arXiv:2212.07388}, 2022.

\bibitem{bovzivc2021transformerfusion}
Alja{\v{z}} Bo{\v{z}}i{\v{c}}, Pablo Palafox, Justus Thies, Angela Dai, and
  Matthias Nie{\ss}ner.
\newblock Transformerfusion: Monocular rgb scene reconstruction using
  transformers.
\newblock {\em arXiv preprint arXiv:2107.02191}, 2021.

\bibitem{7900211}
E. {Bylow}, C. {Olsson}, and F. {Kahl}.
\newblock Robust online 3d reconstruction combining a depth sensor and sparse
  feature points.
\newblock In {\em 2016 23rd International Conference on Pattern Recognition
  (ICPR)}, pages 3709--3714, 2016.

\bibitem{cao2018real}
Yan-Pei Cao, Leif Kobbelt, and Shi-Min Hu.
\newblock Real-time high-accuracy three-dimensional reconstruction with
  consumer rgb-d cameras.
\newblock {\em ACM Transactions on Graphics (TOG)}, 37(5):1--16, 2018.

\bibitem{chen2013scalable}
Jiawen Chen, Dennis Bautembach, and Shahram Izadi.
\newblock Scalable real-time volumetric surface reconstruction.
\newblock {\em ACM Transactions on Graphics (ToG)}, 32(4):1--16, 2013.

\bibitem{Chen2023catnips}
Timothy Chen, Preston Culbertson, and Mac Schwager.
\newblock Catnips: Collision avoidance through neural implicit probabilistic
  scenes, 2023.

\bibitem{chen2019learning}
Zhiqin Chen and Hao Zhang.
\newblock Learning implicit fields for generative shape modeling.
\newblock In {\em IEEE/CVF conference on computer vision and pattern
  recognition}, pages 5939--5948, 2019.

\bibitem{cho2021sp}
Hae~Min Cho, HyungGi Jo, and Euntai Kim.
\newblock Sp-slam: Surfel-point simultaneous localization and mapping.
\newblock {\em IEEE/ASME Transactions on Mechatronics}, 27(5):2568--2579, 2021.

\bibitem{choe2021volumefusion}
Jaesung Choe, Sunghoon Im, Francois Rameau, Minjun Kang, and In~So Kweon.
\newblock Volumefusion: Deep depth fusion for 3d scene reconstruction.
\newblock In {\em IEEE/CVF International Conference on Computer Vision (ICCV)},
  pages 16086--16095, October 2021.

\bibitem{choi2015robust}
Sungjoon Choi, Qian-Yi Zhou, and Vladlen Koltun.
\newblock Robust reconstruction of indoor scenes.
\newblock In {\em IEEE Conference on Computer Vision and Pattern Recognition},
  pages 5556--5565, 2015.

\bibitem{chung2022orbeez}
Chi-Ming Chung, Yang-Che Tseng, Ya-Ching Hsu, Xiang-Qian Shi, Yun-Hung Hua,
  Jia-Fong Yeh, Wen-Chin Chen, Yi-Ting Chen, and Winston~H Hsu.
\newblock Orbeez-slam: A real-time monocular visual slam with orb features and
  nerf-realized mapping.
\newblock {\em arXiv preprint arXiv:2209.13274}, 2022.

\bibitem{curless1996volumetric}
Brian Curless and Marc Levoy.
\newblock {Volumetric method for building complex models from range images}.
\newblock In {\em SIGGRAPH Conference on Computer Graphics}. ACM, 1996.

\bibitem{Dai2017ScanNet}
Angela Dai, Angel~X. Chang, Manolis Savva, Maciej Halber, Thomas Funkhouser,
  and Matthias Nie{\ss}ner.
\newblock {ScanNet: Richly-annotated 3D reconstructions of indoor scenes}.
\newblock In {\em Conference on Computer Vision and Pattern Recognition
  (CVPR)}. IEEE/CVF, 2017.

\bibitem{dai2017bundlefusion}
Angela Dai, Matthias Nie{\ss}ner, Michael Zollh{\"o}fer, Shahram Izadi, and
  Christian Theobalt.
\newblock Bundlefusion: Real-time globally consistent 3d reconstruction using
  on-the-fly surface reintegration.
\newblock {\em ACM Transactions on Graphics (ToG)}, 36(4):1, 2017.

\bibitem{fuhrmann2011fusion}
Simon Fuhrmann and Michael Goesele.
\newblock Fusion of depth maps with multiple scales.
\newblock {\em {ACM} Trans. Graph.}, 30(6):148:1--148:8, 2011.

\bibitem{hane2011stereo}
Christian H{\"a}ne, Christopher Zach, Jongwoo Lim, Ananth Ranganathan, and Marc
  Pollefeys.
\newblock Stereo depth map fusion for robot navigation.
\newblock In {\em 2011 IEEE/RSJ International Conference on Intelligent Robots
  and Systems}, pages 1618--1625. IEEE, 2011.

\bibitem{horn1988closed}
Berthold~KP Horn, Hugh~M Hilden, and Shahriar Negahdaripour.
\newblock Closed-form solution of absolute orientation using orthonormal
  matrices.
\newblock {\em JOSA A}, 5(7):1127--1135, 1988.

\bibitem{huang2021di}
Jiahui Huang, Shi-Sheng Huang, Haoxuan Song, and Shi-Min Hu.
\newblock Di-fusion: Online implicit 3d reconstruction with deep priors.
\newblock In {\em IEEE/CVF Conference on Computer Vision and Pattern
  Recognition}, pages 8932--8941, 2021.

\bibitem{johnson2019billion}
Jeff Johnson, Matthijs Douze, and Herv{\'e} J{\'e}gou.
\newblock Billion-scale similarity search with {GPUs}.
\newblock {\em IEEE Transactions on Big Data}, 7(3):535--547, 2019.

\bibitem{kahler2015hierarchical}
Olaf K{\"a}hler, Victor Prisacariu, Julien Valentin, and David Murray.
\newblock Hierarchical voxel block hashing for efficient integration of depth
  images.
\newblock {\em IEEE Robotics and Automation Letters}, 1(1):192--197, 2015.

\bibitem{Kahler2015infiniTAM}
Olaf K{\"{a}}hler, Victor~Adrian Prisacariu, Carl~Yuheng Ren, Xin Sun, Philip
  H.~S. Torr, and David~William Murray.
\newblock Very high frame rate volumetric integration of depth images on mobile
  devices.
\newblock {\em {IEEE} Trans. Vis. Comput. Graph.}, 21(11):1241--1250, 2015.

\bibitem{keller2013real}
Maik Keller, Damien Lefloch, Martin Lambers, Shahram Izadi, Tim Weyrich, and
  Andreas Kolb.
\newblock Real-time 3d reconstruction in dynamic scenes using point-based
  fusion.
\newblock In {\em International Conference on 3D Vision (3DV)}, pages 1--8.
  IEEE, 2013.

\bibitem{li2023dense}
Heng Li, Xiaodong Gu, Weihao Yuan, Luwei Yang, Zilong Dong, and Ping Tan.
\newblock Dense rgb slam with neural implicit maps.
\newblock {\em arXiv preprint arXiv:2301.08930}, 2023.

\bibitem{li2022bnv}
Kejie Li, Yansong Tang, Victor~Adrian Prisacariu, and Philip~HS Torr.
\newblock Bnv-fusion: Dense 3d reconstruction using bi-level neural volume
  fusion.
\newblock In {\em IEEE/CVF Conference on Computer Vision and Pattern
  Recognition}, pages 6166--6175, 2022.

\bibitem{Lin2021BARF:Fields}
Chen~Hsuan Lin, Wei~Chiu Ma, Antonio Torralba, and Simon Lucey.
\newblock {BARF: Bundle-Adjusting Neural Radiance Fields}.
\newblock In {\em International Conference on Computer Vision (ICCV)}.
  IEEE/CVF, 2021.

\bibitem{liu2020neural}
Lingjie Liu, Jiatao Gu, Kyaw Zaw~Lin, Tat-Seng Chua, and Christian Theobalt.
\newblock Neural sparse voxel fields.
\newblock {\em Advances in Neural Information Processing Systems},
  33:15651--15663, 2020.

\bibitem{lorensen1987marching}
William~E Lorensen and Harvey~E Cline.
\newblock Marching cubes: A high resolution 3d surface construction algorithm.
\newblock {\em ACM siggraph computer graphics}, 21(4):163--169, 1987.

\bibitem{mahdi2022eslam}
Mohammad Mahdi~Johari, Camilla Carta, and Fran{\c{c}}ois Fleuret.
\newblock Eslam: Efficient dense slam system based on hybrid representation of
  signed distance fields.
\newblock {\em arXiv e-prints}, pages arXiv--2211, 2022.

\bibitem{marniok2017efficient}
Nico Marniok, Ole Johannsen, and Bastian Goldluecke.
\newblock An efficient octree design for local variational range image fusion.
\newblock In {\em German Conference on Pattern Recognition (GCPR)}, pages
  401--412. Springer, 2017.

\bibitem{mescheder2019occupancy}
Lars Mescheder, Michael Oechsle, Michael Niemeyer, Sebastian Nowozin, and
  Andreas Geiger.
\newblock Occupancy networks: Learning 3d reconstruction in function space.
\newblock In {\em IEEE/CVF conference on computer vision and pattern
  recognition}, pages 4460--4470, 2019.

\bibitem{mihajlovic2021deepsurfels}
Marko Mihajlovic, Silvan Weder, Marc Pollefeys, and Martin~R Oswald.
\newblock Deepsurfels: Learning online appearance fusion.
\newblock In {\em IEEE/CVF Conference on Computer Vision and Pattern
  Recognition}, pages 14524--14535, 2021.

\bibitem{Mildenhall2020NeRF:Synthesis}
Ben Mildenhall, Pratul~P. Srinivasan, Matthew Tancik, Jonathan~T. Barron, Ravi
  Ramamoorthi, and Ren Ng.
\newblock {NeRF: Representing Scenes as Neural Radiance Fields for View
  Synthesis}.
\newblock In {\em European Conference on Computer Vision (ECCV)}. CVF, 2020.

\bibitem{muller2022instant}
Thomas M{\"u}ller, Alex Evans, Christoph Schied, and Alexander Keller.
\newblock Instant neural graphics primitives with a multiresolution hash
  encoding.
\newblock {\em arXiv preprint arXiv:2201.05989}, 2022.

\bibitem{Mur-Artal2017ORB-SLAM2:Cameras}
Raul Mur-Artal and Juan~D. Tardos.
\newblock {ORB-SLAM2: An Open-Source SLAM System for Monocular, Stereo, and
  RGB-D Cameras}.
\newblock {\em IEEE Transactions on Robotics}, 33(5):1255--1262, 2017.

\bibitem{murez2020atlas}
Zak Murez, Tarrence van As, James Bartolozzi, Ayan Sinha, Vijay Badrinarayanan,
  and Andrew Rabinovich.
\newblock Atlas: End-to-end 3d scene reconstruction from posed images.
\newblock In {\em Computer Vision--ECCV 2020: 16th European Conference,
  Glasgow, UK, August 23--28, 2020, Proceedings, Part VII 16}, pages 414--431.
  Springer, 2020.

\bibitem{newcombe2011kinectfusion}
Richard~A Newcombe, Shahram Izadi, Otmar Hilliges, David Molyneaux, David Kim,
  Andrew~J Davison, Pushmeet Kohli, Jamie Shotton, Steve Hodges, and Andrew~W
  Fitzgibbon.
\newblock Kinectfusion: Real-time dense surface mapping and tracking.
\newblock In {\em ISMAR}, volume~11, pages 127--136, 2011.

\bibitem{newcombe2011dtam}
Richard~A Newcombe, Steven~J Lovegrove, and Andrew~J Davison.
\newblock Dtam: Dense tracking and mapping in real-time.
\newblock In {\em International Conference on Computer Vision (ICCV)}, 2011.

\bibitem{niessner2013voxel_hashing}
Matthias Nießner, Michael Zollhöfer, Shahram Izadi, and Marc Stamminger.
\newblock Real-time 3d reconstruction at scale using voxel hashing.
\newblock {\em ACM Transactions on Graphics (TOG)}, 32, 11 2013.

\bibitem{Oechsle2021UNISURF:Reconstruction}
Michael Oechsle, Songyou Peng, and Andreas Geiger.
\newblock {UNISURF: Unifying Neural Implicit Surfaces and Radiance Fields for
  Multi-View Reconstruction}.
\newblock In {\em International Conference on Computer Vision (ICCV)}.
  IEEE/CVF, 2021.

\bibitem{Oleynikova2017voxblox}
Helen Oleynikova, Zachary Taylor, Marius Fehr, Roland Siegwart, and Juan~I.
  Nieto.
\newblock Voxblox: Incremental 3d euclidean signed distance fields for on-board
  {MAV} planning.
\newblock In {\em 2017 {IEEE/RSJ} International Conference on Intelligent
  Robots and Systems, {IROS} 2017, Vancouver, BC, Canada, September 24-28,
  2017}, pages 1366--1373. {IEEE}, 2017.

\bibitem{ortiz2022isdf}
Joseph Ortiz, Alexander Clegg, Jing Dong, Edgar Sucar, David Novotny, Michael
  Zollhoefer, and Mustafa Mukadam.
\newblock isdf: Real-time neural signed distance fields for robot perception.
\newblock {\em arXiv preprint arXiv:2204.02296}, 2022.

\bibitem{park2019deepsdf}
Jeong~Joon Park, Peter Florence, Julian Straub, Richard Newcombe, and Steven
  Lovegrove.
\newblock Deepsdf: Learning continuous signed distance functions for shape
  representation.
\newblock In {\em IEEE/CVF conference on computer vision and pattern
  recognition}, pages 165--174, 2019.

\bibitem{peng2020convolutional}
Songyou Peng, Michael Niemeyer, Lars Mescheder, Marc Pollefeys, and Andreas
  Geiger.
\newblock {Convolutional Occupancy Networks}.
\newblock In {\em European Conference Computer Vision (ECCV)}. CVF, 2020.

\bibitem{Rematas2021UrbanFields}
Konstantinos Rematas, Andrew Liu, Pratul~P. Srinivasan, Jonathan~T. Barron,
  Andrea Tagliasacchi, Thomas Funkhouser, and Vittorio Ferrari.
\newblock {Urban Radiance Fields}.
\newblock In {\em Conference on Computer Vision and Pattern Recognition
  (CVPR)}. IEEE/CVF, 2021.

\bibitem{Rosinol2022NeRF-SLAM:Fields}
Antoni Rosinol, John~J. Leonard, and Luca Carlone.
\newblock {NeRF-SLAM: Real-Time Dense Monocular SLAM with Neural Radiance
  Fields}.
\newblock {\em arXiv}, 2022.

\bibitem{ross2022bev}
James Ross, Oscar Mendez, Avishkar Saha, Mark Johnson, and Richard Bowden.
\newblock Bev-slam: Building a globally-consistent world map using monocular
  vision.
\newblock In {\em 2022 IEEE/RSJ International Conference on Intelligent Robots
  and Systems (IROS)}, pages 3830--3836. IEEE, 2022.

\bibitem{Sandstrom2022LearningFusion}
Erik Sandstr{\"{o}}m, Martin~R. Oswald, Suryansh Kumar, Silvan Weder, Fisher
  Yu, Cristian Sminchisescu, and Luc Van~Gool.
\newblock {Learning Online Multi-Sensor Depth Fusion}.
\newblock In {\em European Conference Computer Vision (ECCV)}. CVF, 2022.

\bibitem{sayed2022simplerecon}
Mohamed Sayed, John Gibson, Jamie Watson, Victor Prisacariu, Michael Firman,
  and Cl{\'e}ment Godard.
\newblock Simplerecon: 3d reconstruction without 3d convolutions.
\newblock In {\em European Conference on Computer Vision}, pages 1--19.
  Springer, 2022.

\bibitem{schops2019bad}
Thomas Schops, Torsten Sattler, and Marc Pollefeys.
\newblock {BAD SLAM}: Bundle adjusted direct {RGB-D} {SLAM}.
\newblock In {\em CVF/IEEE Conference on Computer Vision and Pattern
  Recognition (CVPR)}, 2019.

\bibitem{steinbrucker2013large}
Frank Steinbrucker, Christian Kerl, and Daniel Cremers.
\newblock Large-scale multi-resolution surface reconstruction from rgb-d
  sequences.
\newblock In {\em IEEE International Conference on Computer Vision}, pages
  3264--3271, 2013.

\bibitem{stier2021vortx}
Noah Stier, Alexander Rich, Pradeep Sen, and Tobias H{\"o}llerer.
\newblock Vortx: Volumetric 3d reconstruction with transformers for voxelwise
  view selection and fusion.
\newblock In {\em 2021 International Conference on 3D Vision (3DV)}, pages
  320--330. IEEE, 2021.

\bibitem{straub2019replica}
Julian Straub, Thomas Whelan, Lingni Ma, Yufan Chen, Erik Wijmans, Simon Green,
  Jakob~J Engel, Raul Mur-Artal, Carl Ren, Shobhit Verma, et~al.
\newblock The replica dataset: A digital replica of indoor spaces.
\newblock {\em arXiv preprint arXiv:1906.05797}, 2019.

\bibitem{Sturm2012ASystems}
Jürgen Sturm, Nikolas Engelhard, Felix Endres, Wolfram Burgard, and Daniel
  Cremers.
\newblock {A benchmark for the evaluation of RGB-D SLAM systems}.
\newblock In {\em International Conference on Intelligent Robots and Systems
  (IROS)}. IEEE/RSJ, 2012.

\bibitem{Sucar2021IMAP:Real-Time}
Edgar Sucar, Shikun Liu, Joseph Ortiz, and Andrew~J. Davison.
\newblock {iMAP: Implicit Mapping and Positioning in Real-Time}.
\newblock In {\em International Conference on Computer Vision (ICCV)}.
  IEEE/CVF, 2021.

\bibitem{sun2021neuralrecon}
Jiaming Sun, Yiming Xie, Linghao Chen, Xiaowei Zhou, and Hujun Bao.
\newblock Neuralrecon: Real-time coherent 3d reconstruction from monocular
  video.
\newblock In {\em IEEE/CVF Conference on Computer Vision and Pattern
  Recognition}, pages 15598--15607, 2021.

\bibitem{tancik2020fourier}
Matthew Tancik, Pratul Srinivasan, Ben Mildenhall, Sara Fridovich-Keil, Nithin
  Raghavan, Utkarsh Singhal, Ravi Ramamoorthi, Jonathan Barron, and Ren Ng.
\newblock Fourier features let networks learn high frequency functions in low
  dimensional domains.
\newblock {\em Advances in Neural Information Processing Systems},
  33:7537--7547, 2020.

\bibitem{vakalopoulou2018atlasnet}
Maria Vakalopoulou, Guillaume Chassagnon, Norbert Bus, Rafael Marini,
  Evangelia~I Zacharaki, M-P Revel, and Nikos Paragios.
\newblock Atlasnet: multi-atlas non-linear deep networks for medical image
  segmentation.
\newblock In {\em International Conference on Medical Image Computing and
  Computer-Assisted Intervention}, pages 658--666. Springer, 2018.

\bibitem{wang2022gosurf}
Jingwen Wang, Tymoteusz Bleja, and Lourdes Agapito.
\newblock Go-surf: Neural feature grid optimization for fast, high-fidelity
  rgb-d surface reconstruction.
\newblock In {\em International Conference on 3D Vision}, 2022.

\bibitem{wang2022neuris}
Jiepeng Wang, Peng Wang, Xiaoxiao Long, Christian Theobalt, Taku Komura,
  Lingjie Liu, and Wenping Wang.
\newblock Neuris: Neural reconstruction of indoor scenes using normal priors.
\newblock In {\em Computer Vision--ECCV 2022: 17th European Conference, Tel
  Aviv, Israel, October 23--27, 2022, Proceedings, Part XXXII}, pages 139--155.
  Springer, 2022.

\bibitem{Wang2021NeuS:Reconstruction}
Peng Wang, Lingjie Liu, Yuan Liu, Christian Theobalt, Taku Komura, and Wenping
  Wang.
\newblock {NeuS: Learning Neural Implicit Surfaces by Volume Rendering for
  Multi-view Reconstruction}.
\newblock In {\em Advances in Neural Information Processing Systems (NeurIPS)},
  2021.

\bibitem{wang2022neus2}
Yiming Wang, Qin Han, Marc Habermann, Kostas Daniilidis, Christian Theobalt,
  and Lingjie Liu.
\newblock Neus2: Fast learning of neural implicit surfaces for multi-view
  reconstruction.
\newblock {\em arXiv preprint arXiv:2212.05231}, 2022.

\bibitem{wang2004image}
Zhou Wang, Alan~C Bovik, Hamid~R Sheikh, and Eero~P Simoncelli.
\newblock Image quality assessment: from error visibility to structural
  similarity.
\newblock {\em IEEE transactions on image processing}, 13(4):600--612, 2004.

\bibitem{wang2021nerf}
Zirui Wang, Shangzhe Wu, Weidi Xie, Min Chen, and Victor~Adrian Prisacariu.
\newblock Nerf--: Neural radiance fields without known camera parameters.
\newblock {\em arXiv preprint arXiv:2102.07064}, 2021.

\bibitem{Weder2020RoutedFusion}
Silvan Weder, Johannes Schonberger, Marc Pollefeys, and Martin~R Oswald.
\newblock Routedfusion: Learning real-time depth map fusion.
\newblock In {\em IEEE/CVF Conference on Computer Vision and Pattern
  Recognition}, pages 4887--4897, 2020.

\bibitem{weder2021neuralfusion}
Silvan Weder, Johannes~L Schonberger, Marc Pollefeys, and Martin~R Oswald.
\newblock Neuralfusion: Online depth fusion in latent space.
\newblock In {\em IEEE/CVF Conference on Computer Vision and Pattern
  Recognition}, pages 3162--3172, 2021.

\bibitem{whelan2015elasticfusion}
Thomas Whelan, Stefan Leutenegger, Renato Salas-Moreno, Ben Glocker, and Andrew
  Davison.
\newblock Elasticfusion: Dense slam without a pose graph.
\newblock In {\em Robotics: Science and Systems (RSS)}, 2015.

\bibitem{whelan2012kintinuous}
Thomas Whelan, John McDonald, Michael Kaess, Maurice Fallon, Hordur Johannsson,
  and John~J. Leonard.
\newblock Kintinuous: Spatially extended kinectfusion.
\newblock In {\em Proceedings of RSS '12 Workshop on RGB-D: Advanced Reasoning
  with Depth Cameras}, 2012.

\bibitem{xu2022point}
Qiangeng Xu, Zexiang Xu, Julien Philip, Sai Bi, Zhixin Shu, Kalyan Sunkavalli,
  and Ulrich Neumann.
\newblock Point-nerf: Point-based neural radiance fields.
\newblock In {\em IEEE/CVF Conference on Computer Vision and Pattern
  Recognition}, pages 5438--5448, 2022.

\bibitem{yan2021continual}
Zike Yan, Yuxin Tian, Xuesong Shi, Ping Guo, Peng Wang, and Hongbin Zha.
\newblock Continual neural mapping: Learning an implicit scene representation
  from sequential observations.
\newblock In {\em IEEE/CVF International Conference on Computer Vision (ICCV)},
  pages 15782--15792, October 2021.

\bibitem{yang2022vox}
Xingrui Yang, Hai Li, Hongjia Zhai, Yuhang Ming, Yuqian Liu, and Guofeng Zhang.
\newblock Vox-fusion: Dense tracking and mapping with voxel-based neural
  implicit representation.
\newblock In {\em IEEE International Symposium on Mixed and Augmented Reality
  (ISMAR)}, pages 499--507. IEEE, 2022.

\bibitem{yang2022fd}
Xingrui Yang, Yuhang Ming, Zhaopeng Cui, and Andrew Calway.
\newblock Fd-slam: 3-d reconstruction using features and dense matching.
\newblock In {\em 2022 International Conference on Robotics and Automation
  (ICRA)}, pages 8040--8046. IEEE, 2022.

\bibitem{yariv2021volume}
Lior Yariv, Jiatao Gu, Yoni Kasten, and Yaron Lipman.
\newblock Volume rendering of neural implicit surfaces.
\newblock {\em Advances in Neural Information Processing Systems},
  34:4805--4815, 2021.

\bibitem{yu2018ds}
Chao Yu, Zuxin Liu, Xin-Jun Liu, Fugui Xie, Yi Yang, Qi Wei, and Qiao Fei.
\newblock Ds-slam: A semantic visual slam towards dynamic environments.
\newblock In {\em 2018 IEEE/RSJ International Conference on Intelligent Robots
  and Systems (IROS)}, pages 1168--1174. IEEE, 2018.

\bibitem{yu2022monoSDF}
Zehao Yu, Songyou Peng, Michael Niemeyer, Torsten Sattler, and Andreas Geiger.
\newblock Monosdf: Exploring monocular geometric cues for neural implicit
  surface reconstruction.
\newblock In {\em Advances in Neural Information Processing Systems (NeurIPS)},
  2022.

\bibitem{zhang2020dense}
Heng Zhang, Guodong Chen, Zheng Wang, Zhenhua Wang, and Lining Sun.
\newblock Dense 3d mapping for indoor environment based on feature-point slam
  method.
\newblock In {\em 2020 the 4th International Conference on Innovation in
  Artificial Intelligence}, pages 42--46, 2020.

\bibitem{zhang2018unreasonable}
Richard Zhang, Phillip Isola, Alexei~A Efros, Eli Shechtman, and Oliver Wang.
\newblock The unreasonable effectiveness of deep features as a perceptual
  metric.
\newblock In {\em IEEE conference on computer vision and pattern recognition},
  pages 586--595, 2018.

\bibitem{zhou2013dense}
Qian-Yi Zhou and Vladlen Koltun.
\newblock Dense scene reconstruction with points of interest.
\newblock {\em ACM Transactions on Graphics (ToG)}, 32(4):1--8, 2013.

\bibitem{zhou2013elastic}
Qian-Yi Zhou, Stephen Miller, and Vladlen Koltun.
\newblock Elastic fragments for dense scene reconstruction.
\newblock In {\em IEEE International Conference on Computer Vision}, pages
  473--480, 2013.

\bibitem{zhu2023nicer}
Zihan Zhu, Songyou Peng, Viktor Larsson, Zhaopeng Cui, Martin~R Oswald, Andreas
  Geiger, and Marc Pollefeys.
\newblock Nicer-slam: Neural implicit scene encoding for rgb slam.
\newblock {\em arXiv preprint arXiv:2302.03594}, 2023.

\bibitem{zhu2022nice}
Zihan Zhu, Songyou Peng, Viktor Larsson, Weiwei Xu, Hujun Bao, Zhaopeng Cui,
  Martin~R Oswald, and Marc Pollefeys.
\newblock Nice-slam: Neural implicit scalable encoding for slam.
\newblock In {\em IEEE/CVF Conference on Computer Vision and Pattern
  Recognition}, pages 12786--12796, 2022.

\bibitem{zollhofer2018state}
Michael Zollh{\"o}fer, Patrick Stotko, Andreas G{\"o}rlitz, Christian Theobalt,
  Matthias Nie{\ss}ner, Reinhard Klein, and Andreas Kolb.
\newblock State of the art on 3d reconstruction with rgb-d cameras.
\newblock In {\em Computer graphics forum}, volume~37, pages 625--652. Wiley
  Online Library, 2018.

\bibitem{zou2022mononeuralfusion}
Zi-Xin Zou, Shi-Sheng Huang, Yan-Pei Cao, Tai-Jiang Mu, Ying Shan, and Hongbo
  Fu.
\newblock Mononeuralfusion: Online monocular neural 3d reconstruction with
  geometric priors.
\newblock {\em arXiv preprint arXiv:2209.15153}, 2022.

\end{thebibliography}
}

\end{document}